\documentclass[manuscript,screen]{acmart} % acmart %  preprint
\AtBeginDocument{%
  \providecommand\BibTeX{{%
    \normalfont B\kern-0.5em{\scshape i\kern-0.25em b}\kern-0.8em\TeX}}}
    
\setcopyright{acmcopyright}
\copyrightyear{2021}
\acmYear{2021}

\usepackage{natbib}
\usepackage{tikz}
\usetikzlibrary{mindmap,trees,calc}
\usepackage[T1]{fontenc}

\usepackage{natbib} % \cite{xxx} ==> [123]; \citet{xxx} ==> xxx et al.[123] !!!! Notice here!!!
\usepackage{graphicx}
\usepackage{booktabs}
\usepackage{url}
\usepackage{color}
\usepackage{multirow}
\usepackage{acronym}

\acrodef{PLM}{Pre-trained Language Model}
\acrodef{DS}{Dialogue System}
\acrodef{NLP}{Natural Language Processing}
\acrodef{PLM}{Pre-trained Language Model}
\acrodef{NLU}{Natural Language Understanding}
\acrodef{DST}{Dialogue State Tracking}
\acrodef{DPL}{Dialogue Policy Learning}
\acrodef{NLG}{Natural Language Generation}
\acrodef{MLM}{Masked Language Modeling}
\acrodef{NSP}{Next Sentence Prediction}
\acrodef{EHR}{Electronic Health Record}
\acrodef{BioNER}{Biomedical Named Entity Recognition}
\acrodef{BioRE}{Biomedical Relation Extraction}
\acrodef{NLM}{Neural Language Model}
\acrodef{FFN}{Feed-forward Network}
\acrodef{SAN}{Self-attention Network}
\acrodef{BERT}{Bidirectional Encoder Representations from Transformers}
\acrodef{LM}{Language Model}
\usepackage{caption}
\usepackage{subcaption}
\usepackage{makecell} % https://tex.stackexchange.com/questions/9796/how-to-add-todo-notes
\usepackage[colorinlistoftodos,prependcaption]{todonotes}
\usepackage{xargs}                      % Use more than one optional parameter in a new commands
\usepackage[graphicx]{realboxes}
\usepackage[font=small,labelfont=bf]{caption}
\usepackage{tablefootnote}
\usepackage{threeparttable}
\usepackage{subcaption}
\newcommandx{\improvement}[2][1=]{\todo[linecolor=Plum,backgroundcolor=Plum!25,bordercolor=Plum,#1]{#2}}
\newcommandx{\unsure}[2][1=]{\todo[linecolor=red,backgroundcolor=red!25,bordercolor=red,#1]{#2}}
\newcommandx{\change}[2][1=]{\todo[linecolor=blue,backgroundcolor=blue!25,bordercolor=blue,#1]{#2}}
\newcommandx{\info}[2][1=]{\todo[linecolor=OliveGreen,backgroundcolor=OliveGreen!25,bordercolor=OliveGreen,#1]{#2}}
\newcommandx{\thiswillnotshow}[2][1=]{\todo[disable,#1]{#2}}

\usepackage{amsmath,amsfonts,bm}

\def\eqref#1{equation~\ref{#1}}
\def\1{\bm{1}}

\def\mW{{\bm{W}}}

\DeclareMathAlphabet{\mathsfit}{\encodingdefault}{\sfdefault}{m}{sl}
\SetMathAlphabet{\mathsfit}{bold}{\encodingdefault}{\sfdefault}{bx}{n}

\begin{document}
%
% \title{\aclp{PLM} in Biomedical Domain: A Survey from Multiscale Perspective}
\title{\aclp{PLM} in Biomedical Domain: A Systematic Survey}

% \benyou{if we could summarize a summary paragraph for each subsection in Sec. 5}
%
%\titlerunning{Abbreviated paper title}
% If the paper title is too long for the running head, you can set
% an abbreviated paper title here
%
%!TEX root = ./main.tex

% \author{Qianqian Xie \inst{1}\orcidID{0000-1111-2222-3333} \and
% Benyou Wang\inst{2,3}\orcidID{1111-2222-3333-4444} \and
% Zhao Li\inst{3}\orcidID{2222--3333-4444-5555}}
% %
% \authorrunning{F. Author et al.}
% % First names are abbreviated in the running head.
% % If there are more than two authors, 'et al.' is used.
% %
% \institute{Princeton University, Princeton NJ 08544, USA \and
% Springer Heidelberg, Tiergartenstr. 17, 69121 Heidelberg, Germany
% \email{lncs@springer.com}\\
% \url{http://www.springer.com/gp/computer-science/lncs} \and
% ABC Institute, Rupert-Karls-University Heidelberg, Heidelberg, Germany\\
% \email{\{abc,lncs\}@uni-heidelberg.de}}
% %

\author{Benyou Wang}
\email{wangbenyou@cuhk.edu.cn}
\affiliation{%
  \institution{SRIBD \& SDS, The Chinese University of Hong Kong, Shenzhen}
%   \streetaddress{P.O. Box 1212}
%   \city{Dublin}
%   \state{Ohio}
  \country{China}
%   \postcode{43017-6221}
}
\author{Qianqian Xie}
\authornote{Qianqian Xie is the corresponding author: xqq.sincere@gmail.com.}
\email{qianqian.xie@manchester.ac.uk}
% \orcid{1234-5678-9012}
\affiliation{%
  \institution{Department of Computer Science, University of Manchester}
  \country{United Kingdom}
}
\author{Jiahuan Pei}
% \authornotemark[1]
\email{j.pei@uva.nl}
\affiliation{%
  \institution{University of Amsterdam}
  \country{Netherlands}
}

\author{Zhihong Chen}
% \authornotemark[1]
\email{zhihongchen@link.cuhk.edu.cn}
\affiliation{%
  \institution{SRIBD \& SSE, The Chinese University of Hong Kong, Shenzhen}
  \country{China}
}

\author{Prayag Tiwari}
% \authornotemark[1]
\email{prayag.tiwari@ieee.org}
\orcid{0000-0002-2851-4260}
\affiliation{%
  \institution{School of Information Technology, Halmstad University}
  \country{Sweden}
}
\author{Zhao Li}
% \authornotemark[1]
\email{lizhao.informatics@gmail.com}
\affiliation{%
  \institution{The University of Texas Health Science Center at Houston}
  \country{USA}
}
\author{Jie Fu}
% \authornotemark[1]
\email{jie.fu@polymtl.ca}
\affiliation{%
  \institution{Mila, University of Montreal}
  \country{Canada}
}

% \author{Lars Th{\o}rv{\"a}ld}
% \affiliation{%
%   \institution{The Th{\o}rv{\"a}ld Group}
%   \streetaddress{1 Th{\o}rv{\"a}ld Circle}
%   \city{Hekla}
%   \country{Iceland}}
% \email{larst@affiliation.org}

% \author{Valerie B\'eranger}
% \affiliation{%
%   \institution{Inria Paris-Rocquencourt}
%   \city{Rocquencourt}
%   \country{France}
% }

% \author{Aparna Patel}
% \affiliation{%
%  \institution{Rajiv Gandhi University}
%  \streetaddress{Rono-Hills}
%  \city{Doimukh}
%  \state{Arunachal Pradesh}
%  \country{India}}

% \author{Huifen Chan}
% \affiliation{%
%   \institution{Tsinghua University}
%   \streetaddress{30 Shuangqing Rd}
%   \city{Haidian Qu}
%   \state{Beijing Shi}
%   \country{China}}

% \author{Charles Palmer}
% \affiliation{%
%   \institution{Palmer Research Laboratories}
%   \streetaddress{8600 Datapoint Drive}
%   \city{San Antonio}
%   \state{Texas}
%   \country{USA}
%   \postcode{78229}}
% \email{cpalmer@prl.com}

% \author{John Smith}
% \affiliation{%
%   \institution{The Th{\o}rv{\"a}ld Group}
%   \streetaddress{1 Th{\o}rv{\"a}ld Circle}
%   \city{Hekla}
%   \country{Iceland}}
% \email{jsmith@affiliation.org}

% \author{Julius P. Kumquat}
% \affiliation{%
%   \institution{The Kumquat Consortium}
%   \city{New York}
%   \country{USA}}
% \email{jpkumquat@consortium.net}

%%
%% By default, the full list of authors will be used in the page
%% headers. Often, this list is too long, and will overlap
%% other information printed in the page headers. This command allows
%% the author to define a more concise list
%% of authors' names for this purpose.
\renewcommand{\shortauthors}{Wang. et al.}

% such as text classification, sequential labeling, and language generation. 

% There have been attempts to develop biomedical pre-trained language models, 
%However, it seems that the vast majority of existing work does not take into account related and previous work and for the most part seems to exist in isolation from each other, 

\begin{abstract}
Pre-trained language models (PLMs) have been the de facto paradigm for most natural language processing (NLP) tasks. This also benefits the biomedical domain:  researchers from informatics, medicine, and computer science (CS) communities \textcolor{black}{propose} various PLMs trained on biomedical datasets, e.g., biomedical text, electronic health records, protein, and DNA sequences for various biomedical tasks.
However, the cross-discipline characteristics of biomedical PLMs hinder their spreading among communities; some existing works are isolated from each other without comprehensive comparison and discussions.
It is nontrivial to make a survey that not only systematically reviews recent advances in biomedical PLMs and their applications but also standardizes terminology and benchmarks. 
This paper summarizes the recent progress of pre-trained language models in the biomedical domain and their applications in downstream biomedical tasks.
Particularly, we discuss the motivations \textcolor{black}{of PLMs in the biomedical domain and introduce the key concepts of pre-trained language models.
We then propose a taxonomy of existing biomedical PLMs, which categorizes them from various perspectives systematically}. 
Plus, their applications in biomedical downstream tasks are exhaustively discussed, respectively.
At last, we illustrate various limitations and future trends, 
which aims to provide inspiration for the future research of the research community. 
\end{abstract}

% \ccsdesc[300]{Computing methodologies~Classification and regression trees}

% \keywords{datasets, neural networks, gaze detection, text tagging}

%\listoftodos[Notes]
\maketitle              % typeset the header of the contribution
\newpage
\setcounter{tocdepth}{2}
\tableofcontents
\newpage

\section{Introduction}
% \benyou{ 

% https://arxiv.org/abs/2204.03905 BioBART \cite{yuan2022biobart}

% https://academic.oup.com/bib/advance-article-abstract/doi/10.1093/bib/bbac409/6713511 BIOGPT  \cite{luo2022biogpt}

% https://arxiv.org/abs/2201.11838  Clinical-Longformer and Clinical-BigBird: Transformers for long clinical sequences    \cite{li2022clinical}

% https://arxiv.org/pdf/2210.10341.pdf

% https://academic.oup.com/jamia/advance-ariticle/doi/10.1093/jamia/ocac225/6855145  long text  \cite{10.1093/jamia/ocac225}  A comparative study of pretrained language models for long clinical text

% @benyou: BioLinkBERT: https://arxiv.org/pdf/2203.15827.pdf \cite{yasunaga2022linkbert}

% }

% \benyou{what is the knowledge and how do we use it!}

% \benyou{https://academic.oup.com/bioinformatics/article-abstract/38/2/494/6374496}

% \benyou{Jie suggested to refer to a survey from Gao Huang, https://arXiv.org/pdf/2102.04906.pdf}

% \qq{this github project have some resources including datasets, workshop and so on.https://github.com/caufieldjh/awesome-bioie}

% \qq{a github project which introduces progress on the biomedical ner.  https://github.com/lingluodlut/BioNER-Progress}

% \qq{a survey paper very similar to our's: AMMU – A Survey of Transformer-based Biomedical Pretrained Language Models}

% \benyou{\url{https://github.com/lrs1353281004/Chinese_medical\_NLP}}

% \benyou{\url{https://github.com/caufieldjh/awesome-bioie}}
As the principal method of communication, humans usually record information and knowledge in a format of \textit{token} sequences, 
% resulting in languages such as natural language, constructed language, and programming language. 
e.g., natural languages, time series, constructed knowledge base, etc. 
For biomedical information and knowledge, tokens in sequences could be of various types, including words, disease codes, amino acids, and DNAs. 
Tremendous biomedical information and knowledge in nature and human history are implicitly encapsulated in these natural token sequences in nature (\textit{a.k.a.}, data).

% Based on the \textit{abstraction degree of biomedical knowledge}, we organize various types of data as a pyramid in Fig. \ref{fig:pyramid}. 

{ \color{black}
There exist many data that involve biomedical information with different abstraction degrees of biomedical knowledge. However, there is a trade-off between the high abstraction degree and its scale. For data that explicitly conveys biomedical knowledge (i.e. at a high abstraction degree), it is usually small-scaled, see biomedical knowledge bases and EHR data (maybe in multi-modality).
One example of data that may not directly convey biomedical knowledge could be protein and DNA sequences,  since one can hardly know what a short protein or DNA sequence really means for humans and it needs more effort for abstraction. Fortunately, these data are usually tremendous. 
In the current stage, existing work pays more attention to data at a high abstraction level (biomedical knowledge-intensive data, e.g., EHR,  biomedical knowledge bases, and biomedical encyclopedia); however, it is usually relatively small-scale. 
We argue that biomedical knowledge on various abstraction degrees should be paid attention to.
To capture and mine the biomedical information and knowledge from various abstraction degrees, there is recently growing attention in the biomedical natural language processing (NLP) community to adopt pre-trained language models (PLMs); 
since PLMs could leverage these massive sequences without biomedical knowledge abstraction and human annotations, including but not limited to plain biomedical text, biomedical images, general text, protein sequences, and DNA sequences. 
}

% (sometimes called `self-supervision').
% \begin{figure}[t]
%     \begin{minipage}[t]{0.38\linewidth}
%     \centering
%     \includegraphics[height=5.2cm]{figures/pyramid.pdf}
%     \caption{
%     Data pyramid for biomedical knowledge. 
%     The level of a type of data is determined by to what extent humans abstract biomedical knowledge from data. \benyou{more explanation}
%     %Data at a higher level contains explicit  biomedical knowledge while data at a lower level \textcolor{black}{needs} more effort to mine biomedical knowledge inside the data.
%     } % \benyou{ELECTRAMed, }
%     \label{fig:pyramid}
%   \end{minipage}
%   \begin{minipage}[t]{0.6\linewidth}
%     \centering
%     \includegraphics[width=1\linewidth]{figures/overview_bioPLM.pdf}
%     \caption{Overview of selected released Biomedical pre-trained language models. One can see a more detailed list in Sec. \ref{sec:pertraining}. Note that there is a BERT-like language model embedded in the overall architecture of AlphaFold2. \benyou{add more models like BioGPT, BioGPT, and dell-e, Clip, Flamigo, ect...} } % \benyou{ELECTRAMed, }
%     \label{fig:overview}
%     \end{minipage}
% \end{figure}

\begin{figure}[t]
    \centering
   \includegraphics[width=0.7\linewidth]{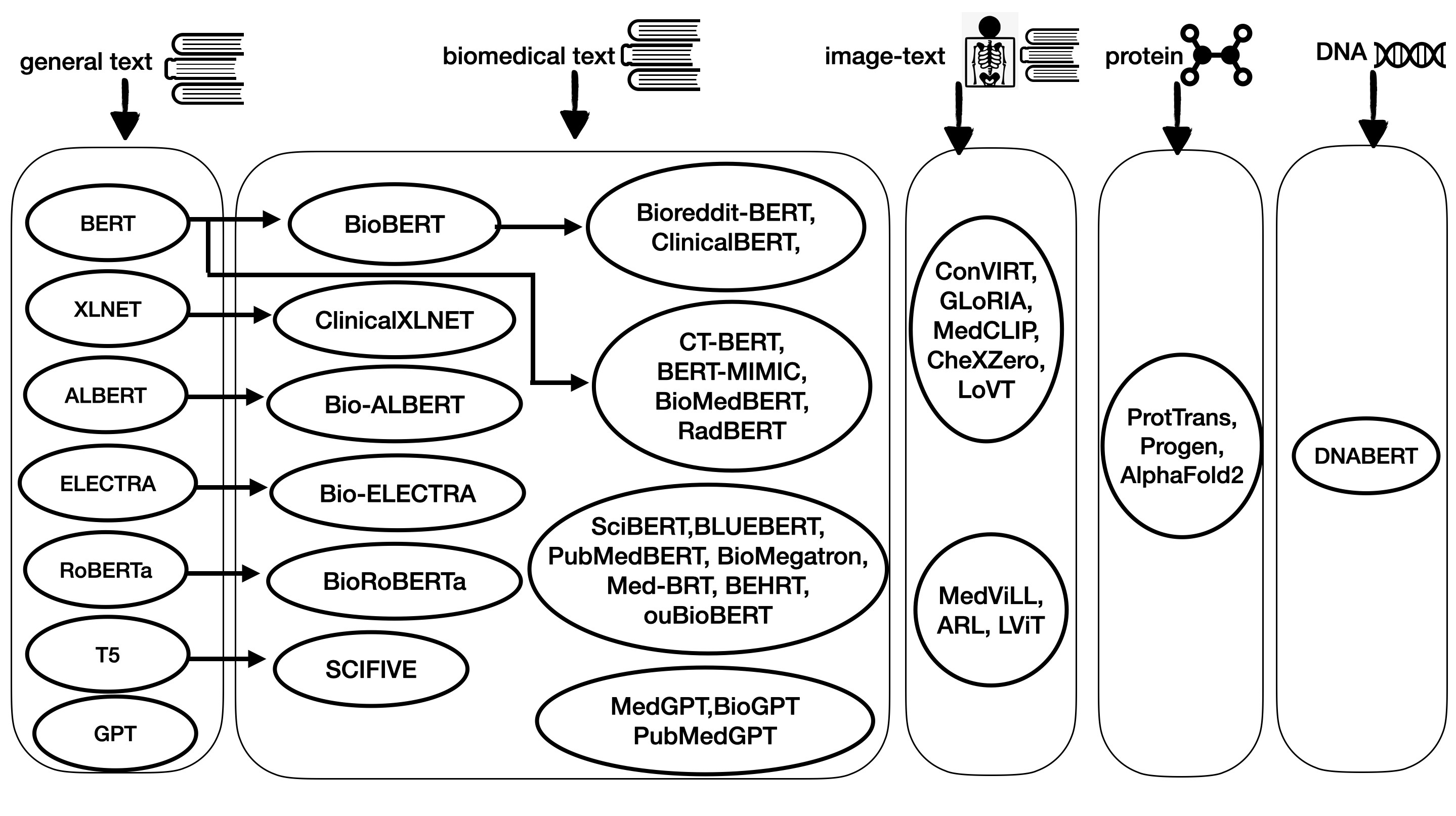}
   \caption{Overview of selected released Biomedical pre-trained language models. One can see a more detailed list in Sec. \ref{sec:pertraining}. Note that there is a BERT-like language model embedded in the overall architecture of AlphaFold 2.} % \benyou{ELECTRAMed, }
    \label{fig:overview}
\end{figure}

The biomedical NLP is a cross-discipline research direction from various communities such as bioinformatics, medicine, and computer science (especially a major frontier of artificial intelligence, \textit{i.e.}, natural language processing \textit{a.k.a.} NLP).
The computational biology community \cite{10.1145/234313.234358} and biomedical informatics community \cite{10.1145/1031120.1031122} have made a substantial effort to make use of NLP tools for information mining and extraction of widespread-adopted electronic health records, medical scientific publications, medical WIKI pages, etc. 
For many decades, NLP has been investigating various biomedical tasks~\cite{cohen2014biomedical,cohen2005survey} such as classification, information extraction, question answering, drug discovery et al. 
Meanwhile, the approaches in the NLP community are changing rapidly, as one can witness exponentially increasing submitted papers in top conferences like ACL, EMNLP, and NAACL. 
Tailoring these NLP approaches that have been evidenced effectively in the NLP community to a specific biomedical domain is beneficial. 

Unfortunately, there is usually a delay for newly proposed NLP approaches being applied to the biomedical domain. 
Especially, since the adoption of various pre-trained language models (\textit{\textit{e.g.}}, ELMo \cite{peters2018deep}, GPT \cite{radford2018improving}, BERT \cite{devlin2018bert}, XLNET \cite{clinicalxlnet}, RoBERTa \cite{liu2019roberta}, T5 \cite{raffel2020exploring} and ELECTRA \cite{clark2020electra}) \cite{qiu2020pre} have nearly shifted the paradigm in NLP, their biomedical variants trained using biomedical data comes sooner or later. 
With this hot trend of the biomedical pre-trained language model, this survey aims to bridge the gap between pre-trained language models and their applications in the biomedical domain. 
% \begin{table}[h]
%     \centering
%     \caption{ Cross-discipline nature of biomedical NLP.  }
%     \begin{tabular}{lllll}
%     \toprule
%          Community & Representative conferences/journals \\
%     \midrule
%          Informatics  &   Bioinformatics, AMIA, JAMIA, JBI;  \\
%         Medicine &  JAMA, Nature Medicine; \\ %Radiology, Drug Safety
%       Computer science  &  ACL, EMNLP, NAACL, NeurIPs, AAAI, ICLR;\\
%     \bottomrule
%     \end{tabular}
%     \label{tab:my_label}
% \end{table}
% \begin{table} \small[]
%     \centering
%     \caption{Publication statistics in AI conference and biomedical journals.  }
%     \begin{tabular}{lllll}
%     \toprule
%          community & representative conference/journal & name (\# papers) \\
%     \midrule
%          \multirow{2}{*}{informatics}  & conference & AMIA, JAMIA, or JBI;  \\
%           & journal\\
%          \multirow{2}{*}{medicine}  & conference\\
%           & journal & Radiology, Drug Safety, or Nature
% Medicine; \\
%         \multirow{2}{*}{computer science}   &conference  & ACL (1) ; EMNLP (1); NAACL (1)\\
%         &journal & \\
%     \bottomrule
%     \end{tabular}

%     \label{tab:my_label}
% \end{table}
\paragraph{\textbf{Motivation of pre-trained language models in biomedical domain}}
The current NLP paradigm is gradually shifting to a two-stage (pre-training and fine-tuning) paradigm, thanks to recently proposed pre-trained language models. \textcolor{black}{Compared} to the previous paradigm with purely supervised learning that relies on feature engineering or neural network architecture engineering \cite{liu2021pre}, the current two-stage paradigm is more friendly to the scenario when supervised data is limited while large-scaled unsupervised data is tremendous. Fortunately, the biomedical domain is a typical case of such a scenario. 

The motivation to use pre-trained language models in the biomedical domain \textcolor{black}{is} pretty straightforward. 
First, annotated data in the biomedical domain is usually not large-scale. Therefore, a well-trained pre-trained language model is more crucial to provide a richer feature extractor, which may slightly reduce the dependence on annotated data. 
Second, the biomedical domain is more knowledge-intensive than the general domain. 
At the same time, pre-trained language models could serve as an easily-used soft knowledge base \cite{petroni2019language} that
captures implicit knowledge from large-scale plain documents without human annotations. More recently, GPT3 has been shown to have the potential to `remember' many complicated common knowledge~\cite{brown2020language}. 
Lastly, large-scaled biomedical corpora and biomedical sequences (including proteins and DNAs), which are previously thought as difficult to handle, can be effectively handled by pre-trained language models (especially  transformers networks).
% , to capture efficient information via model pre-training in the large scale data.

\begin{figure}[t]
    \centering
    \includegraphics[width=0.95\linewidth]{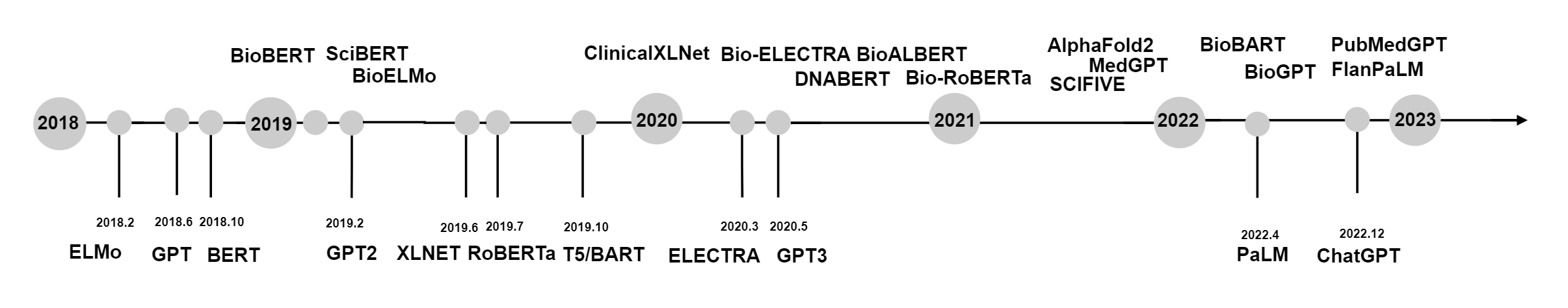}
    \caption{Parallel development of general and biomedical pre-trained language models. The time is determined by the released date of the paper, for example, in arXiv. General pre-trained language models are shown below in the timeline, and biomedical pre-trained language models are shown above the timeline (refer to Tab. \ref{tab:models_overview} for detailed dates). } % \benyou{ see \url{https://gitmind.cn/app/flowchart/28b1949428}}
    \label{fig:timeline}
\end{figure}

As shown in Fig. \ref{fig:timeline}, in recent three years, we have witnessed a rapid development of pre-trained language models (\textit{e.g.}, ELMo \cite{peters2018deep}, GPT \cite{radford2018improving}, BERT \cite{devlin2018bert}, XLNet \cite{clinicalxlnet}, RoBERTa \cite{liu2019roberta}, T5 \cite{raffel2020exploring} and ELECTRA \cite{clark2020electra}) in the general NLP domain. 
\textcolor{black}{Following these progresses}, there are efforts to tailor these pre-trained language models to their corresponding biomedical variants, via in-domain data.
% Thanks for the publicly available pre-trained language models for general domain, many biomedical pre-trained language models have been proposed.  
For example, BERT, the most typical pre-trained language, has many variants in the biomedical domain, \textit{e.g.}, 
Med-BERT \cite{abs-2005-12833}, BioBERT \cite{LeeYKKKSK20}, publicly available Clinical BERT Embeddings \cite{abs-1904-03323}, SciBERT \cite{BeltagyLC19},
ClinicalBERT \cite{abs-1904-05342}, and COVID-twitter-BERT \cite{abs-2005-07503} et al. We draw an overview for these models in Fig.~\ref{fig:overview}. It shows that the extensions of general domain pre-trained language models to the biomedical domain attract great attention from researchers in both NLP and bioinformatics communities.
%this is a very popular direction. 
Interestingly, we can observe that once the general NLP community develops a new variant of PLM, it usually leads to a biomedical counterpart after some months. 
This parallel development between general PLMs and biomedical PLMs shows a strong demand and even a necessity to summarize the existing works, which could help beginners to start their contributions in this field easily.

\paragraph{\textbf{Difference with existing surveys}}
There are a few reviews to summarize the NLP applications in the biomedical, clinical, bioinformatic domain, such as an early one \cite{spyns1996natural} and recent ones \cite{zeng2015survey,wu2020deep,percha2020modern}.  They cover many general methods and applications of biomedical/clinical NLP.  Specifically, \cite{spyns1996natural} mainly discuss either based on statistics-based NLP pipeline (including lexicon, co-occurrence patterns, syntactic/semantic parsing), or word embeddings based neural network approaches (it was mentioned that 60.8\% of them are based on recurrent neural networks)~\cite{wu2020deep} for NLP applications (\textit{e.g.}, information extraction, text classification, named entity recognition, and relation extraction et al). Especially, two reviews \cite{khattak2019survey,kalyan2020secnlp} discuss the word embeddings used in biomedical NLP.
% Also, \cite{percha2020modern} provide a review for these reviews.

All the above reviews made thorough summarization of existing work before the pre-trained language model era of NLP. The NLP techniques in these reviews are mainly about feature engineering, or architecture engineering \cite{liu2021pre}.
However, the NLP recently has been shifted to a pre-training and then fine-tuning paradigm with large-scale pre-trained language models (see existing surveys \cite{qiu2020pre,han2021pretrained,liu2020survey,liu2021pre,bommasani2021opportunities} for pre-trained language model in the general domain). 
\cite{bommasani2021opportunities} called these pre-trained models as `foundation models' to underscore their critically central.
We believe the biomedical NLP applications have \textcolor{black}{benefited} and will continually benefit from the development of pre-trained language models. 

More recently, \cite{kalyan2021ammu} %\footnote{The paper is publicly-available on the arXiv without peer review} 
reviews biomedical textual pre-training, especially using BERT. 
The difference between \cite{kalyan2021ammu} and this review is that \textcolor{black}{Our paper provides a more inclusive taxonomy of biomedical PLMs than \cite{kalyan2021ammu}, which are three fold.}. \textcolor{black}{First, biomedical PLMs summarized in our review are not limited to that trained on texts like \cite{kalyan2021ammu}, but also other data resources including protein, DNA, and even biomedical text-image pairs.  In general, any data that involves biomedical information could be used in biomedical PLMs.}
        % we envision biomedical multi-modal pre-trained language models based on the multi-modal data (e.g., in electronic health records)
Second, in contrast to \cite{kalyan2021ammu} which only discusses Transformer-based pre-trained language models, this review also discusses RNN-based language models (like ELMO \cite{jin2019probing}, which is typically considered as the first pre-trained language model in NLP). We also summarize decoder involved \textit{generative} pre-trained language models (like GPT \cite{kraljevic2021medgpt} and T5 \cite{phan2021scifive}), while \citep{kalyan2021ammu} mainly discusses encoder-based PLMs (BERT or BERT variants). 
Third, to the best of our knowledge, this is the first survey paper to discuss  pre-trained \textbf{vision-language} models in the biomedical  domain.
        % \item 
Last, \textcolor{black}{our paper provides a more comprehensive overview of the applications of PLMs in the biomedical domain compared with~\cite{kalyan2021ammu}. Except for biomedical NLP tasks such as natural language inference, text summarization~\cite{xie2023survey}, relation extraction et al that are summarized in \cite{kalyan2021ammu}, our paper further reviews recent PLMs-based methods for event detection, dialogue systems, as well as protein and DNA sequence. Moreover, compared with \cite{kalyan2021ammu} that only reviews recent methods of biomedical NLP tasks coarsely, we make a thorough categorization and discussion of PLMs-based methods for biomedical NLP tasks and their benchmark datasets. Our paper also introduces competitions and venues such as shared tasks.} 
% \item to be continued
Therefore, we believe there is a requirement for a more thorough survey paper to review the recent progress of pre-trained language models in the biomedical domain from a multi-scale perspective.
% The efficacy of BERT in pre-trained language models has been studied in  [BERT efficacy on scientific and medical datasets: a systematic literature review] that evidences that [] with something.

% raghu2020survey. :  A Survey of Deep Learning for Scientific Discovery
% \benyou{A paragraph about 'what this paper will discuss  and what will not discuss ?'} 
\paragraph{\textbf{Contribution}}
The contributions of the paper can be summarized as follows: 
\begin{itemize}
    \item  We give a comprehensive review to summarize existing \textcolor{black}{PLMs-based methods for the biomedical domain, which thoroughly categorizes and discusses biomedical data sources, biomedical PLMs,} model variants, downstream tasks, shared competitions, etc.
    \item We propose a taxonomy of biomedical PLMs, which classifies existing PLMs in the biomedical domain from various perspectives: \textcolor{black}{training data sources, model architecture, etc.} 
    \item We enumerate existing resources for PLMs and their detailed configuration, facilitating their spreading for beginners.
    \item We discuss the limitations of existing methods and prospect future trends.
    \item {\color{black} To the best of our knowledge, this is the first survey paper to summarize generative pre-trained language models,  protein/DNA language models,  pre-trained \textbf{vision-language} models  in the biomedical  domain.}
    % This will be beneficial for beginners from both the computer science and bioinformatics fields.     
\end{itemize} 
% \begin{figure}
%     \centering
%     \includegraphics[width=0.95\linewidth]{figures/paper-arthietecture.png} %,height=18cm
%     \caption{\textcolor{black}{Architecture of this paper.}} 
%     \label{fig:art}
% \end{figure}
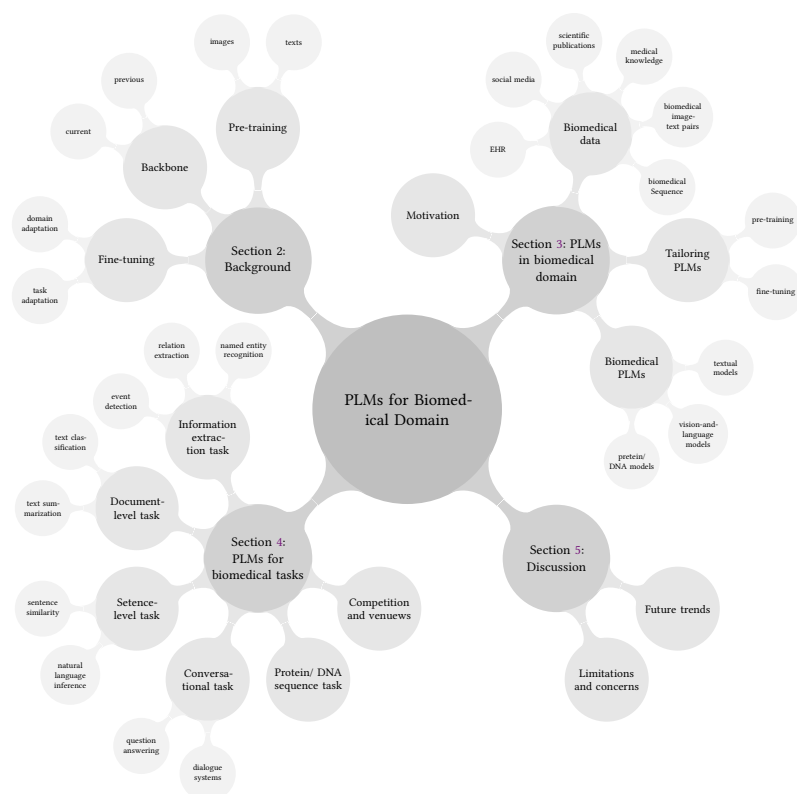
\begin{figure}[t]
\centering
\resizebox{0.70\textwidth}{!}{% \documentclass{standalone}
 
% Required package
% \usepackage{tikz}
% \usetikzlibrary{mindmap,trees,calc}
% \usepackage[T1]{fontenc}
% \tikzset{every picture/.style={/utils/exec={\sffamily}}}

\begin{tikzpicture}[
mindmap,
grow cyclic,
every annotation/.style={fill=red!20},
every node/.style=concept,
concept color=black!25, 
level 1/.append style={concept color=black!18,level distance=4.5cm,sibling angle=90},
level 2/.append style={concept color=black!10,level distance=2.8cm,sibling angle=45},
level 3/.append style={concept color=black!5,level distance=2cm,sibling angle=45},
]

\node{PLMs for Biomedical Domain}
    child {node {Section \ref{sec:finetuning}: PLMs for biomedical tasks}
        child {node {Information extraction task}
            child {node {named entity recognition}}
            child {node {relation extraction}}
            child {node {event detection}}
        }
        child {node {Document-level task}
            child {node {text classification}}
            child {node {text summarization}}
        }
        child {node {Setence-level task}
            child {node {sentence similarity}}
            child {node {natural language inference}}
        }
        child {node {Conversa-tional task}
            child {node {question answering}}
            child {node {dialogue systems}}
        }
        child {node {Protein/ DNA sequence task}}
        child {node {Competition and venuews}}
    }
    child {node {Section \ref{sec:discussion}: Discussion}
        child {node {Limitations and concerns}}
        child {node {Future trends}}
    }
    child {node {Section \ref{sec:pertraining}: PLMs in biomedical domain}
        child [clockwise from=-100] {node {Biomedical PLMs}
            child [clockwise from=-45] {node {textual models}}
            child [clockwise from=-45] {node {vision-and-language models}}
            child [clockwise from=-45] {node {pretein/ DNA models}}
        }
        child [clockwise from=0] {node {Tailoring PLMs}
            child [clockwise from=-20] {node {pre-training}}
            child [clockwise from=-20] {node {fine-tuning}}
        }
        child [clockwise from=120] {node {Biomedical data}
            child [clockwise from=145] {node {EHR}}
            child [clockwise from=145] {node {social media}}
            child [clockwise from=145] {node {scientific publications}}
            child [clockwise from=145] {node {medical knowledge}}
            child [clockwise from=145] {node {biomedical image-text pairs}}
            child [clockwise from=145] {node {biomedical Sequence}}
        }
        child [clockwise from=-110] {node {Motivation}}
    }
    child {node {Section 2: Background}
        child {node {Pre-training}
            child {node {texts}}
            child {node {images}}
        }
        child {node {Backbone}
            child {node {previous}}
            child {node {current}}
        }
        child {node {Fine-tuning}
            child {node {domain adaptation}}
            child {node {task adaptation}}
        }
    };
\end{tikzpicture}
% \end{document}
}
\caption{\textcolor{black}{Architecture of this survey.}} 
\label{fig:art}
\end{figure}

\paragraph{\textbf{How do we collect the papers?}}
In this survey, we collected over a hundred related papers. We used Google Scholar as the main search engine, and also adopted MedPub, Web of Science, as an essential tool to discover related papers. 
In addition, we screened most of the related conferences and journals such as ACL, EMNLP, NAACL, AAAI, Bioinformatics, JAMIA, AMIA, etc. The major keywords we used \textcolor{black}{included} medical pre-trained language model, clinical pre-trained language model, biological language model, etc. Plus, we take Med-BERT \cite{abs-2005-12833}, BioBERT \cite{LeeYKKKSK20}, SciBERT \cite{BeltagyLC19},
ClinicalBert \cite{abs-1904-05342}, COVID-twitter-BERT \cite{abs-2005-07503} as the seed papers to check papers that cited them.

\paragraph{\textbf{Organization}}
\textcolor{black}{The overall architecture of this paper is shown in Figure \ref{fig:art}.}
The paper is organized as below: Sec.\ref{sec:background} introduces the general pre-trained language models \textcolor{black}{including backbone networks, pre-training objective, pre-training corpora, fine-tuning, and categorization of PLMs.
%Biomedical data sources are introduced and categorized in Sec. \ref{fig:data}.
Sec.\ref{sec:pertraining} introduces the pre-trained language models for the biomedical domain and proposes a taxonomy, including motivations for using PLMs, biomedical data sources, domain-specific pre-training, biomedical PLMs, and their categorization.
Sec.\ref{sec:finetuning} summarizes the applications of biomedical PLMs for various downstream tasks and categorizes existing methods for these tasks respectively.} 
More discussions about limitations and future directions are in Sec. \ref{sec:discussion}. We conclude in Sec.\ref{sec:conclusion}.

% \benyou{language models, what is language models (bi-gram language model) and its context here, "text has always been a key direction in NLP, from early linguistic insightabout distributional similarity to the use of Brown cluster and more traditional language models (e.g., HMMs) to context-free wordembedding right preceding the current wave of neural language models. The advance is in the technical front, with compute andmodern architectures providing a more effective way to leverage unlabeled text." This is not included in this survey!}

\section{Background: \aclp{PLM}}
% \subsection{\acf{PLM}}
\label{sec:background}
Pre-trained language models (PLMs) have been widely used in natural language processing, etc., \textcolor{black}{due to their effectiveness to learn useful representations from unannotated  data such as} natural languages. 
In this paper, we mainly discuss pre-trained language in sequential tokens \footnote{Tokens usually refers to words or subwords in NLP, and also protein sequences in the biomedical domain.}. 
We will introduce the textual pre-training in Sec.~\ref{sec:text-pre-training}, one can read the review paper of PLMs in \cite{qiu2020pre} for more details. Thanks to the popularity of  CLIP,  \textcolor{black}{pre-trained language models are also usually jointly trained with a visual pre-trained model in the image-text pre-training  scenario. We will also discuss visual pre-training in Sec.~\ref{sec:image-pre-training}.
Note that models in the visual pre-training usually treats image patches as visual tokens, this makes it language model-like pre-training; we, therefore, include visual pre-training models in this survey.}

In this section, we will introduce the basic ingredients of pre-training models: the training objective with self-supervised tasks \textcolor{black}{ and corpora} in Sec.~\ref{sec:text-pre-training} and Sec.~\ref{sec:image-pre-training} for text and images respectively, basic neural network models in  Sec. \ref{sec:backbone}, and training paradigm in Sec. \ref{sec:paradiam}.

\subsection{Backbone Networks in Language Models}
\label{sec:backbone}
% \begin{figure}
%     \centering
%     \includegraphics[width=0.65\textwidth]{figures/LSTM3-chain.png}
%     \caption{The repeating module in an LSTM. {The figure is from \url{https://colah.github.io/posts/2015-08-Understanding-LSTMs/} } }
%     \label{fig:lstm} %\footnote{\url{https://colah.github.io/posts/2015-08-Understanding-LSTMs/}}
% \end{figure}
The success of pre-trained language models is also attributed to the development of their base backbone network, from LSTM~\cite{hochreiter1997long} to Transformer~\cite{vaswani2017attention}. Before Transformer was invented, LSTM was widely used as the base architecture of pre-trained language models such as ELMO.  However, because of its recurrence structure, it is computationally expensive to scale up LSTM to be deeper in layers. To this end, Transformer is proposed and becomes the backbone of modern NLP.
Transformers are better architecture can be attributed to: 1) efficiency: a recurrent-free architecture that could compute the individual token in parallel, 2) effectiveness: attention allows spatial interaction across tokens that dynamically depends on the input itself. In this section, we briefly introduce the two typical architectures in pre-trained language models, namely, LSTM and Transformers.

\subsubsection{\textcolor{black}{Previous backbone networks in texts}}

\paragraph{LSTM} Long short-term memory
(LSTM) is a recurrent neural network (RNN) architecture for sequential modeling. Unlike standard feed-forward neural networks processing single data points (such as images), LSTM can deal with entire sequences of data (such as text, speech, or video).
% As shown in Fig. \ref{fig:lstm}, 
A common LSTM unit is composed of a cell, an input gate, an output gate, and a forget gate. The cell learns hidden states  over arbitrary time intervals and the three gates regulate the flow of information into and out of the cell.
LSTM networks are well-suited for time series data and were developed to deal with the vanishing gradient problem that can be encountered when training traditional RNNs. Peters et al
\cite{peters2018deep} tried to adopt a Long and Short term memory network (LSTM) in pre-trained language, which naturally processes tokens sequentially. 
% \begin{figure}
%      \centering
%      \begin{subfigure}[b]{0.25\textwidth}
%          \centering         \includegraphics[width=\textwidth]{figures/transformer.PNG}
%          \caption{Transformer Architecture}
%          \label{fig:y equals x}
%      \end{subfigure}
%      \hfill
%      \begin{subfigure}[b]{0.6\textwidth}
%          \centering
%          \includegraphics[width=\textwidth]{figures/mha.PNG}
%          \caption{Multi-head Attention module in Transformer}
%          \label{fig:three sin x}
%      \end{subfigure}
%     \caption{Transformer Architecture and its core module `Multi-head Attention' \cite{vaswani2017attention} }
% \end{figure}
\subsubsection{\textcolor{black}{Previous backbone networks in images}}

{\color{black}

\paragraph{CNNs}

Convolutional neural networks~\cite{lecun1998cnn} (CNNs) are a type of neural networks that are particularly suited for vision tasks. Typically, CNNs are made up of four main types of layers: convolution, pooling, activation, and fully connected layers. The convolution layers are trainable filters that can learn to recognize patterns in images, such as edges, textures, and objects; The pooling layers are used to reduce the dimensionality of the data; The activation layers are used to introduce non-linearity to the network; The fully connected layers are used to make predictions based on the extracted features. Note that CNNs are also a good choice for language understanding~\cite{kim2014textcnn}.

}

\subsubsection{\textcolor{black}{The current backbone networks in texts and images}}

\paragraph{Transformer}
The backbone of most pre-trained language models (\textit{e.g.}, BERT, its variants, GPT, T5 et al) is a neural network called `Transformer ' building upon self-attention networks (SANs) and feed-forward networks (FFNs). SAN is used to facilitate interaction between tokens, while FNN is used to refine the token presentation using non-linear transformation. 
Since Transformer has been the de facto backbone to replace recurrent and convolutional units, almost all language models adopt the Transformer as the backbone network. The transformer is superior in terms of capacity and scalability thanks to, 1) discarding recurrent units and process tokens more efficiently in parallel with the position embeddings\cite{wang2019encoding,wang2021position}, 2) relieving saturation issue of expressive power with large-scale data and very deep layers due to the well-designed architecture including residual connections, layer normalization, and etc.

{ \color{black}

% \begin{figure}
%     \centering
%     \includegraphics[width=0.8\textwidth]{figures/transformer.PNG}
%     \caption{Transformer architecture.}
%     \label{fig:transformer}
% \end{figure}
% (see Fig. \ref{fig:transformer})

A Transformer layer  consists of a self-attention (SAN) module and a feed-forward network (FFN) module. An input $X$ \footnote{  $X$ is the word embedding of each individual input token which are tokenized using  subword tokenization. Moreover, the input is usually concatenated with position embeddings~\cite{wang2020position} to perceive word order } for SAN will be linearly transformed  into  query, key, value, and  output space $\{ Q, K, V\}$ as below \footnote{For all linear transformation in this paper, the bias term is in default omitted}: 
\begin{equation}
\small
% \textrm{APE:} \;
\begin{bmatrix}
Q \\
K \\
V  \\
\end{bmatrix}
=
X \times \begin{bmatrix}
\mW^Q \\
\mW^K  \\
\mW^V  \\
\end{bmatrix}
\end{equation}

The self-attention mechanism (a.k.a  Scaled Dot-Product Attention) is calculated as 
\begin{equation}
% \small
\textrm{Attention(Q,K,V)} = \textrm{softmax}( \frac{QK}{\sqrt{d_k}}) V
\end{equation}

% Since SAN layer does not perceive word order, the input of word embedding and position embeddings designed as a sum of word embedding and position embedding, namely, $X = \textrm{WE} + \textrm{PE}$.

For a multi-head version  of the self-attention mechanism, it  linearly projects $Q,K,V$ with $h$ times using individual linear projections to smaller  dimensions (e.g.  $d_k = \frac{d_\textrm{model}}{h}$), instead of performing a single attention function with $d_\textrm{model}$-dimensional keys, values and queries. Finally, the output of SAN is 
\begin{equation}
 \begin{aligned}
    \textrm{SAN} (X) &=  [\textrm{head}_1;\cdots ;\textrm{head}_h ] \mW^O \\
     \textrm{head}_i &= \textrm{Attention}(Q_i,K_i,V_i),
\end{aligned}
\end{equation}
where $\quad   Q = [Q_1;\cdots Q_h]$, $ K = [K_1;\cdots K_h]$,  and $ V = [V_1;\cdots V_h]$. The individual attention  heads are independently calculated.
Since the output of SAN is a linear transformation (using $\mW^O$)  of $V$, which is a weighted sum of $V$. A stack of many purely SAN layers is not expressive~\citep{dong2021attention}, since it is equivalent to a single linear transformation.  To this end, a  feed-forward network with non-linear activation is alternately used with each SAN layer,
\begin{equation}
    \textrm{FFN}(X) = \delta( X \mW^\textrm{in}) \mW^\textrm{out}.
    \label{equ:ffn}
\end{equation}

Since some neurons after the activation function (e.g., $\delta$ is ReLU or GELU~\citep{hendrycks2016gaussian}) become inactivated (zero), $d_\textrm{in}$ is usually bigger than  $d_\textrm{model}$ to avoid the low-rank bottleneck, typically,  $d_\textrm{in} = 4 \times  d_\textrm{model} = d_\textrm{out}$.
Other tricks, such as layer normalization, residual connection, dropout, and weight decay  are also adopted to relieve the optimization and overfitting problems when it goes deeper, resulting in better stability when training large neural networks.
It is generally believed that Transformer is better than LSTM in terms of generalization since its performance usually does not get to saturation  as early as LSTM. When models become large,  the performance of the Transformer is consistently increasing when feeding  more  data while LSTM gets saturation if a certain amount of data is fed.

% \benyou{stability to finetune larger neural networks, feature representation, domain-specific for long text}

}

% \textcolor{red}{Interestingly, AlphaFold2~\cite{jumper2021highly} also borrows some insights to design the so-called `Evoformer' as the core component in its architecture.}
\textcolor{black}{Interestingly, the computer vision  \cite{li2022lvit} and computational biology  communities  also borrow some insights to design their models, see ViT \cite{li2022lvit} for vision and AlphaFold2~\cite{jumper2021highly}  for protein.}
In Table \ref{tab:PLM}, we introduce some typical pre-trained language models in general NLP domains, based on these two backbone neural networks.

% \benyou{vocab, position embedding and transformer module  }

\subsection{\textcolor{black}{Pre-training for texts}}
\label{sec:text-pre-training}

Previously, there were many typical methods to build token representation (\textit{e.g.}, word vectors) from plain corpora. For example, \cite{pennington2014glove,mikolov2013efficient} build a one-to-one mapping between words and their vectors, which is called `static word embedding' since it is static and not related to word context. However, it is well known that words often express different meanings in different contexts.
To achieve this, most recently many pre-trained language models~\cite{peters2018deep} are proposed to learn `contextualized word embedding' that models the bi-directional contexts of words.
%Inspired by \cite{peters2018deep}, many pre-trained language models adopt `contextualized word embedding' to model words in a specific context.  
For `contextualized word embedding', the vector for a word depends on its specific usage in a context. For example, the meanings of `{\tt bank}' in `{\tt river bank}' and in `{\tt money bank}' are supposed to have some difference. 
% the `contextualized word embedding' may not only arguable help word meaning disambiguation, but also improve the quality of word representation in various tasks \cite{devlin2018bert}.
Compared with `static word embedding', the `contextualized word embedding' largely improves the quality of word representation in various tasks \cite{devlin2018bert}. 

{\color{black}
A \textit{language model} aims to assign a probability to a given piece of text (e.g., a sentence or an n-gram.) \cite{jurafsky2000speech}, see below:
\begin{equation}
    \Theta : \mathbb{V}^N \rightarrow \mathbb{R}^+
\end{equation}
While, in the scenario of natural language processing, a generally-called \textit{language model} is usually a  \textit{conditional language model} that assigns a probability to a next word $w_n$ given some conditioning context (denoted as $[w_1, \cdots, w_{n-1}]$). A conditional language model is a generalization of \textit{language model} in a sense the former could be obtained by dividing the  probability of the concatenated sentence (i.e., $[ w_1, \cdots, w_{n-1}, w_n ]$) by that of the context, namely 
\begin{equation}
    P( w_n \vert w_1, \cdots, w_{n-1}, w_n) = \frac{ \Theta  ( w_1, \cdots, w_{n-1})}{\Theta  ( w_1, \cdots, w_{n-1})}
\end{equation}

In the earliest, neural language models~\cite{bengio2003neural,Mikolov2013Distributed} and their variants such as Skip-Gram~\cite{mikolov2013efficient}, CBow~\cite{mikolov2013efficient} and Glove~\cite{pennington2014glove}, were the backbones of modern NLP to provide pre-trained word features.
The pre-training task of classical neural language models~\cite{bengio2003neural} 
is the unidirectional language modeling (ULM), that predicts the next word conditionally on history words.
To learn better word embeddings, several classical models further improved the pre-training task.
For example, the training objective of Skip-Gram~\cite{mikolov2013efficient} is predicting context words given the input word.
CBow~\cite{mikolov2013efficient} aims to predict the next word based on its bidirectional context words.
The training task of Glove~\cite{pennington2014glove} is to predict the log co-occurrence of words.
These models typically use shallow neural network architecture to conduct calculations between word vectors, for efficient training. 

Language models could be considered as an instance of self-supervision.
Compared to data-hungry supervised learning, which usually needs annotations from humans, language models could make use of massive amounts and cheap plain corpora from the internet, books, etc. 
% The pre-training paradigm largely benefits from self-supervised NLP tasks. 
In language models, a next word is a natural label for a context sentence as a next word prediction task, or one can artificially mask a known word and then predict it. 
The paradigm that uses the unstructured data itself to generate labels (for example, the next word or the masked word in language models) and train language models to predict labels thereof is called `self-supervision learning'.
Language model pre-training is therefore referred to as an `auxiliary task', in which the learned representations in language models can be used as an initial model for various downstream supervised tasks.
The pre-training objective/task is critical for learning efficient representations that are generalizable and universal for downstream tasks.

Recently, efforts have been proposed to learn contextualized word representations based on deep neural networks, such as the pioneer method ELMO~\cite{peters2018deep}, GPT~\cite{radford2018improving}, and the breakthrough work: BERT~\cite{devlin2018bert}.
Similar to traditional neural language models, GPT uses the unidirectional language model task as the pre-training objective.
ELMO proposed the pre-training task for bidirectional language modeling based on both the forward language model and backward language model task. 
The forward language model task aims to model the probability of the word given its previous words, while the backward language model task predicts the word based on its future words.
To better model bi-directional contexts during pre-training, BERT proposed the masked language model (MLM) pre-training objective with the inspiration of the Cloze task.
It randomly masks tokens of input sequences and aims to predict masked tokens with the masked text sequences.
Different from ELMO which concatenates the forward and backward language model, MLM can train the deep bidirectional contextual representations with only one language model.
Based on MLM, Encoder-Decoder language models such as T5~\cite{raffel2020exploring}, proposed the pre-training objective of generating the given sequences in an auto-regressive way taking the masked sequences as input.
The language models based on the auto-regressive pre-training objective are more suitable for the text generation tasks such as abstractive summarization and question answering.
The overview of pre-training tasks is shown in Table~\ref{tab:word_vector}. } 
\textcolor{black}{ Recently, Open AI have released many API services on their trained model, including GPT 3, InstuctGPT, Codex, and ChatGPT. Especially, ChatGPT could interact in a conversational and makes it possible to answer follow-up questions, admit mistakes, challenge incorrect premises, and reject inappropriate requests. 
%representation that can be easily adapted to downstream tasks. 
}

\textcolor{black}{
These pre-training tasks in language modeling are sometimes called `pretext tasks'. In conclusion, by pre-training multi-layer transforms in plain text using pretext tasks, it  learns general text representation that can easily be adapted to downstream tasks.
}
\begin{table}[t]
    \setlength{\tabcolsep}{1pt}
    \centering
    \scriptsize
    \caption{Typical ways for word vectors and language models. $X=\{a,b,c,d,e\}$ is an example text sequence. ELMO, BERT, and GPT usually work on much longer sequences than neural language models (NLMs), Skip-gram and CBOW.}
    % \addtolength\tabcolsep{-0.5pt}
    \begin{tabular}{llllp{7cm}}
    \toprule
    Model & Type & Architecture & Task & Loss function\\
    \midrule
    % \multirow{2}{*}{Methods} & \multicolumn{2}{c}{accuracy}   \\ & test     \\
    \multirow{2}{*}{NLM~\cite{bengio2003neural}}   &  \multirow{2}{*}{static} &  \multirow{2}{*}{1-layer MLP}   & $  (a,b)   \rightarrow c $\\
    % &   & $  p(w_i \vert w_1, w _k ) = w_i (\sum_{w_j \in \mathcal{C}_k(w_i)} w_j) ^T$\\
    &&& predicting the next word & $-\sum_{i=1}^T log p(x_i|\{x_1, ..., x_{i-1}\})$ \\
    \hline
   \multirow{2}{*}{Skip-Gram~\cite{mikolov2013efficient}}    &  \multirow{2}{*}{static} &  \multirow{2}{*}{1-layer MLP}   & $ b  \rightarrow c, \quad b  \rightarrow a  $\\
    % &   & $  p(w_i \vert w_1, w _k ) = w_i (\sum_{w_j \in \mathcal{C}_k(w_i)} w_j) ^T$\\
    &&& predicting neighboring words & $-\sum_{i=1}^T log p(\{x_{i-o},...,x_{i-1}, x_{i+1},..., x_{i+o}\}|x_i)$, ($o$ is the window size)\\
    \hline
    \multirow{2}{*}{CBow \cite{mikolov2013efficient}}   &     \multirow{2}{*}{static} &  \multirow{2}{*}{1-layer MLP}   & $ (a, c)  \rightarrow b  $\\
        &&& predicting central words & $-\sum_{i=1}^T log p(x_i|\{x_{i-o},...,x_{i-1}, x_{i+1},..., x_{i+o}\})$, ($o$ is the window size)\\
    \hline
        \multirow{2}{*}{Glove \cite{pennington2014glove}}  &\multirow{2}{*}{ static} &  \multirow{2}{*}{1-layer MLP}   &  $  \vec{w_i} ^T\vec{w_j}    \propto  log p( \#( w_i  w_j  )) $\\
        &&& predicting the log co-occurrence count & $-\sum_{i=1,j=1}^T f(x_{ij})(\vec{w_i} ^T \vec{w_j} + b_i + c_j - log x_{ij}), (x_{ij} = p( \#( w_i  w_j  )))$\\
    \hline
     \multirow{2}{*}{ELMO \cite{peters2018deep}}  & \multirow{2}{*}{contextualized} &  \multirow{2}{*}{LSTM}  & $ (a,b,c,d) \rightarrow e, \quad  (e,d,c,b) \rightarrow a $ &  \multirow{2}{*}{$-\sum_{i=1}^T log p(x_i|\{x_1, ..., x_{i-1}\})+log p(x_i|\{x_{i+1}, ..., x_{T}\})$}\\ 
         &&&bi-directional language model \\
    \hline
    
       BERT \cite{devlin2018bert}, Roberta \cite{liu2019roberta}  & \multirow{2}{*}{contextualized} &  {Transformers}  & $ (a,[\textrm{\tt mask}],c) \rightarrow  (\_,b,\_) $ & \multirow{2}{*}{$-\sum_{x \in mask(x)} log p(x|\hat{X})$,  $\hat{X}$ is the corrupted sentence with masks}\\ 
         ALBERT \cite{lan2019albert},XLNET \cite{yang2020xlnet}  && or Transformer-XL& predicting masked words \\
    \hline
        % \multirow{2}{*}{XLNET \cite{yang2020xlnet}} &  \multirow{2}{*}{contextualized}   &   \multirow{2}{*}{Transformer-XL}  & %  a generalized  autoregressive pre-training  using  bidirectional contexts  \\
       % &&&  generalized  autoregressive pre-training  \\
    %\hline
         \multirow{2}{*}{Electra \cite{clark2020electra}}  &  \multirow{2}{*}{contextualized}   &  \multirow{2}{*}{Transformer}  &  $ (a,\hat{b},c,\hat{d}) \rightarrow (0,1,0,1) $ &  \multirow{2}{*}{$-\sum_{i=1}^T log p(b_i| \hat{X} )$, $b_i$ indicates whether $x_i$ is replaced.}\\
        &&& replaced token prediction  \\

            \hline
T5~\cite{raffel2020exploring}  & \multirow{2}{*}{contextualized}  &  \multirow{2}{*}{Transformers}  & $ (a,b,c,) \rightarrow (d, e) $& \multirow{2}{*}{$-\sum_{i=1}^T log p(y_i|{X}, y_0, \cdots, y_{i-1})$, ${X}$ and $Y=\{y_1,\cdots,y_T\}$ are the  input/output}\\
{ BART~\cite{lewis2019bart}}         &&& predicting the sequence  \\
      % ChatCPT \\  %  $ (a,[mask],c,[mask],e) \rightarrow (a, b, c, d, e) 
      \hline
        \multirow{2}{*}{GPT \cite{radford2018improving}}  & \multirow{2}{*}{contextualized}  &  \multirow{2}{*}{Transformers}  & $ (a,b,c,d) \rightarrow e $ autoregressively& \multirow{2}{*}{$-\sum_{i=1}^T log p(x_i|\{x_1, ..., x_{i-1}\})$, $\{x_1, ..., x_{T}\}$ is the sequence}\\
         &&& predicting the next word  \\
    \bottomrule
    \end{tabular}
    \label{tab:word_vector}
\end{table}

\paragraph{Pre-training corpora}

{\color{black}Except for the superior pre-training objective, it usually requires a large scale of raw texts to pre-training language models effectively.
On the internet, unlabelled raw texts are abundant ranging from news texts, and web pages, to online encyclopedias.
The training corpora for pre-trained language models mainly include: 1) online encyclopedia like Wikipedia \footnote{\url{https://dumps.wikimedia.org/}}, which was widely used for training BERT and its variants. 
2) existing books and stories that have been digitized like BooksCorpus \cite{zhu2015aligning} and ,   
3) web texts extracted from online websites/URL, such as crawled online corpora\footnote{\url{https://commoncrawl.org/}}. 
PLMs trained by these corpora are usually able to capture the common sense knowledge inherited in the raw training texts.
For specific domains such as the biomedical domain, it, therefore, needs other efforts such as domain-specific pre-training with domain-specific texts, to capture the domain knowledge (will further be introduced in the next section). 
Moreover, the vocabulary with limited words is unable to cover all words in the large-scale training texts.
To address the out-of-vocabulary (OOV) problem, they proposed to split words into sub-words to formulate the vocabulary via the Byte-Pair Encoding (BPE)~\cite{sennrich2016neural} or WordPiece~\cite{kudo2018sentencepiece} methods.

}

% \begin{table} \footnotesize
%     \centering
%     \caption{Representative pre-trained language models in the general domain. NSP means the next sentence prediction task.}
%     \addtolength\tabcolsep{-4pt}
%     \begin{tabular}{llll}
%         \toprule
%         Model &  Objective & Backbone network  & Comments\\
%         \midrule
%         ELMO \cite{peters2018deep} & bidirectional LM & Bi-LSTM &  the first
%          contextualized word representation \\
%          BERT \cite{devlin2018bert} & masked LM, NSP & Transformer (Encoder) & the most commonly-used pre-trained language model \\
%          Roberta \cite{liu2019roberta} & masked LM & Transformer (Encoder) & a longer-trained BERT variant using more data \\
%          ALBERT \cite{lan2019albert} & masked LM, NSP &  Transformer (Encoder) & a BERT variant with shared weights and a factorized word embedding\\
%          XLNET \cite{yang2020xlnet} &    generalized  autoregressive pre-training   &  Transformer (Encoder) &   a generalized  autoregressive pre-training  using  bidirectional contexts  \\
%          Electra \cite{clark2020electra} &   replaced token prediction  & Transformer (Encoder) &  a pre-trained based LM trained by replaced token prediction\\
%          GPT \cite{radford2018improving} & autoregressive language model & Transformer (Decoder) & a pre-trained based LM for autoregressive generation \\
%          T5 \cite{raffel2020exploring} & Seq2Seq &  Transformer (En-Decoder) & a pre-trained based LM for seq2seq generation \\
%          \bottomrule
%     \end{tabular}
%     \label{tab:PLM}
% \end{table}
\begin{minipage}{\textwidth}
 \begin{minipage}[b]{0.55\textwidth}
    \centering
    \includegraphics[width=0.8\textwidth]{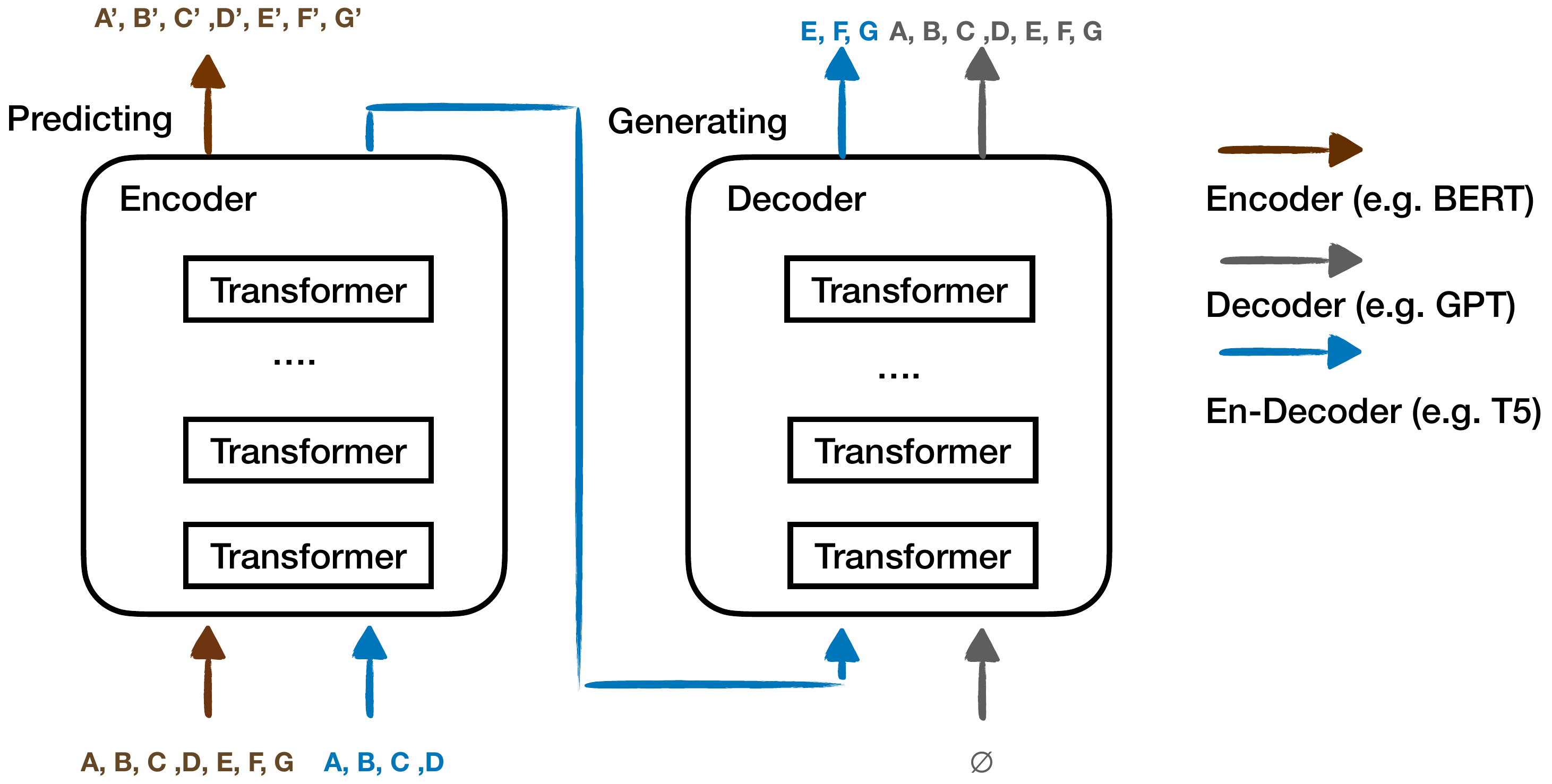}
    \captionof{figure}{The difference between Encoder, Decoder and En-Decoder pre-trained language models.
    %{ \{\tt A,B,C,D,E,F,G\}} is an example token sequence, while {\{\tt A',B',C',D',E',F',G'\}} is the corresponding abstract representation or labels. 
    }
    \label{fig:En-Decoder}
  \end{minipage}
  \hfill
\begin{minipage}[b]{0.44\textwidth}
    \centering
\small
    \centering
    \begin{tabular}{llll}
    \toprule
        Category &  Data &  Task \\
    \midrule
         Pre-training & general domain & pre-training task \\
         Domain adaption & target domain & pre-training task\\
         Task adaption & general domain & downstream task \\
         Fine-tuning &  target domain & downstream task\\
    \bottomrule
    \end{tabular}
     \captionof{table}{Categories to tailor pre-trained language models}
    \label{tab:categories}
\end{minipage}
  \end{minipage}
\paragraph{Representative PLMs}
 {
Pre-trained language models can generally be categorized into three principal types, based on whether the input or output constitutes a text sequence or label: Encoder-only, Decoder-only, and Encoder-Decoder models. Models such as BERT~\cite{devlin2018bert}, RoBERTa~\cite{liu2019roberta}, and ALBERT~\cite{lan2019albert} fall under the Encoder-only category and are primarily utilized for text classification and sequence labeling tasks. RoBERTa~\cite{liu2019roberta} is a BERT variation that has undergone a more extended training phase and employs additional data.  ALBERT~\cite{lan2019albert} serves as a lightweight BERT variant but features shared weights and a factorized word embedding. 
}

% Based on whether the input or output is a textual sequence or the label, pre-trained language models are mainly divided into three categories: \textit{Encoder-only}, \textit{Decoder-only}, and \textit{En-Decoder}. Encoder-only pre-trained models  {(such as BERT~\cite{devlin2018bert}, RoBERTa~\cite{liu2019roberta} and ALBERT~\cite{lan2019albert}) are mainly for text classification and sequential labeling tasks.  RoBERTa~\cite{liu2019roberta} is a longer-trained BERT variant using more data and ALBERT~\cite{lan2019albert} is a BERT variant with shared weights and a factorized word embedding. A special model XLNEt  is   a generalized   autoregressive pre-training  using  bidirectional contexts,  which performs masked word prediction  like BERT.}
 
Pre-trained models equipped with the decoder such as GTP series, T5, BART, could deal with generation-related tasks like translation, summarization, and language models~\footnote{ XLNet~\cite{yang2020xlnet} provides a generalization of autoregressive pre-training by leveraging bidirectional contexts to conduct masked word prediction akin to BERT. It could also deal with text generation.}. See Fig. \ref{fig:En-Decoder} for the difference: an Encoder model predicts labels for each input tokens (in brownish yellow); a Decoder model generates a sequence of tokens w.r.t. a probability distribution (in blue); an En-Decoder model predicts a new sequence conditioned on a given sequence (in grey), \textit{a.k.a.} Seq2Seq.

{\color{black}
\paragraph{Knowledge in PLMs} As a pioneer, LAMA \cite{petroni2019language} has explored the ability about how much PLMs could capture factual and commonsense knowledge  (in the format of triplets in knowledge bases). It concludes that large PLMs (e.g., BERT-Large) can recall  knowledge slightly better than small competitors and remarkably better than with non-neural and supervised alternatives \cite{petroni2019language}.
However, \cite{cao2021knowledgeable} revise the ability that PLMs can potentially be a reliable knowledge source. %how much PLMs could capture factual and commonsense knowledge.  
Cao et al \cite{cao2021knowledgeable}  claims that the way PLMs capture knowledge is vulnerable;  it might overfit dataset artifacts and make use of answer leakage. 
% It shows the three paradigms of such patterns to capture factual knowledge. 
In the biomedical domain, it needs more domain knowledge and it is therefore more knowledge-intensive than the general domain. Some existing work (e.g. \cite{jha2022continual}) has explored injecting biomedical domain knowledge in PLMs.

}

% \begin{itemize}
%     \item \textit{Encoder-only} pre-trained language models: for exmaple, BERT learns a encoder to encode text for classification and sequential labeling. 
%     \item  \textit{Decoder-only}
%     \item  \textit{Encoder-Decoder}.
% \end{itemize}

\subsection{\textcolor{black}{Pre-training for images}}
\label{sec:image-pre-training}
% \benyou{pre-trained CNN, contrastive pre-training, CLIP, Dall-E, Flamigo, etc.; do we also mention contrative pre-training in NLP?}

{\color{black}

Deep neural networks have achieved excellent performance in the imaging domain on various vision tasks, e.g., image classification, object detection, and instance segmentation. One of the major reasons behind this is pre-training. However, different from language models in the NLP field, `pre-training' in the earliest means training vision models on large annotated image datasets, e.g., ImageNet~\cite{deng2009imagenet}. Subsequently, different self-supervised learning approaches are proposed to overcome the shortcoming of supervised learning, e.g., generalization error and spurious correlations. Next, we detail different types of pre-training for images.

\paragraph{Supervised pre-training} 
In supervised pre-training, the most commonly-used dataset is ImageNet which contains over one million labeled images. Supervised pre-training~\cite{krizhevsky2017alexnet,he2016resnet} involves training a deep learning model on the entire ImageNet dataset to learn generic features that can be useful for various downstream tasks. Once the model has been pre-trained on the large dataset, it can be fine-tuned on a smaller, task-specific dataset relevant to the specific task. This can help the model learn valuable features that can be generalized to different tasks at hand.

\paragraph{Contrastive self-supervised Learning}
Different from supervised pre-training, contrastive self-supervised learning~\cite{chen2020simclr,he2020moco,grill2020byol} is a method for representation learning without needing labeled data. It involves training a model to distinguish between different variations of a given input image. For example, the model might be trained to identify whether two images are a rotated version of the same image or whether they are two completely different images. By learning to predict these labels, the model can learn useful features that can be applied to various tasks, such as object detection and semantic segmentation.

\paragraph{Masked self-supervised Learning} 
Motivated by BERT in NLP, masked self-supervised learning has attracted attention in the computer vision field~\cite{bao2021beit,he2022mae,xie2022simmim}. It is a type of generative pre-training approach. Models are trained to reconstruct images from incomplete data, in which part of the input image is removed or masked before it is fed into the model. This allows the model to learn the underlying structure of the image.

\paragraph{Contrastive language-image pre-training} 
Contrastive language-image pre-training \cite{radford2021clip} (CLIP) aims to train a vision model on a wide variety of image-text datasets. The model is trained to pair images and texts in a mini-batch through contrastive learning. CLIP showed excellent zero-shot transfer ability, where the pre-trained model can achieve comparable results with the original ResNet \cite{he2016resnet} on ImageNet in a zero-shot manner. One of the primary reasons is that texts provide rich, detailed information about the visual content of an image. For example, a text description of an image can include information about the objects and scenes depicted in the image, as well as their spatial relationships and attributes. This information can help a machine learning model to identify and understand an image's visual content. Additionally, texts can be easily generated and collected in large quantities, making them a convenient and scalable source of supervision for visual representation learning.

}

%In this paper, we mainly discuss the pre-trained language models, \textit{e.g.}, BERT and GPT. The main difference between them is that BERT is a textual encoder to encode a given document, while GPT is a textual decoder to decode a new document. This can also be considered as the difference between the discriminative model and generative model in machine learning. BERT is mainly used for the discriminative prediction/inference for a given text, like information extraction, text classification, named entity recognition, relation extraction, and non-generative question answering, as shown in Sec. \ref{sec:finetuning}. The latter is to generate texts, for example, text summarization, text completions, generative question answering and translations.

\subsection{Fine-tuning Paradigm in PLMs}
\label{sec:paradiam}
One challenge to use PLMs in downstream tasks is that there are two gaps between PLMs and downstream tasks, the \textit{task gap} and \textit{domain gap}. The \textit{task gap} means the meta-task in PLMs (usually masked language model in BERT or causal language model in GPT) usually can not directly be tailored to most downstream tasks (\textit{e.g.} sentimental classification). The \textit{domain gap} refers to the difference between the trained corpora in PLMs and the needed domain in a specific downstream task. The adaptation of both \textit{task gap} and \textit{domain gap} is crucial.% \begin{figure}
%     \centering
%     \includegraphics[width=0.8\textwidth]{figures/bert.png}
%     \caption{BERT architecture and its pipeline for pre-training and fine-tuning procedures~\cite{devlin2018bert}. \benyou{revised figure} }
%     \label{fig:bert}
% \end{figure}

\paragraph{Adaption}
To use the pre-trained language model in a downstream task, it is suggested to adopt both the domain and task adaption~\cite{rongali2021continual,zhang2020multistage,DBLP:conf/acl/GururanganMSLBD20,gu2020domain}, see Table. \ref{tab:categories} for the difference. The domain adaption suggests continuing training pre-trained models trained from a general domain, in the target domain, \textit{e.g.}, biomedical domain. Task adaption refers to fine-tuning on similar downstream tasks. In this paper, without specifying, we mainly discuss the domain-adapted pre-trained models in various downstream tasks. Task adaption is not the main concern in this review. Take BERT as an example, BERT is first trained using next-sentence predictions (NSP) and masked language models in the pre-training phase. Such pre-trained BERT will be used as the initial feature extractor. BERT with an additional classifier layer is then fine-tuned to optimize the objective of down-stream tasks (like MNLI \cite{N18-1101}, NER \cite{tjong-kim-sang-de-meulder-2003-introduction}, and SQuAD \cite{rajpurkar2016squad}).

\section{PLMs in Biomedical Domain}
%\section{PLMs in Biomedical Domain: Why and How?}
\label{sec:pertraining}
Recently, the pre-trained language models have been widely applied to various NLP tasks and achieved significant improvement in performance, because: 1) Pre-training on the huge text corpus can learn universal language representations and help with the downstream
tasks.
2) Pre-training provides a better model initialization,
which usually leads to a better generalization performance and speeds up convergence on the target task.
3) Pre-training can be regarded as a kind of regularization
to avoid overfitting on small data \cite{qiu2020pre}. 
Self-supervised learning, which pre-trained language models rely on, usually adopts plain unstructured corpora in a format of a sequence of tokens.
\textcolor{black}{At first, most pre-trained language models focus on pre-training in general plain corpora from the Internet, like Wikipedia or crawled webpages. 
Except for the general domain, efforts have been proposed to extend PLMs in specific domains such as:~\cite{feng2020codebert} trains CodeBERT in the programming language and \cite{BeltagyLC19} trains SciBERT on scientific publications and biological sequence.} This paper aims to discuss pre-trained language models in the biomedical domain.
It is believed that the pre-trained language model can always benefit from more training corpora~\cite{gu2020domain}. To achieve better performance in the domain-specific downstream tasks, it is also intuitive that the in-domain data pre-training is necessary. 

We will first introduce the motivation of using pre-trained language models in the biomedical domain in the Sec.~\ref{sec:motivation}.
Then, we will illustrate the main components on tailoring PLMs to the biomedical domain including the in-domain data in the Sec.~\ref{sec:data}, and the pre-training and fine-tuning strategy in 
the Sec.~\ref{sec:tailor}.
%In the biomedical domain, the in-domain data can be text in the electronic health records, scientific literature, and online social media, or biological sequence (\textit{e.g.}, DNA pieces), which will be introduced in the Sec.~\ref{sec:data}. 
Next, in the Sec.~\ref{sec:model}, we will introduce existing pre-trained models in the biomedical domain, which are pre-trained from the in-domain data as introduced in the Sec.~\ref{sec:data}. We will give an overview of these models, \textcolor{black}{catagorization of them, and discussion} differences between them.  We expect to help one from both the bioinformatics and computer science communities to get knowledge of the biomedical domain-specific pre-trained language model quickly.

\subsection{Motivation}
\label{sec:motivation}
In the biomedical domain, the motivation for using pre-trained language models is manyfold. 
\begin{itemize}
    \item Firstly, the biomedical domain involves biomedical data in the format of sequential tokens (like biomedical texts and the history of electronic health records) that usually lack annotations. However, these sequential data were previously thought of as difficult to model. Thanks to pre-trained language models, it has been empirically demonstrated to train these sequential data in a self-supervised manner effectively. This would open a new door for \textcolor{black}{processing biomedical data with} pre-trained language models. 
    \item Second, annotated data in the biomedical domain is usually limited at scale. Some extreme cases in machine learning are called `zero-shot' or `few-shot'. More recently, language models such as GPT3 show that language models have the potential for few-shot learning and even zero-shot learning \cite{brown2020language}. Therefore, a well-trained pre-trained language model \textcolor{black}{in the biomedical domain} is more crucial to provide a richer feature extractor, which may slightly reduce the dependence on annotated data. 
    \item Plus, the biomedical domain is more knowledge-intensive than the general domain, since most tasks may need domain expert knowledge, while pre-trained language models could serve as an easily-used soft knowledge base \cite{petroni2019language} that
captures implicit knowledge from large-scale plain \textcolor{black}{biomedical corpora} without human annotations. More recently, GPT3 has been shown to have the potential to `remember' many complicated common knowledge  \cite{brown2020language}. 
\item Lastly, beyond text, there exist various types of biological sequential data in the biomedical domain, like protein and DNA sequences. Using these data to train language models has shown great success in biological tasks like protein structure predictions. Therefore, it is expected that pre-trained language models could solve more challenging problems in biology.
\end{itemize}
% \begin{itemize}
%     \item Biomedical domains involve many sequential tokens like (text in electronic health record and protein), which  naturally fit pre-trained language models.
%     \item In the biomedical domain,  supervised tasks usually  have limited training examples, thus it heavily relies on pre-trained language models for better feature engineering.
%     \item Biomedical domain is knowledge-intensive since expert  knowledge is essential. Pre-trained language models can be used as the soft knowledge base \cite{petroni2019language}.
%     \item proteins and DNA

\subsection{Biomedical Data for Pre-training}
\label{sec:data}

\begin{table}[t]
\small
    \centering
     \addtolength\tabcolsep{-1pt}
     \resizebox{.95\textwidth}{!}{
    \begin{tabular}{lllllllll}
         \toprule
            dataset  &  types & size & characteristics \\
         \midrule
            MIMIC III   &  EHR  & 58,976  hospital admissions for 38,597 patients   & from  Beth Israel Deaconess Medical Center in 2001-2012                                                      \\
            CPRD        &  EHR  & 11.3M patients &     anonymized medical records  from  674 UK GP practices            \\
            BREATHE     &  Scientific Publications  & 6M articles and about 4 billion words&     sources are diverse.     \\
            PubMed      &  Scientific Publications &  35M citations and abstracts of biomedical literature& It provide only links to  journal articles         \\
            COMETA in Reddit      & Social Media  &  800K Reddit posts &  68 health-themed subreddits with  entity annotation        \\
            Tweets      & Social Media & up-to-date    & one could crawl real-time Tweets using its official API       \\
            UMLS        &  Knowledge Bases  &  2M names for 900K concepts   & well-organized medical knowledge source  \\
            IU-Xray     & image-Text Pairs  &  3,955 reports and 7,470 images &    XML reports with  findings, indications, comparisons,  etc.     \\
            MIMIC-CXR   & image-Text Pairs  & 77,110 images  & images corresponding to 227,835 radiographic studies       \\
            ROCO        & image-Text Pairs   & 81,000 radiology images and  corresponding captions  &  figures and their corresponding captions in PubMed articles    \\ 
            MedICaT     &  image-Text Pairs  &     17,000 images includes captions  & open-access biomedical papers and their captions  \\
         \bottomrule
         & 
    \end{tabular}}
    \caption{ \color{black} Summary of Biomedical Data for pre-training. }
    \label{tab:Summary_dataset_general}
\end{table}

Unstructured plain data for pre-trained language models mainly include electronic health records, scientific publications, social media text, biomedical image-text pairs, and other biological sequences like protein, { \color{black} see Tab.~\ref{tab:Summary_dataset_general}}.  An overview of EHR mining can be seen in \cite{yadav2018mining,10.1145/1118890.1118891}, and \cite{gonzalez2017capturing} discussed both health records and social media text. One can also check \cite{kalyan2020secnlp} for some systematic overview of biomedical textual corpora.

% \begin{table}[]
%     \centering
%     \begin{tabular}{c|c}
%          MMic-iii &  \\
%          pubmed & \\
%          arxiv bio \\
%          ....
%     \end{tabular}
%     \caption{ \color{black}  biomedical data }
%     \label{tab:my_label}
% \end{table}

% \benyou{A table for dataset, indicates the size, type and characteristics.}

\subsubsection{Electronic Health Record} Electronic health record (EHR) is a collection of patient and population electronically-stored health information in a digital format that may include demographics, medical history, medication and allergies, immunization status, laboratory test results, radiology images, vital signs, personal statistics like age and weight, and billing information. One can check \cite{solares2020deep,weng2019representation} for details about EHR with deep learning.
Assessing  such records  may be  restricted to limited organizations, which hinders its widespread to the public. The reason may involve some privacy issues.

\paragraph{MIMIC III}
Medical Information Mart for Intensive Care III  dataset~\cite{johnson2016mimic} \footnote{\url{https://mimic.mit.edu/}} is one of the most popular EHR datasets, which consists of 58,976 unique hospital admissions from 38,597 patients in the intensive care unit of the Beth Israel Deaconess Medical Center between 2001 and 2012. In addition, there are 2,083,180 de-identified notes associated with the admissions.

\paragraph{CPRD} Clinical Practice Research Datalink (CPRD)~\cite{herrett2015data} is the primary care database of anonymized medical records  from  674 general physicians (GP) practices in the UK,  which involves over 11.3 million patients. It consists of data on demographics, symptoms, tests, diagnoses, therapies, and health-related behaviors. It is also linked to secondary care (\textit{i.e.}, hospital episode statistics, or HES) and other health and administrative databases (\textit{e.g.}, office for national statistics’ death registration).  
With 4.4 million active (alive, currently registered) patients meeting quality criteria, approximately 6.9\% of the UK population are included, this shows that patients are broadly representative of the UK general population in terms of age, sex, and ethnicity. As a result, CPRD has been widely used across countries and spawned a lot of scientific research output. 
% \paragraph{\textbf{Cerner}} is a de-identified EHR database that consists of over 600 hospitals and clinics in the United States. It represents over 68 million unique patients and includes longitudinal data from 2000 to 2017
% used  in Med-BERT

% https://pubmed.ncbi.nlm.nih.gov/26481140/
% radiology reports
\subsubsection{Scientific Publications}
Scientific publications are another source for biomedical pre-trained language models since we expect that  biomedical knowledge may be encapsulated in scientific publications. 
% Moreover, such knowledge may not be limited to traditional common knowledge, but also involves some state-of-art research output that may be discovered by recent literature.

\paragraph{BREATHE}
Biomedical Research Extensive Archive To Help Everyone (BREATHE) \footnote{\url{https://cloud.google.com/blog/products/ai-machine-learning/google-ai-community-used-cloud-to-help-biomedical-researchers}}, is a large and diverse dataset collection of biomedical research articles from leading medical archives. It contains titles, abstracts, and full-body texts. 
The dataset collection process was done with public APIs that were used when available. The primary advantage of the BREATHE dataset is its source diversity. BREATHE is from nine sources including BMJ, arXiv, medRxiv, bioRxiv, CORD-19, Springer Nature, NCBI, JAMA, and BioASQ \cite{chakraborty2020biomedbert}.
BREATHE v1.0 contains more than 6M articles and about 4 billion words. BREATHE v2.0 is the most recent version. 

\paragraph{PubMed}
PubMed \footnote{\url{https://pubmed.ncbi.nlm.nih.gov/}} is a free search engine accessing the MEDLINE database of references and abstracts on life sciences and biomedical topics primarily. 
% The United States National Library of Medicine (NLM) at the National Institutes of Health maintains the database as part of the Entrez system of information retrieval.
PubMed comprises more than 32 million citations for biomedical literature from MEDLINE, life science journals, and online books. Citations may include links to full-text content from PubMed Central and publisher websites.
PubMed abstracts (PubMed) have 4.5B words, and PubMed Central full-text articles (PMC) have 13.5B words.

\subsubsection{ Social Media} 
Users post information on social media, which may contain biomedical information. We mainly introduce Reddit and Tweets as examples.

\paragraph{Reddit}
Reddit is an American social news aggregation, web content rating, and discussion website. Registered members submit content to the site, such as links, text posts, images, and videos, then voted up or down by other members. Posts are organized by subject into user-created boards called "communities" or "subreddits", which cover a variety of topics such as news, politics, religion, science, movies, video games, music, books, sports, fitness, cooking, pets, and image-sharing. Submissions with more up-votes appear towards the top of their subreddit and, if they receive enough up-votes, ultimately on the site's front page. Despite strict rules prohibiting harassment, Reddit's administrators have to moderate the communities and, on occasion, close them.
COMETA corpus~\cite{basaldella2020cometa} crawled health-themed forums on Reddit using Pushshift (Baumgartner et al., 2020) and Reddit’s own APIs. 
% Wechoose forums satisfying strict constraints, i.e. selecting subreddits where: (i) new content was posted daily, (ii) the quality of the content was sufficient (e.g. avoiding spam-ridden forums), (iii) the focus was the personal experiences or questions of the users.3 Applying these criteria, we
% They selected a list of 68 subreddits and crawled all the threads from 2015 to 2018, obtaining a collection of more than 800K discussions. This collection was then pruned by removing deleted posts, comments by bots or moderators, and so on.

\paragraph{Tweets}
Twitter is an American micro-blogging and social networking service on which users post and interact with messages known as "tweets". Registered users can post, like, and retweet tweets.  Tweets were originally restricted to 140 characters, but the limit was doubled to 280 for non-CJK languages in November 2017. Audio and video tweets remain limited to 140 seconds for most accounts. 
The COVID-twitter-BERT \cite{abs-2005-07503} is trained on a corpus of 160M tweets about the coronavirus collected through the Crowdbreaks platform \cite{muller2019crowdbreaks} during the period from January 12 to April 16, 2020.
% Crowdbreaks uses the Twitter filter stream API to listen to a set of COVID-19-related keywords in the English language. Prior to training, the original corpus was cleaned for retweet tags. Each tweet was pseudonymised by replacing all Twitter usernames with a common text token. A similar procedure was performed on all URLs to web pages. We also replaced all unicode emoticons with textual ASCII representations (e.g. :smile: for ,) using the Python emoji library. In the end, all retweets, duplicates and close duplicates were removed from the dataset, resulting in a final corpus of 22.5M tweets that comprise a total of 0.6B words. The domain-specific pretraining dataset therefore consists of 1/7th the size of what is used for training the main base model. Tweets were treated as individual documents and segmented into sentences using the spaCy library [8].
\subsubsection{ Online Medical Knowledge Sources}  Other than unstructured text, there is some online medical knowledge source that is well-organized. For example, UMLS provides biomedical concepts that may benefit biomedical pre-trained language models.
\paragraph{UMLS}
Unified Medical Language System (UMLS)~\cite{bodenreider2004unified}
(http://
umlsks.nlm.nih.gov) is a repository of biomedical
vocabularies developed by the US National Library
of Medicine. The UMLS has over 2 million
names for  900, 000 concepts from more
than 60 families of biomedical vocabularies, as well
as 12 million relations among these concepts. These vocabularies include the NCBI taxonomy, the Medical Subject Headings (MeSH), Gene Ontology, OMIM, and the
Digital Anatomist Symbolic Knowledge Base. 
% UMLS concepts are  inter-related  and also linked to external resources like GenBank. In addition to data, the UMLS includes tools for customizing the Metathesaurus (MetamorphoSys), for generating lexical variants of concept names (lvg) and for extracting UMLS concepts from text (MetaMap). 
The UMLS knowledge sources are
updated every quarter. In addition, all vocabularies are freely available for research purposes within an institution if a license agreement is signed.
% The UMLS knowledge sources are distributed on CD-ROM and by FTP.  % used in https://arXiv.org/pdf/2011.02947.pdf  

\subsubsection{Biomedical Image-Text Pairs}
{\color{black}

Besides texts, there are many medical texts paired with their corresponding images. This type of data is a good resource for learning the cross or joint representations of medical images and texts.

\paragraph{IU-Xray}
IU-Xray \cite{demner2016iuxray} has a collection of chest X-Ray images from the Indiana University hospital network. The data includes two files: one for the images and the other for the XML reports of the radiography. Each report may have multiple images, typically having two views: frontal and lateral. The XML reports contain information such as findings, indications, comparisons, and impressions. In total, there are 3,955 reports and 7,470 images.

\paragraph{MIMIC-CXR}
Medical Information Mart for Intensive Care Chest X-Ray \cite{johnson2019mimiccxr} is a large publicly available dataset of chest radiographs with free-text radiology reports. It contains 377,110 images corresponding to 227,835 radiographic studies performed at the Beth Israel Deaconess Medical Center in Boston, MA.

\paragraph{ROCO}
Radiology Objects in COntext \cite{pelka2018roco} is a large-scale medical and multimodal imaging dataset from the articles of PubMed Central, an open-access biomedical literature database. They are figures and their corresponding captions in articles. It has over 81,000 radiology images (from various imaging modalities) and their corresponding captions.

\paragraph{MedICaT}
MedICaT \cite{subramanian2020medicat} is also a dataset of medical figure-caption pairs also extracted from PubMed Central. Different from ROCO, 74\% of its figures are compound figures, including several sub-figures. It contains more than 217,000 images from 131,000 open-access biomedical papers and includes captions, inline references, and manually annotated sub-figures and sub-captions.

}

% \benyou{@zhihong images only or, image-text pair}

\subsubsection{Biological Sequences}
Other than text, there are various types of biomedical token sequences, e.g., amino acids for proteins.
The structure of each protein is fully determined by a sequence of amino acids \cite{anfinsen1973principles}. These amino acids are from a limited-size amino acid vocabulary, of which 20 are commonly observed. This is similar to  text that is composed of words in a lexicon vocabulary. In this subsection, we introduce a protein dataset called `Pfam' and a DNA sequence dataset from Human Genome Project.

% \paragraph{proteins}
\paragraph{Pfam Protein Dataset} The Pfam database \footnote{\url{http://pfam.xfam.org/}} is a large collection of protein families, in which each protein is represented by multiple sequence alignments using hidden Markov models. The newest version is Pfam 34.0, which was released in March 2021 and contains 19,179 families (or called `entries') and 645 clans \footnote{ Clans are the generated higher-level groupings of related entries in Pfam. A clan is a collection of entries that are related by sequence similarity, structure, or profile-HMM.}. 
The original purpose of the Pfam database is for the classification of protein families and domains. It creates the database using a semi-automated method of curating information on known protein families.
Pfam 34.0 contains 47 million sequences, which could be used to train protein language models.

% to improve the efficiency of annotating genomes.

% The Pfam classification of protein families has been widely adopted by biologists because of its wide coverage of proteins and sensible naming conventions

% Of the sequences that are in reference proteomes, 74.5\% have at least one Pfam match, and 48.8\% of all residues fall within a Pfam family.

\paragraph{DNA Dataset}
The DNA sequence is composed of a genomic sequence.
% primarily finished clones that were sequenced as part of the Human Genome Project.
The Human Genome Project was the international research effort to determine the DNA sequence of the entire human genome. Human Genome Project Results. In 2003, an accurate and complete human genome sequence was finished two years ahead of schedule and at a cost less than the original estimated budget. \cite{Ji2020} uses  the reference human genome GRCh38.p13 primary assembly from GENCODE Release \footnote{\url{https://www.ncbi.nlm.nih.gov/assembly/GCF_000001405.39/}}. The total  sequence length is about 3 Billion.
% \end{itemize}
% \section{Taxonomy of Biomedical PTMs}
\subsection{How to tailor PLMs to the Biomedical Domain}
\label{sec:tailor}
The pre-trained language model~\cite{devlin2018bert} is a new two-stage paradigm for NLP. In the first phase, it trains a language model (\textit{e.g.}, masked language model and casual language model) with a self-supervised meta-task in task-agnostic corpora. In the second phase, it fine-tunes the pre-trained language model to a (usually small-scaled) specific downstream task. 
\textcolor{black}{To tailor pre-trained language models on the biomedical domain, methods~\cite{gu2020domain,LeeYKKKSK20,abs-1904-05342} have explored conducting the domain-specific adaptation on both the pre-training and fine-tuning stage.
In the pre-training stage, the domain-specific adaption of existing efforts involves in the continual pre-training or training from scratch with a large scale of raw biomedical data.
This yield many efficient foundation models in the biomedical domain such as BioBERT~\cite{LeeYKKKSK20} and PubMedBERT~\cite{gu2020domain} et al, that can be directly used for downstream domain-specific tasks in the fine-tuning stage.
%In the fine-tuning stage, most efforts fine-tunes the pre-trained language models in the biomedical domain with the specific task, which is similar to the general domain.
}
%The trivial way to use a pre-trained language model on the biomedical domain is to fine-tune it with the domain data. 
%However, additional adaption is usually adopted to transfer the learned domain knowledge and task characteristics to the target domain and task. 
%In this paper, we group the usage of the pre-trained language model into four categories in Fig. \ref{fig:taxonomy}.
%The adaption is basically two-fold: transfer the domain or task characteristics. The former refers to  transferring a general pre-trained language model to the biomedical domain.
% Tab. \ref{tab:categories}
%\begin{figure}
%     \centering
%     \includegraphics[height=4cm]{figures/catogories.PNG}
%     \caption{Categories of usage of pre-trained models}
%     \label{fig:categories}
% \end{figure}

\subsubsection{\textcolor{black}{Biomedical Language Model Pre-training}}
One challenge in the biomedical domain is that medical jargon and abbreviations consist of many terms that are composed of Latin or Greek parts.  
\textcolor{black}{Moreover, clinical notes have different syntax and grammar from books or encyclopedias.
These lead to the semantic and domain-knowledge gap between the general pre-trained language models and the biomedical domain.
Therefore, many existing approaches have investigated the biomedical language models pre-training on the basis of pre-trained language models in the general domain, to tailor pre-trained language models to the biomedical domain.}
%This needs to design different vocabulary and therefore existing general pre-trained models with different vocabularies probably cannot be directly used;  training from scratch is sometimes necessary.

%something else.
\paragraph{\textcolor{black}{Continual pre-training}}
\textcolor{black}{The general way used by many methods~\cite{LeeYKKKSK20,peng2019transfer,abs-1904-05342} is to conduct the continual pre-training based on the general pre-trained language models such as BERT.
They directly initialize the model with existing general PLMs and further pre-training it with the self-supervised task and domain-specific corpora such as PubMed texts and MIMIC-III et al.
The representative works include the BioBERT~\cite{LeeYKKKSK20} that conducts continual pre-training based on the BERT with the PubMed abstracts and PubMed Central full-text articles, BlueBERT~\cite{peng2019transfer} that uses PubMed texts and MIMIC-III, Clinical BERT~\cite{abs-1904-05342} that further pre-trains BERT with clinical notes.
In this case, they use the same vocabulary as the general PLMs, which cover words in a corpus of the general domain such as Wikipedia and BookCorpus.
However, as mentioned before, biomedical texts consist of many domain-specific terms.
Using the same vocabulary as the general PLMs can be ineffective for modeling biomedical texts~\cite{gu2020domain}.
}
%One may reuse a pre-trained language model from the general domain (\textit{e.g.}, general Wikipedia pages \footnote{\url{https://en.wikipedia.org/}} or Google books) and then continue pre-training a few epochs in the new (target) domain (\textit{i.e.}, biomedical domain). In the case when the corpora in the target domain are large-scale enough, one can also directly train the model from scratch since there is no need to reuse the general knowledge.

\paragraph{\textcolor{black}{Pre-training from scratch}}
\textcolor{black}{To conduct better pre-training for biomedical language models, some efforts~\cite{gu2020domain,BeltagyLC19} have explored the way of pre-training from scratch.
Different from the continual pre-training, they propose to build the new vocabulary from the raw biomedical training corpora. 
SciBERT~\cite{BeltagyLC19} is the representative work, that constructs the new vocabulary with the size of 30K and trains the model with the mix-domain corpora, where 18\% training texts from the computer science domain, and 82\% from the biomedical domain.
However, one recent work~\cite{gu2020domain} has argued that the mixed domain pre-training doesn't make sense for the biomedical domain, since the target data of downstream applications in the biomedical domain is highly domain-specific.
Instead, they proposed the superior domain-specific pre-training from scratch that uses the training corpora from only the biomedical domain.
}

% \benyou{Regarding biomedical pretraining itself, a key insight emerging from recent work is the importance of domain-specific vocabulary and pretraining as proposed in the "Domain-Specific Language Model Pretraining for Biomedical Natural Language Processing" paper (thecitation for the paper should be Special Issue on Computational Methods for Biomedical Natural Language Processing, ACMTransactions on Computing for Health, 2021.). While the authors recited insights from this important paper (e.g., Continue Training orfrom scratch, line 451; Reusing existing vocabulary or building a new one, line 724), there is no mention or citation to make the properattribution. In Sec 4.2.1, in listing existing biomed PLMs, the authors also missed this important work and failed to point out the keydifferentiation that most models there were just straightforward continued pretraining of standard transformer models.}

\textcolor{black}{
\paragraph{Summary} Our observation is that the core factors that affect the decision between training from scratch or continuously  training are twofold: the scale of pre-training biomedical corpora and the domain specificity for  biomedicine, where we need to make a trade-off. Pre-training is in general data-hungry, one could fully leverage a large amount of biomedical corpora without inheriting parameters from a well-trained general PLM if  there already exist enough biomedical corpora. Early work (e.g., \cite{yu2019biobert}) tends to  continuously  train biomedical PLMs from an initial BERT. Nowadays, it becomes more popular to directly train biomedical PLMs from scratch thanks to the large scale of collected data and adequate computing resources \cite{luo2022biogpt}. Interestingly, \cite{singhal2022large} reused and tailored a giant general PLM (PaLM) to a clinical one, since giant models are economically expensive. We might expect some approach to decompose existing models and reuse part of them; afterward one can inject biomedical modules into it. 
}

\subsubsection{Fine-tuning}
Based on well-trained biomedical language models, one has to adapt them to downstream tasks. This is typically implemented to replace the mask language model prediction head and next sentence prediction head with a downstream prediction head, \textit{e.g.}, classification head, or sequence labeling heads.

Since the downstream tasks usually have much less training data than those used in pre-training, fine-tuning is an unstable process. 
Sun et al \cite{Sun2019} investigate different fine-tuning methods of BERT on the natural text classification tasks. 
Mosbach et al \cite{Mosbach2020} argues that the fine-tuning instability is due to vanishing gradients. 
Merchant et al \cite{Merchant2020} observe that fine-tuning mainly modifies the top layers of BERT. 
Unfortunately, the solutions (\textit{e.g.} hyper-parameters of which layer to fine-tune) proposed in those papers cannot be easily translated to other settings. 
To automate this process, automatic hyper-parameter tuning (\textit{e.g.} Bayesian optimization \cite{Brochu2010,Turner2021}) can come into help.
{ \color{black}
Tinn et al
\cite{tinn2021fine} systematically study fine-tuning stability in biomedical NLP. Particularly,  it finds that freezing lower layers is beneficial for small models, while layerwise decay is  beneficial for larger models.  In most cases, it facilitates  robust fine-tuning by using domain specific vocabulary and pre-training.}

\subsection{Biomedical Pre-trained Language Models}
\label{sec:model}
Based on the types of training corpora in the biomedical domain as introduced in the above section \ref{sec:data}, we mainly introduce two groups of biomedical pre-trained language models: biomedical textual language models and protein language models.
\textcolor{black}{Based on the types of training corpora in the biomedical domain as introduced in the section \ref{sec:data}, we mainly introduce biomedical pre-trained language models in three scenarios: pure language models, vision-and-language modeling, and protein/DNA language models.}

% \zhihong{The above paragraph could be placed in a better place?}

\subsubsection{Overview of Existing Biomedical Textual Language Models}
Since BERT was released, various biomedical pre-trained language models have been proposed via continued training with in-domain corpora based on the BERT model or training from scratch.
Tab. \ref{tab:models_overview} presents existing pre-trained language models with used corpora, size, release date, and related web pages.
% \Rotatebox{90}{%
\begin{table} \small
    \centering
    \setlength{\tabcolsep}{2pt}
    \scriptsize
    \caption{Existing textual biomedical pre-trained models. The base setting is with 0.1B parameters, and the large setting is with 0.3B parameters. The date is based on the submission in arXiv or published date of the journal or conference proceeding. }
    \Rotatebox{0}{
    \begin{threeparttable}
    \begin{tabular}{lllllp{7cm}}
    \toprule
    Model & Corpora & {\color{black} Architecture} & Size & Date  & Link \\
    \midrule
    BioBERT \cite{LeeYKKKSK20} &  PubMed  and PMC   & BERT &  base \& large & 2019.01 & \url{https://github.com/dmis-lab/biobert}  \\
    BERT-MIMIC \cite{si2019enhancing} & MIMIC III & BERT & base and large & 2019.02 & - \\
    SciBERT \cite{BeltagyLC19} & Semantic Scholar papers &BERT & base & 2019.03 & \url{https://github.com/allenai/SciBERT} \\
    BioELMo~\cite{jin2019probing} &  PubMed abstracts  & ELMo & 93.6 M  & 2019.04 & \url{https://github.com/Andy-jqa/bioelmo} \\
     Clinical BERT \cite{abs-1904-03323} & EHR (MIMIC-III) & BERT & base & 2019.04 & \url{https://github.com/EmilyAlsentzer/clinicalBERT}  \\ % \footnote{all notes or only discharge summaries in}
    Clinical BERT \cite{abs-1904-05342} & EHR (MIMIC-III) & BERT & base & 2019.05 & \url{https://github.com/kexinhuang12345/clinicalBERT}  \\
    BlueBERT \cite{peng2019transfer} &   PubMed+MIMIC-III & BERT & base \& large & 2019.05 &  \url{https://github.com/ncbi-nlp/bluebert} \\
     G-BERT \cite{shang2019pre} &  MIMIC III  &  BERT & - & 2019.06 &  \url{https://github.com/jshang123/G-Bert} \\
    BEHRT \cite{li2020behrt} & Clinical Practice Research Datalink & BERT &  - & 2019.07 & \url{https://github.com/deepmedicine/BEHRT} \\
     BioFLAIR \cite{sharma2019bioflair} & PubMed abstracts & BERT & lagre &  2019.08 & \url{https://github.com/zalandoresearch/flair} \\
     RadBERT \cite{meng2019selfsupervised} &  RadCore radiology reports& BERT & - & 2019.12 & - \\
     EhrBERT \cite{li2019fine} &  MADE corpus & BERT &  base &2019.12 & \url{https://github.com/umassbento/ehrbert}  \\ %20%20s%20(MADE)%20corpus
    Clinical XLNet \cite{clinicalxlnet} &  EHR (MIMIC-III)  & XLNET&   base & 2019.12& \url{https://github.com/lindvalllab/clinicalXLNet} \\
    CT-BERT  \cite{abs-2005-07503} &  Tweets about the coronavirus & BERT & large &2020.05 & \url{https://github.com/digitalepidemiologylab/covid-twitter-bert} \\ %(a.k.a, COVID-twitter-BERT)
    Med-BERT \cite{abs-2005-12833} &  Cerner Health Facts (general EHR) & BERT & - & 2020.05 &  \url{https://github.com/ZhiGroup/Med-BERT} \\
    ouBioBERT \cite{wada2020pre} &  PubMed & BERT & base & 2020.05 & \url{https://github.com/sy-wada/blue_benchmark_with_transformers} \\
    Bio-ELECTRA \cite{ozyurt2020effectiveness}  & PubMed & ELECTRA & base & 2020.05& \url{https://github.com/SciCrunch/bio_electra} \\ % \url{https://zenodo.org/record/3971235#.YNGnn-gzaUk}
    BERT-XML & Anonymous Institution EHR system  & BERT & small and base &2020.06& \\
    PubMedBERT \cite{gu2020domain} &  PubMed & BERT & base &  2020.07 &  \url{https://huggingface.co/microsoft/BiomedNLP-PubMedBERT-base-uncased-abstract}\\
%  TIMBERT \cite{davari-etal-2020-timbert} & 
% TIMBERT
    MCBERT \cite{zhang2020conceptualized} & Chinese social media, wiki and EHR & BERT & base & 2020.08 & \url{https://github.com/alibaba-research/ChineseBLUE} \\
    BioALBERT \cite{naseem2020bioalbert} &  PubMed and PMC & ALBERT & base \& large & 2020.09 & \url{https://github.com/usmaann/BioALBERT} \\
    BRLTM \cite{Meng_2021} &   private EHR & BERT &customized & 2020.09 &  \url{https://github.com/lanyexiaosa/brltm}\\
  BioMegatron \cite{shin2020biomegatron} & PubMed and PMC & BERT & 0.3/0.8/1.2B & 2020.10 &  \url{https://ngc.nvidia.com/} \\
  ClinicalTransformer \cite{10.1093/jamia/ocaa189}& MIMIC III & \tnote{1}   &base & 2020.10  & \url{https://github.com/uf-hobi-informatics-lab/ClinicalTransformerNER}  \\ %  \footnote{including BERT, RoBERTa, ALBERT, ELECTRA, XLNET}  
    Bioreddit-BERT \cite{basaldella2020cometa} &  healththemed forums on Reddit &BERT & base  & 2020.10 & \url{https://github.com/cambridgeltl/cometa} \\
   BioRoBERTa \cite{lewis2020pretrained}  &     PubMed, PMC, and MIMIC-III  & RoBERTa &base \& large & 2020.11 & \url{https://github.com/facebookresearch/bio-lm} \\ % (we renamed it)
   CODER \cite{yuan2020coder} &  UMLS Metathesaurus  & BERT & base   &  2020.11 &\url{https://github.com/GanjinZero/CODER} \\
    bert-for-radiology \cite{10.1093/bioinformatics/btaa668} &  daily clinical reports &BERT & - &2020.11 & \url{https://github.com/rAIdiance/bert-for-radiology} \\
 BioMedBERT \cite{chakraborty2020biomedbert}  &  BREATHE & BERT & large  & 2020.12 & \url{https://github.com/BioMedBERT/biomedbert} \\
 LBERT \cite{warikoo2021lbert} & PubMed  & BERT & base &  2020.12 & \url{https://github.com/warikoone/LBERT}\\
  ELECTRAMED \cite{miolo2021electramed} & PubMed   & ELECTRA  & base &  2021.04 & \url{https://github.com/gmpoli/electramed}  \\
  SCIFIVE \cite{phan2021scifive} & PubMed Abstract and PMC & T5 &220/770M & 2021.06 & \url{https://github.com/justinphan3110/SciFive} \\
  MedGPT \cite{kraljevic2021medgpt} &  King’s College Hospital   and MIMIC-III &GPY  & customized  & 2021.07 & \url{https://pypi.org/project/medgpt/} \\ % (KCH) NHS Foundation Trust, UK
% { \color{black} Proteinbert~\cite{brandes2022proteinbert} } \\
{\color{black}Clinical-Longformer \cite{li2022clinical} }  & MIMIC-III & Longformer \cite{beltagy2020longformer} & base &2022.01&\url{https://github.com/luoyuanlab/Clinical-Longformer}  \\
{\color{black}Clinical-BigBird \cite{zaheer2020big}   \cite{li2022clinical}  }   & MIMIC-III & BigBird & base  & 2022.01& \url{https://github.com/luoyuanlab/Clinical-Longformer} \\
{\color{black}BioLinkBERT~\cite{yasunaga2022linkbert} } & PubMed  with citation links & BERT & base\& large &  2022.03 & \url{https://github.com/michiyasunaga/LinkBERT} \\
{\color{black}BioBART~\cite{yuan2022biobart} } &  PubMed  & BART & base \& large & 2022.04 & \url{https://github.com/GanjinZero/BioBART} \\
{\color{black}BioGPT\cite{luo2022biogpt} } &  PubMed& GPT &  GPT-2$_\textrm{medium}$\tnote{2} & 2022.09 & \url{https://github.com/microsoft/BioGPT}\\
{\color{black}PubMedGPT } & PubMed & GPT & 2.7B & 2022.12 &  \url{https://www.mosaicml.com/blog/introducing-pubmed-gpt}\\
{\color{black}Flan-PaLM \cite{singhal2022large}  }  &  Instruction \tnote{3} & PaLM \cite{chowdhery2022palm}& 8B,62B and 540B & 2022.12& unavailable \\
{Med-PaLM 2 \cite{singhal2023towards}  }  &  Instruction  \tnote{4} & PaLM 2 \cite{anil2023palm}& 8B,62B and 540B & 2023.5& unavailable \\
{HuatuoGPT \cite{huatuogpt-2023}  }  & Instruction + conversation  & GPT (Bloom~\cite{scao2022bloom})  & 7B & 2023.5& \url{https://github.com/FreedomIntelligence/HuatuoGPT} \\
    \bottomrule
    \end{tabular}
    \begin{tablenotes}
      \item [1]  ClinicalTransformer \cite{10.1093/jamia/ocaa189} provides a series of biomedical models based on different architectures including BERT, RoBERTa, ALBERT, ELECTRA, DistilBERT, XLNet, Longformer, and DeBERTa.
      \item [2] BioGPT adopts GPT-2$_\textrm{medium}$  as the backbone network (24 layers, 1024 hidden size and 16 attention heads), resulting 347M 355M parameters in total. Its parameter size is close to BERT-large.
      \item [3]  \cite{singhal2022large} adopts instruction prompt tuning on medical data. The details were not introduced.
      \item [4] Instructions are from MedQA, MedMCQA, HealthSearchQA, LiveQA and MedicationQA. 
      % \item [4] The results are produced without additional task-specific distillation. 
    \end{tablenotes}
    \end{threeparttable}    
    \label{tab:models_overview} }
\end{table}

% Coder (Yuan et al., 2022b) and SapBERT (Liu et al., 2021) take advantage of the synonyms resource from biomedical knowledge base UMLS (Bodenreider, 2004) and enhance the model with entity knowledge by contrastive pretraining.
% KeBioLM (Yuan et al., 2021) uses Entity as Experts (Févry et al., 2020) model to inject biomedical entity knowledge into the language model, starting from the weights of PubMedBERT
%  Bio-lm (Lewis et al., 2020b) is pretrained on data from PubMed, PMC,
% and MIMIC-III based on the RoBERTa model
% BioMed-RoBERTa (Gururangan et al., 2020) is initialized from RoBERTa (Liu et al., 2019), with additional training on the scientific papers from Semantic Scholar
% ScholarBERT
% BioNLP [46] Lewis, P., Ott, M., Du, J. & Stoyanov, V. Pretrained language models for biomedical and clinical tasks: Understanding and extending the state-of-the-art in Proceedings of the 3rd Clinical Natural Language Processing Workshop (2020), 146–157
 % DARE [66],  DARE: Data augmented relation extraction with gpt-2.
% Nadav Brandes, Dan Ofer, Yam Peleg, Nadav Rappoport, and Michal Linial. Proteinbert: A universal deep-learning model of protein sequence and function \cite{brandes2022proteinbert}

We introduce some representative pre-trained language models, including encoder-only pre-trained language models like  BioBERT, ClinicalBERT, SciBERT, and COVID-twitter-BERT, decoder-only pre-trained language models like MedGPT, and encoder-decoder pre-trained language models like SCIFIVE.
\begin{itemize}
\item  \textbf{BioBERT}~\cite{LeeYKKKSK20} is initialized with the general BERT model and pre-trained on PubMed abstracts and PMC full-text articles. 
% It is further fine-tuned for biomedical text mining tasks such as named entity recognition (NER), question answering, and relation extraction. 

\item  \textbf{ClinicalBERT}~\cite{abs-1904-05342} is trained on clinical text from approximately 2M notes in the MIMIC-III database~\cite{johnson2016mimic}, a publicly available dataset of clinical notes. 

% \item  Flair embeddings from PubMed - A language model available through the Flair framework and embedding method. Trained over a 5\% sample of PubMed abstracts until 2015, or > 1.2 million abstracts in total.

\item  \textbf{SciBERT}~\cite{BeltagyLC19} is trained on the large scale of scientific papers from a multi-domain based on the BERT. The training papers are  from 1.14 M full-text papers in Semantic Scholar, in which $82\%$ articles are from the biomedical domain.

\item  \textbf{COVID-twitter-BERT}~\cite{abs-2005-07503}
is a natural language model to analyze COVID-19 content on Twitter. The COVID-twitter-BERT model is trained on a corpus of 160M tweets about the coronavirus collected through the Crowdbreaks platform  during the period from January 12 to April 16, 2020. 

\item \textbf{MedGPT}~\cite{kraljevic2021medgpt} is a  GPT-like language model trained by patients' medical history  in the format of electronic health records (EHRs). Given the sequence of past events, MedGPT aims to predict future events like a diagnosis of a new disorder or complications of an existing disorder.

\item \textbf{SCIFIVE}~\cite{phan2021scifive} is a domain-specific T5 model which is pre-trained on large biomedical corpora. Like T5, SCIFIVE is a typical Seq2seq paradigm to transform an input sequence into an output sequence.
\end{itemize}

\subsubsection{Discussions on Biomedical Pre-trained Language Models}
Here, we will discuss the listed models in various aspects as below:

\paragraph{Training corpora: EHR, literature, social media, etc., or the hybrid?}
Most pre-trained language models are based on scientific publications \textit{e.g.}, PubMed, and EHR notes. Note that EHR datasets are usually relatively smaller than scientific publications datasets or Wikipedia. Hence pre-trained language models with only EHR datasets are typically trained from the initialization of well-trained BERT  \cite{abs-1904-03323,abs-1904-05342}, XLNET\cite{clinicalxlnet}, etc. Furthermore, some PLMs (\textit{e.g.}, BioRoBERTa~\cite{lewis2020pretrained}) adopt both scientific publications and EHRs. A few models such as CT-BERT and Bioreddit-BERT~\cite{abs-2005-07503,basaldella2020cometa} adopt social media, including Twitter and Reddit.

\paragraph{Extra features}
EHR data usually have some extra meaningful features, for example, disease codes, personal information of patients like age, gender. Such extra features can be embedded as dense vectors used in some models such as Med-BERT and BEHRT~\cite{abs-2005-12833,li2020behrt} like word embedding, position embedding, and segment embedding that are used in the embedding layer of Transformer.

\paragraph{Training from scratch or continue training}
The standard approach to obtain a biomedical pre-trained model is to conduct
continual pre-training from a general-domain pre-trained model like BERT~\cite{devlin2018bert}, such as the BioBERT \cite{yu2019biobert}. Specifically, this approach would initialize the model with the standard BERT model, including its word vocabulary, which is pre-trained by general Wikipedia and BookCorpus.  Besides, some literature demonstrated training from scratch may fully make use of in-domain data and reduce the negative effect from out-of-domain corpora, which may be beneficial for downstream tasks such as PubMedBERT~\cite{gu2020domain}. 

\paragraph{Reusing existing vocabulary or building a new one}
To make use of well-trained general pre-trained language models like BERT~\cite{devlin2018bert}, one has to reuse its vocabulary~\cite{gu2020domain}. However, Biomedical NLP is more challenging than general NLP because it involves jargon and abbreviations: clinical notes have different syntax and grammar than books or encyclopedias. Moreover, a totally new vocabulary necessarily leads to training from scratch due to different vocabularies that may be more computationally expensive. 

% SciBERT Vocabulary BERT uses WordPiece (Wu et al., 2016) for unsupervised tokenization of the input text. The vocabulary is built such that it contains the most frequently used words or subword units. We refer to the original vocabulary released with BERT as BASEVOCAB. We construct SCIVOCAB, a new WordPiece vocabulary on our scientific corpus using the SentencePiece1 library. We produce both cased and uncased vocabularies and set the vocabulary size to 30K to match the size of BASEVOCAB. The resulting token overlap between BASEVOCAB and SCIVOCAB is 42\%, illustrating a substantial difference in frequently used words between scientific and general domain texts
\paragraph{Model size}
Typically, big models usually have a bigger capacity that needs more data for training. However, the biomedical domain usually does have as many corpora as the general domain. Thus, biomedical pre-trained language models are relatively smaller than general pre-trained language models.  Another reason is that most of them are based on BERT or BERT-like encoder-based models, while pre-trained models with decoder architecture (\textit{e.g.}, GPT, T5) could be bigger than encoder-based pre-trained models. To the best of our knowledge, the biggest model is Biomegatron \cite{shin2020biomegatron} with 1.2B parameters.  Note  that  bigger models take longer for inference, which is unfriendly for those researchers without enough research computing resources.

% [Scaling laws for language model, OpenAI]

\paragraph{Being publicly available}
Thanks to the open-sourced tradition of computer science, most models have web pages for downloading and documents for usage. Some of them standardized their model in huggingface (\url{https://huggingface.co}), which will largely be beneficial for its wide-spreading. However, some models are not available to the public due to privacy issues even though data might have been anonymized~\cite{lehman2021does}. 

\paragraph{Biomedical pre-trained language models in other languages}
Most of the biomedical pre-trained language models are in English. However, there is an increasing need for biomedical pre-trained language models in other languages. There are typically two solutions: a multilingual solution or a purely second-language solution. The former may be beneficial for low-resource languages, and the latter is usually used in some rich-resource languages like Chinese~\cite{zhang2020conceptualized}.

\subsection{\textcolor{black}{Beyond Text: Biomedical Vision-and-Language Models}}
\label{sec:multi_modal_biomedical_language_models}

% \benyou{a table to list some models}
\begin{table}[t]\small
\centering
\setlength{\tabcolsep}{2pt}
\scriptsize
\caption{\textcolor{black}{Existing biomedical vision-and-language pre-trained models. The date is based on the submission in arXiv or published data of the journal or conference proceeding.}}
\begin{tabular}{@{}llllllll@{}}
\toprule
Model                                  & Date    & Type           & Image Encoder & Text Encoder    & Fusion Module & Corpora                       & Downstream Datasets                        \\ \midrule
UMRL~\cite{hsu2018umrl}                & 2018.11 & Dual-Encoder   & DenseNet      & GloVe           & -             & MIMIC-CXR                     & ICD-9-IT                                   \\
ConVIRT~\cite{zhang2020convirt}        & 2020.10 & Dual-Encoder   & ResNet        & ClinicalBERT    & -             & MIMIC-CXR, RIH-BONE           & CheXpert, COVIDx, MURA, RSNA               \\
MulInfo~\cite{liao2021mulinfo}         & 2021.05 & Dual-Encoder   & ResNet        & ClinicalBERT    & -             & MIMIC-CXR                     & Pathology9, EdemaSeverity                  \\
GLoRIA~\cite{huang2021gloria}          & 2021.10 & Dual-Encoder   & ResNet        & BioClinicalBERT & -             & CheXpert                      & CheXpert, RSNA, SIIM                       \\
LoVT~\cite{muller2022lovt}             & 2021.12 & Dual-Encoder   & ResNet        & ClinicalBERT    & -             & MIMIC-CXR                     & COVID-Rural, NIH-CXR, Object CXR, SIIM     \\
BioViL~\cite{boecking2022biovil}       & 2022.04 & Dual-Encoder   & ResNet        & CXR-BERT        & -             & MIMIC-CXR                     & MS-CXR, RSNA                               \\
BFSPR~\cite{seibold2022bfspr}          & 2022.05 & Dual-Encoder   & CLIP-Image    & CLIP-Text       & -             & MIMIC-CXR                     & CheXpert, MIMIC-CXR, NIH-CXR, PadChest     \\
CheXZero~\cite{tiu2022chexzero}        & 2022.09 & Dual-Encoder   & CLIP-Image    & CLIP-Text       & -             & MIMIC-CXR                     & CheXpert, PadChest                         \\
MedCLIP~\cite{wang2022medclip}         & 2022.10 & Dual-Encoder   & ResNet/ViT    & BioClinicalBERT & -             & CheXpert, MIMIC-CXR           & CheXpert, COVID, MIMIC-CXR, RSNA           \\
MGCA~\cite{wang2022mgca}               & 2022.10 & Dual-Encoder   & ResNet/ViT    & BioClinicalBERT & -             & MIMIC-CXR                     & CheXpert, RSNA, SIIM                       \\
Analysis~\cite{muller2022role}         & 2022.11 & Dual-Encoder   & ResNet        & ClinicalBERT    & -             & MIMIC-CXR                     & COVID-Rural, NIH-CXR, Object CXR, SIIM     \\
Analysis~\cite{li2020comparison}       & 2020.09 & Fusion-Encoder & -             & -               & -             & MIMIC-CXR                     & IU-Xray, MIMIC-CXR                         \\
Analysis~\cite{wang2021mixed}          & 2021.03 & Fusion-Encoder & ResNet        & BERT            & Dual-Stream   & MIMIC-CXR, NIH14-CXR, IU-Xray & MIMIC-CXR, NIH14-CXR, IU-Xray              \\
Med-ViLL~\cite{moon2022medvill}        & 2021.05 & Fusion-Encoder & ResNet        & BERT            & Single-Stream & MIMIC-CXR                     & MIMIC-CXR, IU-Xray, VQA-RAD                \\
Berthop~\cite{monajatipoor2022berthop} & 2021.08 & Fusion-Encoder & ResNet        & BlueBERT        & Single-Stream & IU-Xray                       & IU-Xray                                    \\
LViT~\cite{li2022lvit}                 & 2022.06 & Fusion-Encoder & ViT           & BERT            & Single-Stream & QaTa-COV19, MoNuSeg           & QaTa-COV19, MoNuSeg                        \\
M3AE~\cite{chen2022m3ae}               & 2022.09 & Fusion-Encoder & CLIP-Image    & RoBERTa         & Dual-Stream   & MedICaT, ROCO                 & VQA-RAD, SLAKE, MedVQA-2019, MELINDA, ROCO \\
ARL~\cite{chen2022arl}                 & 2022.09 & Fusion-Encoder & CLIP-Image    & RoBERTa         & Dual-Stream   & MedICaT, MIMIC-CXR, ROCO      & VQA-RAD, SLAKE, MedVQA-2019, MELINDA, ROCO \\ \bottomrule
\end{tabular}
\label{tab:vl_models_overview}
\end{table}

{\color{black}

Biomedical data is inherently multi-modal. 
It includes various types of data: text data, imaging data, tabular data, time-series data, and structured sequence data (e.g., proteins and DNA).
Among them, the joint learning of text and imaging data is one of the most explored directions, and biomedical vision-and-language pre-training has emerged as an attractive direction in both artificial intelligence and clinical medicine.
This owes to two facts: (i) From the technical perspective, computer vision and natural language processing have been the most popular directions in the past few years, and many models and algorithms have been proposed to process these two types of data; (ii) From the data perspective, the text and imaging data are much easier to obtain in the medical domain, and more importantly they are always pair-collected (e.g., radiology images and their corresponding diagnostic reports).

Most existing biomedical vision-and-language models are motivated by the success of the self-supervised pre-training recipe of SimCLR \cite{chen2020simclr} in CV and BERT in NLP.
Most recently, there have also been some studies \cite{chambon2022adapting,chambon2022roentgen} applying the popular text-to-image diffusion models \cite{ramesh2022dalle2,rombach2022stablediffusion,saharia2022imagen} to the medical domain.
In this subsection, we summarize the existing biomedical vision-and-language models in \ref{sec:overview_vl_models} and describe them in detail.

}

\subsubsection{\textcolor{black}{Overview of Existing Biomedical Vision-and-Language Models}}\label{sec:overview_vl_models}

{\color{black}

In biomedical vision-and-language pre-training, most existing studies could be categorized into two classes, i.e., dual-encoder and fusion encoder. These two types of models have different advantages and disadvantages. Dual-encoder models are able to capture the relationship between visual and linguistic elements in input by independently encoding each modality and then performing shallow iteration on the resulting vectors. This allows them to effectively learn representations that can be used for single-modal/cross-modal tasks, e.g., image classification, image captioning, and cross-modal retrieval. However, dual-encoder models are limited in their ability to fully capture the complex interactions between visual and linguistic elements, which can limit their performance on more challenging vision-and-language tasks.

On the other hand, fusion-encoder models aim to overcome this limitation by directly incorporating visual and linguistic elements into a single encoder. This allows them to capture more complex interactions between the two modalities, which can improve their performance on tasks that require a deeper understanding of the relationship between visual and linguistic elements. They jointly process these two modalities with an early interaction to learn multi-modal representations to solve those tasks requiring multi-modal reasoning, e.g., visual question answering. However, it can be more difficult to perform single-modal tasks, as the interactions between visual and linguistic elements are not as easily separated as they are in dual-encoder models. Tab. \ref{tab:vl_models_overview} presents existing dual-encoder and fusion-encoder vision-and-language models.

In addition to dual-encoder and fusion-encoder models, there are other approaches for biomedical vision-and-language pre-training. For example, motivated by the success of diffusion models \cite{ramesh2022dalle2,rombach2022stablediffusion,saharia2022imagen} in the general domain, several medical text-to-image diffusion models \cite{chambon2022adapting,chambon2022roentgen} have been proposed in the medical domains.

}

\subsubsection{\textcolor{black}{Dual-Encoder Vision-Language Models}}\label{sec:dual_encoder_vl_models}
{\color{black}

Dual-encoder models encode images and texts separately to learn uni-modal/cross-modal representations following a shallow interaction layer (e.g., an image-text contrastive layer).
The learned models can be transferred to many single-modal/cross-modal tasks, e.g., image classification and cross-modal retrieval tasks.
Next, we detail some representative dual-encoder models:
\begin{itemize}
\item \textbf{ConVIRT}~\cite{zhang2020convirt} is the first study to apply contrastive learning to images and texts, inspired by its success in the vision field. For the model architecture, it adopts ResNet and BERT as the vision encoder and the language encoder, respectively. Afterward, a bidirectional contrastive loss between two modalities is used to train these two encoders. It is found that the vision encoder can be used to perform the image classification tasks, requiring much fewer annotated training data as an ImageNet-initialized counterpart to achieve comparable or better performance.

\item \textbf{GLoRIA}~\cite{huang2021gloria} proposed to perform the representation learning of medical images from global and local perspectives. Specifically, for global contrastive learning, it is similar to that of ConVIRT. For local contrastive learning, it uses an attention mechanism to learn local representations by matching the words in radiology reports and image sub-regions. 

\item \textbf{MedCLIP}~\cite{wang2022medclip} is trained on both image-text and image-label datasets. The core idea is to pre-compute the matching scores between an image and its text or an image and its label. Subsequently, the scores are used as the target to perform the learning procedure. It is observed that much fewer data are required to learn good representations for zero-shot disease classification.

\item \textbf{CheXZero}~\cite{tiu2022chexzero} is initialized with the pre-trained CLIP model and pre-trained on the medical image-text dataset. With the strong backbone model and curated designs, CheXZero can achieve comparable results in disease classification tasks in a zero-shot manner.

\item \textbf{LoVT}~\cite{muller2022lovt} is the first dual-encoder study targeting localized medical imaging tasks. It proposed a local contrastive loss to align local representations of sentences or image regions while encouraging spatial smoothness and sensitivity. This promotes its performance on many localized downstream tasks.
\end{itemize}
}

\subsubsection{\textcolor{black}{Fusion-Encoder Vision-Language Models}}\label{sec:fusion_encoder_vl_models}
{\color{black}

Fusion-encoder models encode images and texts and then exploit a fusion module to integrate the image and text features.
For the fusion module, normally, there are two types: (i) single-stream: the models use a single Transformer for early and unconstrained fusion between modalities; (ii) dual-stream: the models adopt the co-attention mechanism to interact with different modalities.
For fusion-encoder models, the most common objectives are masked language modeling and image-text matching.
Similarly, we detail some representative fusion-encoder studies:

\begin{itemize}
\item \textbf{Li et al.}~\cite{li2020comparison} adopted four general-domain pre-trained vision-and-language models (i.e., LXMERT \cite{tan2019lxmert}, VisualBERT \cite{li2019visualbert}, UNITER \cite{chen2020uniter}, and PixelBERT \cite{huang2020pixelbert}) to learn multi-modal representations from medical images and texts. The experimental results demonstrated their effectiveness of them in disease classification tasks.

\item \textbf{MedViLL}~\cite{moon2022medvill} adopted a single BERT-based model and designed a masking scheme to improve both vision-language understanding tasks (e.g., disease classification, cross-modal retrieval, and visual question answering) and vision-language generation tasks (e.g., radiology report generation).

\item \textbf{ARL}~\cite{chen2022arl} proposed to integrate medical-domain knowledge bases (e.g., UMLS) into the fusion encoder. Medical knowledge is exploited from three perspectives: (i) aligning through knowledge, (ii) reasoning using knowledge, and (iii) learning from knowledge.

\item \textbf{LViT}~\cite{li2022lvit} is a vision-and-language fusion-encoder model for medical image segmentation. It leverages medical text annotation to improve the quality of generated segmentation results, especially in the semi-supervised setting.

\end{itemize}
}

\subsubsection{\textcolor{black}{Other Vision-Language Models}}\label{sec:other_vl_models}
{\color{black}

Besides the dual-encoder and fusion-encoder models, there are also some biomedical pre-trained models involving vision and language. We mainly introduce medical text-to-image diffusion models. Diffusion models are a type of generative model inspired by non-equilibrium thermodynamics. By defining a Markov chain of diffusion steps to add random noise to data slowly, the model aims to learn to reverse the diffusion process to construct desired data samples from the noise. Recently, different text-to-image diffusion models (e.g., DALLE-2 \cite{ramesh2022dalle2}, Stable Diffusion \cite{rombach2022stablediffusion}, and Imagen \cite{saharia2022imagen}) have been proposed and achieved excellent performance on text-based image generation. In the medical domain, RoentGen \cite{chambon2022adapting,chambon2022roentgen} investigated the adaptation of Stable Diffusion to the medical domain. In specific, they exploited chest X-ray images and their corresponding reports from the MIMIC-CXR dataset to train the model. Then they explored several adaptation approaches (i.e., partially fine-tuning or fully fine-tuning) and different text encoders for adaptation (e.g., domain-agnostic and domain-specific text encoder). The experiments demonstrated the effectiveness of the model with respect to image quality and clinical accuracy.
}

\subsection{Beyond Text: Language Models for Proteins/DNA}
Various biological sequences like proteins and DNA could also be treated like linguistic tokens in natural language. Therefore, many existing works explored training language models for these biological sequences. One crucial difference between language models for biological sequences and the counterparts for natural language is tokenization (see Sec. \ref{sec:tokenization}), which leads to different token vocabularies.  Sec. \ref{sec:language_model_biomedical_sequence} will summarize the existing language models for these biological sequences.

\subsubsection{Tokenization for Proteins/DNAs}
\label{sec:tokenization}
Like words in the text, biological sequences such as proteins and DNA sequences could also be modeled by language models, which typically aim to predict the next token in a sequence. However, in contrast to that words are in a relatively big vocabulary (typically 10k-100k), and the vocabularies for biological sequences are usually small. 

\paragraph{Tokenization in Proteins}
Since the structure of a protein is fully determined by its amino acid sequence~\cite{anfinsen1973principles}, one can represent a protein by its amino acid sequences. Roughly 500 amino acids have been identified in nature; however, only 20 amino acids are found to make up the proteins in the human body. The vocabulary of protein sequences consists of these 20 typical amino acids.

\paragraph{Tokenization in DNAs}
The two DNA strands are known as polynucleotides, and they are composed of simpler monomeric units (\textit{a.k.a.} nucleotides). Each nucleotide  contains  one of four nitrogen-containing nucleobases (\textit{i.e.}, cytosine [C], guanine [G], adenine [A], or thymine [T]). The two separate polynucleotides are bound together, according to deterministic base pairing rules ([A] with [T] and [C] with [G]), with hydrogen bonds. Typically, existing work~\cite{Ji2020} usually adopts a so-called `$K$-mer' representation for DNA sequences \footnote{ `$K$-mer' is like a $k$-size convolutional window for a sequence. For example, a  DNA sequence {ATGGCT \tt} will be tokenized to a sequence of $3$-mers {\tt\{ATG TGC GGC  GCT\} }  or to a sequence of $5$-mers {\tt\{ATGGC TGGCT\}}.} for richer contextual information for DNAs. By doing so, the vocabulary size will increase to the $4^k +5$ which is exponential to $k$ and additionally pluses five special tokens ({[CLS] \tt}, {[SEP] \tt}, {[PAD] \tt}, {[MASK] \tt}, {[UNK] \tt}).

\subsubsection{Language Models for biological sequences}
\label{sec:language_model_biomedical_sequence}

% \benyou{add a table here}

\paragraph{\textbf{Protein language models}}
Since the commonly-found categories of amino acids are relatively small, namely 20. Initially, some work applied character-level language models to protein to deal with limited-size amino acids.
% Character-level LMs have been recently applied to protein sequences, where the amino acids constitute the vocabulary. 
In the beginning, there were many efforts to training RNN-based language models~\cite{Alley2019,bepler2018learning} for protein sequences. \cite{Heinzinger2019,heinzinger2019modeling} trains a deep bi-directional model ELMo for proteins \footnote{https://github.com/Rostlab/SeqVec}.
Other than those protein sequences,  protein language models usually adopt additional features for proteins, \textit{e.g.},  global structural similarity between proteins and  pairwise
residue contact maps for each protein \cite{bepler2018learning}.
Later, \cite{Rao2019} introduces the Tasks Assessing Protein Embeddings (TAPE), a suite of biologically relevant semi-supervised learning tasks. 
The authors also train language models based on LSTM, Transformer, and ResNet on the protein sequences. 
Bepler et al~\cite{bepler2019learning} also proposed a novel framework based on the LSTM model to learn protein sequence embeddings. They make their embeddings publicly available at \footnote{https://github.com/tbepler/protein-sequence-embedding-iclr2019}.
\cite{Rives2021} trains a contextual transformer-based language model\footnote{The trained model and code are available at \url{https://github.com/facebookresearch/esm}.} on 250 million protein sequences. 
The representations learned by this LM encode multi-level information spanning from the biochemical properties of amino acids to the remote homology of proteins. 
Different from the above line of approaches, MSA Transformer \cite{Rao2021} fits a model separately to each family of proteins. 
ProtTrans \cite{Elnaggar2020} trains a variety of LM models with thousands of GPUs, and also makes the trained models publicly available\footnote{\url{https://github.com/agemagician/ProtTrans}}.
ProGen \cite{Madani2020} is a generative LM trained on 280M protein sequences conditioned on taxonomic and keyword tags. 
ProteinLM~\cite{xiao2021modeling} was recently proposed, which trained a large-scale pre-train model for evolutionary-scale protein sequences, and the trained model is available at\footnote{ https://github.com/THUDM/ProteinLM}.
More recently, DeepMind developed Alphafold2~\cite{jumper2021highly} that could predict protein structures with high accuracy in the challenging 14th Critical Assessment of protein Structure Prediction (CASP14). Most interestingly, there is an embedded protein language model in Alphafold2, which makes Alphafold2 feasible to make use of unlabelled protein data. In detail, Alphafold2 adopts an auxiliary BERT-like loss to predict pre-masked residues in multiple sequence alignments (MSAs). More recently, ProteinBERT \cite{brandes2022proteinbert} was proposed to use  a novel task of Gene Ontology (GO) annotation prediction along with masked language modeling and it is also tailored to  make the model highly efficient and flexible to very large sequence lengths.

% \benyou{
% Nadav Brandes, Dan Ofer, Yam Peleg, Nadav Rappoport, and Michal Linial. Proteinbert: A universal deep-learning model of protein sequence and function \cite{brandes2022proteinbert}
% % Roshan Rao, Joshua Meier, Tom Sercu, Sergey Ovchinnikov, and Alexander Rives. Transformer protein language models are unsupervised structure learners. In 9th International Conference on  Learning Representations, ICLR 2021, Virtual Event, Austria, May 3-7, 2021. OpenReview.net, 2021b. URL https://openreview.net/forum?id=fylclEqgvgd.   \cite{rao2020transformer}
% }

\paragraph{\textbf{DNA language models}}
Proteins are translated from DNA through the genetic code.  There are 20 natural amino acids that are used to build the proteins that DNA encodes. Therefore,  amino acids cannot be one-to-one mapped by  only four nucleotides. Some work also explored the potential to build language models on DNA sequences. 
DNABERT \cite{Ji2020} is a bidirectional encoder pre-trained on genomic DNA sequences with up and downstream nucleotide contexts. Yamada et al \cite{Yamada2021} pre-trains a BERT on RNA sequences and RNA-binding protein sequences. 
All the LMs remain largely the same as those used for human language data. 
Designing new architectures and pipelines tailored to protein/DNA sequences is a promising direction. 

\section{Fine-tuning PLMs for Biomedical Downstream Tasks}
\label{sec:finetuning}
\begin{table}
\scriptsize
    \centering
    \caption{\textcolor{black}{The performance of different biomedical pre-trained language models on downstream tasks. For all biomedical language models, we compare the F1 score of the base model on various tasks. The BLURB Score calculates the macro average of F1 test results on all tasks.}}
    \begin{tabular}{lcccccccccc}
    \toprule
    &BERT&RoBERTa&BioBERT&SciBERT&ClinicalBERT&BlueBERT&PubMedBERT&BioM-ELECTRA&BioLinkBERT&BioGPT\\
    \midrule
    \textbf{NER}&\\
    BC5-chem~\cite{li2016biocreative} &89.99&89.43&92.85&92.51&90.80&91.19&93.33&93.1&\textbf{93.75}&-\\
    BC5-disease~\cite{li2016biocreative}&79.92&80.65&84.70&84.70&83.04&83.69&85.62&85.2&\textbf{86.10}&-\\ 
    NCBI-disease~\cite{dougan2014ncbi}&85.87&86.62&89.13&88.25&86.32&88.04&87.82&\textbf{88.4}&88.18&-\\ 
    BC2GM~\cite{smith2008overview}&81.23&80.90&83.82&83.36&81.71&81.87&84.52&-&\textbf{84.90}&-\\ 
    JNLPBA~\cite{kim2004introduction}&77.51&77.86&78.55&78.51&78.07&77.71&\textbf{79.10}&-&79.03&-\\
    \hline
    \textbf{PICO extraction}&\\
    EBM-PICO~\cite{nye2018corpus}&71.70&73.02&73.18&73.06&72.06&72.54&73.38&-&\textbf{73.97}&-\\
    \hline
    \textbf{Relation extraction}&\\
    ChemProt~\cite{krallinger2017overview}&71.54&72.98&76.14&75.00&72.04&71.46&77.24&\textbf{77.6}&77.57&-\\ 
    DDI~\cite{herrero2013ddi}&79.34&79.52&80.88&81.22&78.20&77.78&82.36&-&\textbf{82.72}&-\\
    GAD~\cite{bravo2015extraction}&79.61&80.63&82.36&81.34&80.48&79.15&83.96&-&\textbf{84.39}&-\\ 
    \hline
    \textbf{Sentence similarity}&\\
    BIOSSES~\cite{souganciouglu2017biosses}&81.40&81.25&89.52&87.15&91.23&85.38&92.30&-&\textbf{93.25}&-\\
    \hline
    \textbf{Document classification}&\\
    HoC~\cite{hanahan2000hallmarks}&80.12&79.66&81.54&81.16&80.74&80.48&82.32&-&84.35&
\textbf{85.12}\\
    \hline
    \textbf{Question answering}&\\
    PubMedQA~\cite{jin2019pubmedqa}&49.96&52.84&60.24&51.40&49.08&48.44&55.84&-&70.20&\textbf{78.2}\\ 
    BioASQ~\cite{nentidis2019results}&74.44&75.20&84.14&74.22&68.50&68.71&87.56&-&\textbf{91.43}&-\\
    \hline
    BLURB Score~\cite{gu2020domain}&75.86&76.46&80.34&78.14&77.29&76.27&81.16&-&\textbf{83.39}&-\\
    \bottomrule
    \end{tabular}
    \label{tab:task}
\end{table}
\begin{table}[t] 
\scriptsize
    \centering
    \caption{{Example for each downstream task.}}
    \begin{tabular}{llll}
    \toprule
    Task &Input&Output&Example\\
    \midrule
Named Entity Recognition&Unannotated biomedical text&Annotated text with biomedical entities identified&E.g., identifying drug names, disease terms in text\\
Relation Extraction&Text with annotated entities&Text with relations between entities identified&E.g., recognizing drug-disease treatment relations\\
Event Extraction&Text with annotated entities and relations&Text with biomedical events identified&E.g., identifying gene-mutation-event in the literature\\
Text Classification&Biomedical text&Classified text into pre-defined categories&	E.g., classifying medical reports based on disease types\\
Sentence Similarity&Pair of sentences&Similarity score between the sentences&	E.g., measuring similarity between two medical findings\\
Question Answering&Question and context&Answer to the question based on context&	E.g., answering clinical questions based on medical textbooks\\
Dialogue Systems&User input&System response&E.g., virtual health assistant responding to user health queries\\
Text Summarization&Long biomedical text&Short summary of the text&E.g., summarizing a medical research article\\
Natural Language Inference&Pair of sentences&Inference relation between the sentences&E.g., inferring medical conclusions from patient's symptoms\\
    \bottomrule
    \end{tabular}
    \label{tab:tsex}
  \end{table}
Similar to the general domain, to evaluate the effectiveness and facilitate the research development of biomedical pre-trained language models, the Biomedical Language Understanding Evaluation (BLUE) benchmark has been proposed in~\cite{peng2019transfer}. BLUE includes five text mining tasks in biomedical natural language processing, including sentence similarity, named entity recognition, relation extraction, text classification, and inference task. However, BLUE does not include some important biomedical application tasks such as question answering, and it mixes the applications of clinical data and biomedical literature. To improve it, 
Gu et al \cite{gu2020domain} proposed a novel benchmark, the Biomedical Language Understanding \& Reasoning Benchmark (BLURB). It includes named entity recognition (NER), evidence-based medical information extraction (PICO), relation extraction, sentence similarity, document classification, and question-answering tasks. Moreover, some works proposed the benchmark in other languages, such as Chinese~\cite{zhang2020conceptualized}. 

\textcolor{black}{The development of biomedical pre-trained language models has greatly boosted the performance of these downstream tasks recently.
In Table \ref{tab:task}, we show the performance when directly fine-tuning different biomedical pre-trained language models for downstream tasks.
All biomedical pre-trained language models significantly outperform PLMs in the general domain including BERT and RoBERTa.
Especially for sentence similarity and question-answering tasks, the biomedical PLMs such as PubMedBERT and BioLinkBERT outperform BERT and RoBERTa by more than 10\% percent.
PubMedBERT conducts the domain-specific pre-training from scratch and consistently outperforms other biomedical PLMs such as BioBERT, ClinicalBERT, and BlueBERT in all tasks.
Most recently, BioLinkBERT~\cite{yasunaga2022linkbert} further utilizes the citation links of documents from PubMed abstracts in the pre-training stage, and has achieved the SOTA performance on most tasks.
Specifically, for the document level task such as the question-answering task, it outperforms PubMedBERT by 15\% percent in the PubMedQA dataset and 4\% percent in the BioASQ dataset.
In the PubMedQA dataset and another document-level task: document classification, the BioGPT~\cite{luo2022biogpt} achieves the new SOTA, which conducts the generative pre-training on PubMed abstracts from scratch like GPT.}

Besides directly fine-tuning, there is other research exploring how to better leverage and improve PLMs for various downstream tasks.
In the following, we will introduce the recent progress based on PLMs on these tasks ({we show the example of each downstream task in Table \ref{tab:tsex}}) and other critical tasks in the biomedical domain.

\subsection{Information Extraction}
Information extraction plays a key role in automatically extracting structured biomedical information (entities, concepts, relations, and events) from unstructured biomedical text data ranging from biomedical literature, and electronic health records (EHR) to biomedical-related social media corpus, etc (one can check a review in \cite{wang2018clinical}). 
%It plays an important role in  intelligent healthcare applications such as clinical decision support, biocuration assistance, and health monitoring. 
In the biomedical community, it generally refers to several important sub-tasks, including named entity recognition (NER), relation extraction, and event extraction.

%Relation extraction is to determine the relationships among biomedical entities, concepts and attributes.
%imilar to the general domain, learning methods for biomedical information extraction have been rapidly advanced based on the pre-trained language models recently. 
%We will next introduce the methods progress based on the pre-trained language models in each sub-tasks.
\subsubsection{Named entity recognition}
NER aims to identify the common biomedical concept mentions or entity names (such as genes, drug names, adverse effects, metabolites, and diseases) of biomedical texts.
\textcolor{black}{Singh et al~\cite{sachan2018effective} proposed the first effort to investigate pre-training the bidirectional language model with the PubMed abstract dataset, and then fine-tune the model for the supervised NER task.
Compared with traditional neural network based methods such as BiLSTM, it outperforms them by around 1\% in the F1-score in datasets including NCBI-disease~\cite{dougan2014ncbi}, BC5-disease~\cite{li2016biocreative}, BC2GM~\cite{smith2008overview}, JNLPBA~\cite{kim2004introduction}, and requires less labelled training data to achieve comparable results.
Several methods have shown that further pre-training the language models on the in-domain data can consistently 
improve the performance.
For example, Zhu et al~\cite{zhu2018clinical}\footnote{https://github.com/noc-lab/clinical\_concept\_extraction} trained a domain-specific ELMo model in the mixture data of clinical reports and relevant Wikipedia pages, which outperforms the previous SOTA method based on BiLSTM-CRF by 3.4\% in F1-score in the i2b2 2010~\cite{uzuner20112010} dataset.
Si et al~\cite{si2019enhancing} have shown that the BERT-large further pre-trained on the MIMIC-III achieves the best performance for the i2b2 2010 dataset, and improves the performance by 5\% over that of the traditional neural network method based on the GloVe embedding.
Sheikhshab et al~\cite{sheikhshab2018domain} have shown that directly using the off-the-shelf ELMo embeddings has limited improvement on the performance, while ELMo continually pre-trained on the in-domain data has significant improvement on the performance by 4\% in the F1 score of the JNLPBA dataset.
Gao et al~\cite{gao2021pre}\footnote{\url{https://code.ornl.gov/biomedner/biomedner}} investigated the pre-training and semi-supervised self-training of BiLSTM-CRF and BlueBERT with the in-domain corpora such as MedMentions and SemMed.
They evaluated these models on BioNER with limited labeled training data, and the BluBERT pre-trained on MedMentions has the best performance overall.
Moreover, for the scenarios with very few labeled data, the semi-supervised self-training can significantly boost performance.}

\textcolor{black}{Some methods have explored how to utilize PLMs for BioNER with less time and computational consumption.
Naseem et al~\cite{naseem2020bioalbert}\footnote{\url{https://github.com/usmaann/BioALBERT}} proposed a lightweight domain-specific language model BioALBERT trained on the biomedical domain corpora for biomedical named entity recognition, that captures inter-sentence coherence via the sentence-order-prediction (SOP) task.
For eight benchmark datasets, it outperforms the BioBERT by a significant margin, such as increasing the performance of the F1-score by 7.47\% in the NCBI–disease dataset and 10.63\% in the BC5CDR–disease dataset.
Poerner et al~\cite{poerner2020inexpensive}\footnote{\url{https://github.com/npoe/covid-qa}} proposed the time and memory saving domain-adaption method: training Word2Vec on target domain text and aligning them with the word vectors of existing PLMs, and thus propose the GreenBioBERT.
On eight BioNER datasets, the GreenBioBERT covers 60\% results of BioBERT but only takes 2\% of its cloud compute cost.
Moreover, there are methods incorporating BioNER with the relation extraction task or modeling BioNER beyond the sequence labeling problem.
Khan et al~\cite{khan2020mt} employed PLMs including BERT and BioBERT as the encoder for the multi-task learning of BioNER.
They found that using BioBERT has moderately better performance than BERT, and it requires more training epochs for the BERT based method to achieve comparable results.
Giorgi et al~\cite{giorgi2019end}\footnote{\url{https://github.com/bowang-lab/joint-ner-and-re}} proposed the end-to-end model for jointly extracting named entities and their relations using PLMs as the encoder. 
However, in the i2b2 2010~\cite{uzuner20112010} dataset, it has worse performance than the method proposed by Si et al~\cite{si2019enhancing} and BlueBERT.
Sun et al~\cite{sun2020biomedical}\footnote{\url{https://github.com/CongSun-dlut/BioBERT-MRC}} proposed to model the BioNER as the machine reading comprehension (MRC) problem to incorporate the prior knowledge flexibly, and use PLMs as the text encoder.
Among ClinicalBERT, BlueBERT and BioBERT, the method based on BioBERT achieves the best performance.
Tong et al~\cite{tong2021multi} proposed the auxiliary sentence-level prediction tasks, which can improve the F1 score by 3\% in the low-resource scenario on three benchmark datasets compared with BioBERT.
Banerjee et al~\cite{banerjee2021biomedical}\footnote{\url{https://github.com/kuntalkumarpal/KGQA}} formulated the BioNER as the knowledge-guided question-answer task (KGQA), that outperforms the SOTA by 1.78–12\% on 11 biomedical NER datasets in the exact match F1 score.}
%With the development of general domain language models, 
%Gao et al ~\cite{gao2021pre} explored the effectiveness of transfer learning based on the pre-trained language models and semi-supervised self-training to improve the performance of biomedical NER with very limited labeled data.
%To save the memory and time consuming of domain adaption, Poerner et al~\cite{poerner2020inexpensive} proposed to train Word2Vec on biomedical domain text and incorporated it into the general domain BERT to improve the performance of the biomedical NER task.
 %Besides English, there is much work exploring the pre-trained language models on the BioNER of other languages, including Chinese~\cite{dai2019named,jiang2019bert,li2020chinese,xue2019fine,wang2020bert,li2020towards}, Spanish~\cite{akhtyamova2020named,hakala2019biomedical,miranda2020named}, French~\cite{copara2020contextualized}, Korean~\cite{kim2020korean}, Russian~\cite{miftahutdinov2020biomedical}, Arabic~\cite{boudjellal2021abioner}, Italian~\cite{catelli2020crosslingual}. 
 
\textcolor{black}{
\paragraph{Summary}
In Table~\ref{tab:ner}, we summarize the commonly used datasets in the BioNER task and compare performances of different methods on these datasets in Table \ref{tab:nermethod}.
We can find that the lightweight BioALBERT~\cite{naseem2020bioalbert} model pre-trained on the sentence-order-prediction (SOP) task, is the SOTA method on almost all datasets.
Among various PLMs, several methods~\cite{tong2021multi,khan2020mt,sun2020biomedical} show that using BioBERT generally shows better performance than other PLMs such as BERT, ClinicalBERT, and BlueBERT.
Several methods~\cite{zhu2018clinical,si2019enhancing,sheikhshab2018domain,gao2021pre} show that pre-training various PLMs such as ELMO, BERT and BlueBERT with various in-domain data such as MIMIC-III and MedMentions, can consistently improve the performance.
}
\begin{table} 
\scriptsize
    \centering
    \caption{Datasets used in the BioNER task.}
    \begin{tabular}{llllll}
    \toprule
    Dataset & Language & Entity type &Text type&Text Genre&Size\\
    \midrule
    BC5-chem~\cite{li2016biocreative}& English &Chemical&Abstract&PubMed&1,500\\
    BC4-chem~\cite{krallinger2015chemdner} &English&Chemical&Full text&PubMd&10,000\\
    BC5-disease~\cite{li2016biocreative} & English & Disease&Abstract&PubMed&1,500\\
    NCBI-disease~\cite{dougan2014ncbi} & English & Disease&Abstract&PubMed&793\\
    i2b2 2010~\cite{uzuner20112010} &English&Disease&Report&Clinical records&871\\
    BC2GM~\cite{smith2008overview}& English& Gene/Protein &Sentence&MEDLINE&20,000\\
    JNLPBA~\cite{kim2004introduction} & English & Protein,DNA,RNA,cell line&Abstract&MEDLINE&2,404\\
    LINNAEUS~\cite{gerner2010linnaeus} &English& Species&Full text&PMC&100\\
    Species-800~\cite{pafilis2013species}&English&Species &Abstract&MEDLINE&800\\
    EBM PICO~\cite{nye2018corpus} & English & Participants,interventions, outcomes&Abstract&PubMed&4,993\\
    CCKS 2017&Chinese&Body,disease,symptom,test,treatment&Report&Clinical Records&400\\
    CCKS 2018&Chinese&Anatomy,symptom,independent,drug,operation&Report&Clinical Records&1,000\\ PharmaCoNER~\cite{agirre2019pharmaconer}&Spanish&Protein,chemical&Report&Spanish Clinical Case Corpus&1,000\\
    CANTEMIST~\cite{miranda2020named}&Spanish&Tumor morphology&Report&Spanish Clinical Case Corpus&1,301\\
    CAS~\cite{grabar2018cas}&French&Terms,negation,uncertainty&Clinical cases&PubMed&100\\
    \bottomrule
    \end{tabular}
    \label{tab:ner}
  \end{table} 
\begin{table} 
\scriptsize
    \centering
    \caption{\textcolor{black}{Performances (F1-score) of different methods on benchmark datasets.}}
    \begin{tabular}{lllllllll}
    \toprule
    & BC5-chem & BC4-chem & BC5-disease&i2b2 2010&BC2GM&JNLPBA&LINNAEUS&Species-800\\
    \midrule
    Singh et al~\cite{sachan2018effective}&-&-&89.28&-&81.69&75.03&-&-\\
    Zhu et al~\cite{zhu2018clinical}&-&-&-&88.60&-&-&-&-\\
    Si et al~\cite{si2019enhancing}&-&-&-&89.55&-&-&-&-\\
    Sheikhshab et al~\cite{sheikhshab2018domain}&-&-&-&-&89.72&70.08&-&-\\
    Gao et al~\cite{gao2021pre}&91.80&88.38&84.02&-&80.56&81.44&91.36&72.49\\
    Naseem et al~\cite{naseem2020bioalbert}&\textbf{97.79}&\textbf{96.23}&\textbf{97.61}&-&\textbf{96.33}&\textbf{83.53}&\textbf{99.73}&\textbf{98.72}\\
     Poerner et al~\cite{poerner2020inexpensive}& 93.08& 91.26&85.08&-&83.45 &76.89 &88.34&74.31 \\
      Khan et al~\cite{khan2020mt}&90.52&-&-&-&83.01&-&-&-\\
      Giorgi et al~\cite{giorgi2019end}&-&-&-&89.26&-&-&-&-\\
      Sun et al~\cite{sun2020biomedical}&94.11&92.70&87.56&-&85.11&78.45&-&-\\
      Tong et al~\cite{tong2021multi}\footnote{\url{https://github.com/ zgzjdx/MT-BioNER}}&93.98&-&-&-&84.78&-&-&-\\
      Banerjee et al~\cite{banerjee2021biomedical}&90.50&92.39&-&\textbf{92.67}&83.47&79.19&92.63&-\\
    \bottomrule
    \end{tabular}
    \label{tab:nermethod}
  \end{table} 
\subsubsection{Relation Extraction}
Biomedical relation extraction (BioRE) aims to identify the relationship (semantic correlation) between biomedical entities mentioned (such as genes, proteins, and diseases) in texts and generally be considered as a classification problem to predict the possible relation type of two identified entities in a given sentence. 
%In the past few years, deep neural networks, including recurrent neural networks (RNNs), convolutional neural networks (CNNs), and graph neural networks (GNNs) have shown their effectiveness in the BioRE task, which are efficient to encode the semantic features of entities and sentences. They rely on labeled data to train the model. However, the annotated data is quite limited in the biomedical domain. 
Recently, PLMs have been widely explored in the BioRE. 
%In Table \ref{tab:re}, we summarize the commonly used datasets in the BioRE task. 
\textcolor{black}{Wei et al~\cite{wei2019relation} conducted the first study that investigated fine-tuning BERT and combining additional BIO tag features for the clinical RE.
It shows that the BERT-based model outperforms previous SOTA methods based on deep neural networks on n2c2~\cite{henry20202018} and i2b2~\cite{uzuner20112010} dataset.
%Similarly, Hafiane et al~\cite{hafiane2020experiments} investigated two transfer learning strategies: frozen and fine-tuning to adapt BERT on the biomedical RE. 
Similarly, Thillaisundaram et al~\cite{thillaisundaram2019biomedical} adapted the SciBERT to the BioRE via fine-tuning the representation of the classification token (CLS). 
However, it only compared with and outperformed a simple sampling-based baseline.
To further explore the potential of utilizing full information in the last layer to improve performance, 
Su et al~\cite{su2020investigation} proposed to utilize all outputs of the last layer when fine-tuning the BioBERT model on the BioRE task, which outperforms the BioBERT only using classification token on the DDI~\cite{herrero2013ddi}, PPI~\cite{krallinger2008overview} and ChemProt~\cite{krallinger2017overview} dataset.
Su et al~\cite{su2021improving} proposed to employ the contrastive  learning to improve fine-tuning BERT model for biomedical relation extraction, which outperforms directly fine-tuning BERT on the DDI, PPI and ChemProt dataset. 
Xue et al~\cite{xue2019fine} proposed to fine-tune BERT for joint entity and relation extraction in Chinese medical text, which outperforms the SOTA joint model based on Bi-LSTM by 1.6\%. 
Chen et al~\cite{chen2019general} combined BERT with the one-dimensional convolutional neural network (1D-CNN) for the medical relation extraction, which significantly outperforms the traditional 1D-CNN classifier.
Lin et al~\cite{lin2019bert,lin2020bert} combined the global embeddings and multi-task learning to improve BERT on the clinical temporal relation extraction. 
Guan et al~\cite{guan2020robustly} investigated several PLMs including BERT, RoBERTa, ALBERT, XLNet, BioBERT, ClinicalBERT, in predicting the relationships between clinical events and temporal expressions, and found that RoBERTa generally has the best performance. 
%To make the biomedical relation extraction in the document with long-distance dependencies and complex semantics, Liu et al~\cite{liu2020document} proposed to use the pre-trained self-attention structure for biomedical relation extraction in the document level with the entity replacement method.
%In many real application scenarios, training the medical relation extraction model generally requires collecting and storing privacy-sensitive data, which may conflict with privacy protection. 
To prevent private information leakage, Sui et al~\cite{sui2020feded} proposed the first privacy-preserving medical relation extraction method FedED based on BERT and federated learning, which achieved promising results on three benchmark datasets.}

\textcolor{black}{
\paragraph{Summary}
In Table \ref{tab:re} and Table \ref{tab:remethod}, we summary the commonly used datasets and compare the performances of different methods on these datasets.
In summary, fine-tuning various PLMs significantly outperforms traditional neural network based methods~\cite{wei2019relation,chen2019general}, and improving the fine-tuning strategy can further improve the performance, for example Su et al~\cite{su2021improving} using the contrastive learning as the auxiliary task, achieves the best performance on DDI, PPI and ChemProt.
}
%Moreover, similar to BioNER, several advanced biomedical domain PLMs as mentioned in the last section, such as BioBERT~\cite{LeeYKKKSK20}, BlueBERT~\cite{peng2019transfer}, SciBERT~\cite{BeltagyLC19}, PubMedBERT~\cite{gu2020domain}, BioMegatron~\cite{shin2020biomegatron} have achieved great performance on the BioRE task. 
% They typically adapt to the BioRE task with the extra binary classification layer (linear layer or MLP). 
\begin{table}[t] \footnotesize
% \begin{minipage}{.55\textwidth}
    \centering
    \caption{Datasets used in the BioRE task.}
    \begin{tabular}{lllll}
    \toprule
    Dataset & Entity type &Text type&Relation Size\\
    \midrule
    i2b2 2010~\cite{uzuner20112010}&Medical problem—treatment&Report&5,261\\
    i2b2 2012~\cite{sun2013evaluating} & Event–temporal expression&Summary&8,294\\
   TM~\cite{styler2014temporal}& Event–event&EHRs&355\\
    DDI~\cite{herrero2013ddi}& Drug–drug&Abstract&48,223\\
    PPI~\cite{krallinger2008overview}&Protein-protein&Abstract&5,834\\
    ChemProt~\cite{krallinger2017overview}&Protein–chemical&Abstract&31,784\\
    BioC VI PM~\cite{dogan2017biocreative}&Protein–protein&Full text&1,629\\
    \bottomrule
    \end{tabular}
    \label{tab:re}
   \end{table}
    % \begin{minipage}{.4\textwidth}
\begin{table}[t]
\footnotesize
    \centering
     \caption{\textcolor{black}{Performances (F1-score) of different methods on benchmark datasets.}}
    \begin{tabular}{llllll}
    \toprule
    & i2b2 2010& DDI & PPI&ChemProt&ib2b 2012\\
    \midrule
    Wei et al~\cite{wei2019relation}&76.79&-&-&-&-\\
    Su et al~\cite{su2020investigation}&-&80.7&82.5&76.8&-\\
    Su et al~\cite{su2021improving}\footnote{\url{https://github.com/udel-biotm-lab/BERT-CLRE}}&-&\textbf{82.9}&\textbf{82.7}&\textbf{78.7}&-\\
    Chen et al~\cite{chen2019general}\footnote{\url{https://github.com/chentao1999/MedicalRelationExtraction}}&-&-&-&-&70.85\\
    %Lin et al~\cite{lin2020bert}\footnote{https://github.com/helloeve/mre-in-one-pass}&\\
    Guan et al~\cite{guan2020robustly}&-&-&-&-&70.5\\
    Sui et al~\cite{sui2020feded}&-&-&-&-&\textbf{75.09}\\
    \bottomrule
    \end{tabular}
    \label{tab:remethod}
% \end{minipage}
  \end{table} 
\subsubsection{Event Extraction}
Event extraction is another important task for mining structured knowledge from biomedical data, which aims to extract interactions between biological components (such as protein, gene, metabolic, drug, disease) and the consequences or effects of these interactions~\cite{ananiadou2010event}. 
Similar to BioRE, it is formulated into the multi-classification problem.
%The event generally consists of event triggers and their arguments (event participants). The triggers are signal words (generally verbs or nouns) to indicate the appearance and type of events. The arguments are biomedical entities. The events will finally be formulated to the graph structures, in which the triggers are connected with the appropriate argument along with the path. 
%Commonly used datasets are summarized in Table \ref{tab:ee}.
Many efforts have been proposed to explore the application of PLMs in biomedical event extraction recently. 
\textcolor{black}{
Trieu et al~\cite{trieu2020deepeventmine}\footnote{\url{ https://github. com/aistairc/DeepEventMine}} proposed the model called DeepEventMine with the BERT-based encoder, which significantly outperforms the strong baseline based on CNN. 
Wadden et al~\cite{wadden2019entity}\footnote{\url{https://github. com/dwadden/dygiepp}} explored combining the BERT model and graph propagation to capture long-range cross-sentence relationships, which have been proven to improve the performance of the model-based BERT alone. 
%Zhang et al~\cite{zhang2020pre} investigated transfer learning with the BERT model for Chinese clinical event detection. 
%There are works modeling the biomedical event extraction task as other NLP tasks. 
Ramponi et al~\cite{ramponi2020biomedical}\footnote{\url{https://github.com/cosbi-research/beesl}} modeled the biomedical event extraction as the sequence labeling problem, and proposed the model called BEESL with the BERT model as the encoder. 
It outperformed the baseline based on LSTM by 1.57\% in the GENIA 2011~\cite{kim2011overview} benchmark.
Wang et al~\cite{wang2020biomedical}\footnote{\url{https://github.com/WangXII/bio_ event_qa}} formulated the biomedical event extraction as the multi-turn question-answering problem and utilized the question-answering system based on the SciBERT.
The method can form event structures from the answers to multiple questions and achieves promising results on GENIA 2011~\cite{kim2011overview} and Pathway Curation 2013~\cite{pyysalo2015overview} dataset.
In Table \ref{tab:ee} and Table \ref{tab:eemethod}, we summarize commonly used datasets and compare the performance of different methods.
}
\begin{table} 
\begin{minipage}{.59\textwidth}
\scriptsize
    \centering
    \caption{Datasets used in the Biomedical event extraction.}
    \begin{tabular}{lllll}
    \toprule
    Dataset &  Entities & Triggers & Relations&Events\\
    \midrule
    Cancer Genetics 2013~\cite{nedellec2013overview}& 21,683&9,790&13,613&17,248\\
    EPI 2011~\cite{ohta2011overview}& 16,675&2,035&3,416&2,453\\
    GENIA 2011~\cite{kim2011overview}& 22,673&10,210&14,840&13,560\\
    GENIA 2013~\cite{kim2013genia}& 12,725&4,676&7,045&6,016\\
    Infectious Diseases 2011~\cite{pyysalo2011overview}&12,788&2,155&2,621&2,779\\
    Pathway Curation 2013~\cite{pyysalo2015overview}&15,901&6,220&10,456&8,121\\
    Multi-level event extraction~\cite{pyysalo2012event}& 8,291&5,554&7,588&6,677\\
    \bottomrule
    \end{tabular}
    \label{tab:ee}
    \end{minipage}
        \begin{minipage}{.38\textwidth}
        \scriptsize
    \centering
     \caption{\textcolor{black}{Performances (F1-score) of different methods on benchmark datasets.Genia means the GENIA 2011~\cite{kim2011overview} dataset. PC means the Pathway Curation 2013~\cite{pyysalo2015overview} dataset.}}
    \begin{tabular}{lllll}
    \toprule
    & Genia&PC&\\
    \midrule
  Trieu et al~\cite{trieu2020deepeventmine}&\textbf{63.96}&\textbf{55.67}\\
  %Wadden et al~\cite{wadden2019entity}&\\
  Ramponi et al~\cite{ramponi2020biomedical}&60.22&-\\
  Wang et al~\cite{wang2020biomedical}&58.33&48.29\\ 
    \bottomrule
    \end{tabular}
    \label{tab:eemethod}
\end{minipage}
  \end{table} 
\subsection{Text Classification}
Text classification aims to classify biomedical texts into pre-defined categories, which play an important role in the statistical analysis, data management, retrieval of biomedical data et al. 
%Compared with the general domain, text classification in the biomedical domain has more challenges, such as data unbalancing, semantic ambiguity, and irregular data. 
Fine-tuning pre-trained language models on biomedical text classification has attracted great attention recently.
%in which the commonly used biomedical text classification datasets are summarized in Table \ref{tab:tc}. 
\textcolor{black}{
Gao et al~\cite{gao2021limitations} investigated four methods of adapting the BERT model to handle input sequences up to approximately 400 words long, for the clinical single-label and multi-label clinical document classification. }
However, they found that the BERT or BioBERT model generally has equal or worse performance for clinical data such as the MIMIC-III clinical notes dataset, than a simple CNN model. They suggested that it may be because BERT or BioBERT models don't capture clinical domain knowledge due to trained on the general domain or biomedical literature datasets, and can't handle too long sentences longer than 512 tokens. 
Mascio et al~\cite{mascio2020comparative} made a comprehensive analysis of the performance of various word representation methods (such as Bag-of-Words, Word2Vec, GLoVe, FastText, BERT, BioBERT) and classification approaches (Bi-LSTM, RNN, CNN) on the electronic health records classification. They found that the contextual embeddings from BERT and BioBERT generally outperform the traditional embeddings, and the traditional deep neural networks Bi-LSTM enriched with appropriate entity information and specific domain embeddings have better performance than BERT and BioBERT. 
Guo et al~\cite{guo2020benchmarking} compared the performance of three PLMs including RoBERTa-base, BERTweet, and Clinical BioBERT on 25 social media classification datasets, in which 6 datasets are biomedical related. They found that RoBERTa-base and BERTweet outperform Clinical BioBERT, in which RoBERTa-base can capture general text semantic characteristics, while BERTweet captures more domain knowledge. 
Gutierrez et al~\cite{gutierrez2020document}\footnote{\url{https://github.com/dki-lab/covid19-classification}} also provided an analysis of traditional deep neural networks and fine-tuning PLMs including BERT and BioBERT on the performance of multi-label document classification on the COVID-19 dataset: LitCovid. 
They found that BERT and BioBERT models have better performance than traditional methods such as RNN, CNN, and Bi-LSTM in the datasets, and BioBERT outperforms BERT due to domain-specific pre-training. 

\paragraph{Summary}
{We summarize commonly used biomedical text classification datasets in Table \ref{tab:tc}, and show the performances of different methods in these datasets in Table \ref{tab:tcm}.
In summary, all these methods found that directly using fine-tuning PLMs outperforms the traditional neural network based methods.
The performance of different language models depends on the target datasets, for example, BERTweet pre-trained with large scale English Tweets, significantly outperforms ClinicalBioBERT on the social media dataset PM Abuse~\cite{guo2020benchmarking}.
BioBERT has promising performance on the covid-19 dataset but has worse performance than ClinicalBioBERT on the clinical data such as the MIMIC-III.}

%Sont et al~\cite{song2019classification} proposed to use the character-level BERT model to enhance the text convolutional neural network, for the classification of traditional Chinese medicine cases.
\begin{table}[t] 
% \begin{minipage}{.45\textwidth}
\footnotesize
    \centering
    \caption{{Datasets used in the biomedical text classification task.}}
    \begin{tabular}{llllll}
    \toprule
    Dataset &Label type &{Label Num}&{Avg Label Num}&Text type &Data Size\\
    \midrule
    HoC~\cite{hanahan2000hallmarks}&Multi-label&37&-&PubMed abstracts&1,852\\
{MeSH}~\cite{tsatsaronis2015overview}&Multi-label&26,563&12.55&Biomedical articles&10,876,004\\
    MIMIC-III~\cite{johnson2016mimic}&Multi-label&6,919&11.7&Discharge summaries&49,785\\
    LitCovid~\cite{gutierrez2020document}&Multi-label&8&-&PubMed articles&23,038\\
    CORD-19 Test~\cite{gutierrez2020document}&Multi-label&8&-&PubMed articles&100\\
PM Abuse~\cite{al2021text}&Multi-label&4&-&Tweets&15,100\\
    \bottomrule
    \end{tabular}
    \label{tab:tc}
\end{table}
% \begin{minipage}{.3\textwidth}
\begin{table}[t] 
\footnotesize
    \centering
    \caption{{Performances (accuracy for PM Abuse, macro F1 score for other datasets) of different methods on classification datasets.}}
    \begin{tabular}{llllll}
    \toprule
    Dataset &HoC&MiMIC-III&PM Abuse&LitCovid&CORD-19\\
    \midrule
    LSTM~\cite{adhikari2019rethinking}&-&-&-&83.9&83.2\\
CNN~\cite{kim2014convolutional}&-&-&-&83.3&82.7 \\
BERT~\cite{DevlinCLT19}&80.12&29.3&-&85.5&85.1\\
    BioBERT~\cite{LeeYKKKSK20}&81.54&32.4&-&86.3&86.2\\
    ClinicalBioBERT~\cite{alsentzer2019publicly}&-&44.4&77.4&-&-\\
BERTweet~\cite{nguyen2020bertweet}&-&-&82.4&-&-\\
    \bottomrule
    \end{tabular}
    \label{tab:tcm}
\end{table}
\begin{table}[t]
\footnotesize
    \centering
    \caption{Benchmark datasets in the biomedical sentence similarity task.}
    \begin{tabular}{llll}
    \toprule
    Dataset &Text type &Data Size\\
    \midrule
    BIOSSES~\cite{souganciouglu2017biosses}&PubMed&100\\
    MedSTS~\cite{wang2020medsts}&Clinical report&174,629\\ 
    MedSTS\_ann~\cite{wang2020medsts}&Clinical report&1,068\\
    n2c2/OHNLP~\cite{wang20202019}&Clinical report&1,642\\
    \bottomrule
    \end{tabular}
    \label{tab:ss}
\end{table}
\begin{table}[t]
% \begin{minipage}{.22\textwidth}
\footnotesize
    \centering
    \caption{\textcolor{black}{Performances (F1-score) of different methods on benchmark datasets.}}
    \begin{tabular}{lll}
    \toprule
    &BIOSSES & MedSTS\\
    \midrule
    Chen et al~\cite{chen2019biosentvec} & 84.8&83.6\\
    Chen et al~\cite{chen2019evaluation} &-& 83.8\\
    Chen et al~\cite{chen2020deep}&-&\textbf{85.28}\\
    \bottomrule
    \end{tabular}
    \label{tab:ssmethod}
% \end{minipage}
  \end{table} 
\subsection{Sentence Similarity}
The semantic similarity task is generally formulated into the regression problem to predict the similarity score of each sentence pair. Recent works have focused on fine-tuning various PLMs for this task.
%Existing pre-trained language models in the biomedical and general domains such as BERT, RoBERTa, BioBERT, SciBERT, ClinicalBERT, BlueBERT, PubMedBERT, and BioMegatron have achieved great performance in the task, in which the BioMegatron yielded the best performance due to domain-specific pre-training and task-specific fine-tuning. However, these models are pre-trained for yielding embeddings in the token level with the contextual information.
%To better capture the semantic information at the sentence level, 
Chen et al~\cite{chen2019biosentvec}\footnote{\url{https://github.com/ncbi-nlp/BioSentVec}} proposed the first pre-trained open set sentence embeddings in the biomedical domain called BioSentVec, which is trained on over 30 million documents from both biomedical literature such as PubMed and clinical notes such as the MIMIC-III Clinical Database. Compared with existing word embeddings and sentence encoder-based methods, it yields better performance on both sentence similarity and text classification tasks, due to better capturing the sentence-level semantic information. 
Chen et al~\cite{chen2019evaluation} empirically compared the performance of traditional deep learning methods such as random forest, RNN, CNN with PLMs including BERT and BioBERT, which shows that PLMs are more effective.
\textcolor{black}{Chen et al~\cite{chen2020deep} further show the BioSentVec can improve the performance of traditional deep learning models by 2\% F1 score.}
Yang et al~\cite{yang2020measurement} explored three PLMs including BERT, XLNet, and RoBERTa for the clinical semantic textual similarity task, in which the XLNet achieves the best performance among the three models. 
%Li et al~\cite{li2020cross2self} proposed to integrate the BERT models and bidirectional recurrent neural network (Bi-RNN) to capture both contextual semantic and semantic textual similarity. 
\textcolor{black}{We show commonly used sentence similarity datasets and compare performances of different methods in Table \ref{tab:ss} and Table \ref{tab:ssmethod}.
We found that Chen et al~\cite{chen2020deep} using the pre-trained sentence embedding BioSentVec with the traditional neural networks has better performance than Chen et al~\cite{chen2019evaluation} directly fine-tuning BERT and BioBERT.
}

\subsection{Question Answering}% \benyou{}
Biomedical question answering (BioQA) aims to extract or generate the natural language answers to the given questions, and generally be formulated into the machine reading comprehension approach focusing on predicting the text span of answers with the given questions and passages containing the answers.
%It is a challenging task due to the lack of large-scale annotated data in the biomedical domain, where annotating data requires domain expertise and time-consuming. 
%To facilitate the development of BioQA, some competitions and datasets have been proposed, such as BioASQ~\cite{tsatsaronis2012bioasq} and MEDIQA 2019\footnote{https://sites.google.com/view/ mediqa2019}.
Recently, the fine-tuning and transfer learning of PLMs have been widely explored in the task. 
%Much effort has investigated the BERT and BioBERT on the question answering task of biomedical literature datasets.
\textcolor{black}{Yoon et al~\cite{yoon2019pre}\footnote{\url{https://github.com/dmis-lab/bioasq-biobert}} applied the BioBERT to answer biomedical questions such as factoid, list, and yes-no type questions. 
They show that BioBERT fine-tuned with the question-answering datasets in both the general and biomedical domains and achieved the best performance in the 7th BioASQ Challenge.
%To solve the problem of limited training data in the biomedical domain, they first fine-tune BioBERT on the general domain question answering datasets SQuAD and SQuAD 2.0, and then further fine-tune it on the task dataset: BioASQ. 
Jeong et al~\cite{jeong2020transferability}\footnote{\url{https://github.com/dmis-lab/bioasq8b}} proposed to transfer the knowledge of natural language inference (NLI) to BioQA with BioBERT, which outperforms previous methods on Yes/No, Factoid, and List type questions by 5.59\%, 0.53\%, and 13.58\%, in the 7th BioASQ Challenge.
Chakraborty et al~\cite{chakraborty2020biomedbert}\footnote{\url{https://github.com/BioMedBERT/biomedbert}} proposed a novel language model BioMedBERT for question answering (QA) and information retrieval tasks, which is pre-trained on a large-scale biomedical literature dataset BREATHE based on BERT, and outperforms BERT in the BioQA.
Kamath et al~\cite{kamath2019pre} compared the effectiveness of PLMs based on two different QA models including the machine-reading comprehension and open question-answering method, and show the question-answering model achieves better performance on the BioQA. 
Du et al~\cite{du2020deep} utilized the BERT model as the encoder and then used the scaled dot-product attention mechanism to capture the interaction between the question and passage.
The proposed method outperforms the best performance for factoid questions in 2016 and 2017 BioASQ-Task B.
Zhou et al~\cite{zhou2019dut} utilized the BioBERT and interactive transformer model for both the recognizing question entailment and question answering task, and showed significant improvements on the single task with the shared representations of both tasks. 
Similarly, Akdemir et al~\cite{akdemir2020transfer} also explored multi-task learning to improve the performance of BioBERT on the BioQA task with the biomedical entity recognition task, and show its improvements on the BioASQ 8B challenge.
However, these models can't detect multiple spans of the passage when there are multiple answers to the question. 
To solve the problem, Yoon et al~\cite{yoon2021sequence}\footnote{\url{https://github.com/dmis-lab/SeqTagQA}} reformulated the BioQA task as the sequence tagging problem to detect multiple entity spans simultaneously based on the BioBERT encoder, which achieves the BioASQ 7b and 8b list-type questions.}

Some works tried to incorporate domain knowledge, such as biomedical-named entities, into PLMs.
\textcolor{black}{He et al~\cite{he2020infusing}\footnote{\url{https://github.com/heyunh2015/diseaseBERT}} proposed to infuse the domain knowledge of disease into a series of PLMs including BERT, BioBERT, SciBERT, ClinicalBERT, BlueBERT, and ALBERT, to improve their performance.
They found all these models can be improved by infusing the disease knowledge, and for example, the accuracy of BioBERT on the CHQ dataset can be improved by nearly 4\%.
Rawat et al~\cite{rawat2020entity}\footnote{\url{https://github.com/emrQA/bionlp_acl20}} incorporated the medical entity information with entity embeddings and the auxiliary task on predicting the logical form of the question to improve the accuracy and generalization of the BERT model on answering questions, which improves the BERT model by 5\% F1 score on the paraphrased question answering of the emrQA dataset.
Kommaraju et al~\cite{kommaraju2020unsupervised} introduced the extra biomedical named entities prediction task to improve the BioBERT on Biomedical QA.
They show the BioBERT pre-trained by the prediction task outperforms the previous best model on the 7b-Phase B of the 7th BioASQ Task challenge.}
%To help the prevention of COVID-19, there are works~\cite{lee2020answering,esteva2020co,reddy2020end,ngai2021transformer} build the question answering and information retrieval system based on BioBERT and BERT.

Besides methods for biomedical literature corpora, other works have proposed question-answering models for unstructured electronic health records (EHR).
Soni et al~\cite{soni2020evaluation} investigated the performance of various PLMs including BERT, BioBERT, ClinicalBERT, and XLNet on the clinical question answering, and explored the fine-tuning methods with different datasets, including datasets in the general domain, biomedical and clinical corpora. 
\textcolor{black}{They find that fine-tuning the open-domain dataset SQuAD consistently improves the performance across all the model variants.
Mairittha et al~\cite{mairittha2020improving} explored four different fine-tuned BERT models for personalized EHR question answering and show the extended BioBERT-QA model pre-trained on unstructured EHR data achieves the best performance.
Table \ref{tab:QA} shows commonly used datasets in BioQA 
, and Table \ref{tab:QAM} presents the performances of different methods on these datasets.
Several methods~\cite{yoon2019pre,soni2020evaluation,mairittha2020improving} have shown fine-tuning PLMs with the open domain question-answering dataset, and pre-training PLMs with the in-domain datasets improves the performance of various PLMs.
Generally, it assumes that fine-tuning or pre-training with more corpora is always useful, for example, Soni et al~\cite{soni2020evaluation} fine-tuning PLMs with the general domain, biomedical and clinical corpora achieves the best performance for the clinical question answering.
Moreover, incorporating the domain knowledge including disease knowledge, medical entities, and multi-task learning incorporating BioNER task, can significantly improve the performance.
For example, He et al~\cite{he2020infusing} improved the accuracy of BioBERT on the MEDIQA dataset by nearly 4\% with the disease knowledge and achieved the best performance.
}
\begin{table}[t] 
% \begin{minipage}{.28\textwidth}
\footnotesize
    \centering
    \caption{Datasets used in the biomedical question answering task.}
    \begin{tabular}{lll}
    \toprule
    Dataset &Text type & Data Size\\
    \midrule
    PubMedQA~\cite{jin2019pubmedqa} & PubMed abstracts & 1,000\\
    BioASQ~\cite{nentidis2019results} & MEDLINE articles & 885\\
    MEDIQA~\cite{abacha2019overview}&online community&383\\
    emrQA~\cite{pampari2018emrqa}&Clinical notes&400,000\\
    cMedQA~\cite{zhang2018multi}&online community&61,343\\
    COVID19-QA~\cite{tang2020rapidly}&Literature review&124\\
    \bottomrule
    \end{tabular}
    \label{tab:QA}
    \end{table}
   % \begin{minipage}{.68\textwidth}
   \begin{table}[t]
   \footnotesize
    \centering
    \caption{\textcolor{black}{Performances of different methods. For BioASQ 6b, we compare the Mean Reciprocal Rank (MRR) score on the Factoid question. For the BioASQ, we compare the averaged MRR score on the Factoid question of all batches. For the MEDIQA and emrQA, we compare the accuracy score.}}
    \begin{tabular}{lllll}
    \toprule
     &BioASQ 6b & BioASQ 8b&MEDIQA&emrQA\\
    \midrule
    Yoon et al~\cite{yoon2019pre}&48.41&-&-&-\\
    Jeong et al~\cite{jeong2020transferability}&48.05&\textbf{46.65}&-&-\\
    Chakraborty et al~\cite{chakraborty2020biomedbert}&\textbf{50.50}&-&-&-\\
   Kamath et al~\cite{kamath2019pre}&45.70&-&-&-\\
   Zhou et al~\cite{zhou2019dut}&-&-&75.8&-\\
  Akdemir et al~\cite{akdemir2020transfer}&-&43.61&-&-\\
  Yoon et al~\cite{yoon2021sequence}&-&37.95&-&-\\
  He et al~\cite{he2020infusing}&-&-&\textbf{79.49}&-\\
  Rawat et al~\cite{rawat2020entity}&-&-&-&59.00\\
  Kommaraju et al~\cite{kommaraju2020unsupervised}&-&43.93&-&-\\
  Soni et al~\cite{soni2020evaluation}&-&-&-&75.56\\
   Soni et al~\cite{soni2020evaluation}&-&-&-&\textbf{86.97}\\
    \bottomrule
    \end{tabular}
    \label{tab:QAM}
   \end{table}
  % \end{table} 
  
\subsection{Dialogue Systems}
% \benyou{retrieval or generation}

%\acfp{DS} in the biomedical domain has attracted continuous attention due to the utility applications (\textit{e.g.}, virtual health consultants and therapists)~\cite{laranjo2018conversational}.
The dialogue system aims to produce a proper response in either a selective~\cite{zhang2020mie,whang2020effective} or generative~\cite{liu2020meddg,zeng2020meddialog,zhang2020dialogpt} way given a dialogue context for the biomedical goals of a user.
The context includes historical utterances from users and systems, biomedical knowledge base, electronic health records of users, etc.
The format of a response could be various, \textit{e.g.}, a set of structured user goal data~\cite{wei2018task}, a distribution of biomedical labels for diagnosis~\cite{lin2019enhancing,zhang2020mie} and natural language utterances~\cite{zeng2020meddialog}.
For different types of contexts and responses, recent work focuses on end-to-end \ac{DS}~\cite{xu2019end,zeng2020meddialog} or parts of four typical \ac{DS} modules, \textit{i.e.}, \acf{NLU}~\cite{du2019learning,shi2020understanding}, \acf{DST}~\cite{lin2019enhancing,wei2018task}, \ac{DPL}~\cite{wei2018task,xia2020generative} and \ac{NLG}~\cite{zeng2020meddialog}.
Recently, \acsp{PLM} are well-known for natural language modeling, but it is nontrivial to pre-train on task datasets that are based on a specific domain~\cite{whang2020effective}.
%Pre-training \ac{DS} models in biomedical can be seen as a task-specific pre-training problem~\cite{gururangan2020don}.
%It involves two essential aspects, \textit{i.e.}, biomedicine-domain adaptation and dialogue-task adaptation. 
\textcolor{black}{To adapt PLMs to the medical domain, the dominant solution is to pre-train a language model on a large-scale general/medical corpus and then fine-tune the model with a medical dialogue dataset.
Yan et al~\cite{yan2022remedi}\footnote{\url{https://github.com/yanguojun123/Medical-Dialogue}} first explored fine-tuning PLMs including BER-WWM, BERT-MED, MT5 and GPT2 on $M^2$-MedDialog dataset for understanding the intents and slots of patients, in which MT5 achieves the best performance.
%BioBERT and MIMIC-BERT~\cite{wei2018task} are pre-trained using MIMIC III dataset~\cite{si2019enhancing} and PubMed articles, respectively, followed by fine-tuning on MZ dataset for predicting diagnosis actions.
Zeng et al~\cite{zeng2020meddialog}\footnote{\url{https://github.com/UCSD-AI4H/Medical-Dialogue-System}} pre-trained Transformer, BERT-GPT, and GPT on dialog datasets and other large-scale texts, and then fine-tune models on the Chinese MedDialog dataset for generating clinically correct and human-like medical responses.
BERT-GPT has been shown to have lower perplexity compared to both Transformer and GPT, while maintaining similar diversity metrics as Transformer.}
%Naturally, \acsp{PLM} are talented for those challenges because: 
%(i) \acsp{PLM} can capture longer-term dependency with the transformer architecture for learning complex dialogues effectively and efficiently~\cite{henderson2020convert,vaswani2017attention}.
%Yan et al~\cite{yan2022remedi} unifies \acs{NLU}, \acs{DPL}, \acs{NLG} tasks into one context-to-response generation framework and use pretrained GPT2 and MT5 to model complex context for generating responses.
%The empirical study proves the positive impact of \acsp{PLM} on $M^2$-MedDialog dataset.
%(ii) \acsp{PLM} can incorporate external knowledge by pre-training on large-scared corpora~\cite{kalyan2021ammu,li2020task}.
\textcolor{black}{
Shi et al~\cite{shi2020understanding}\footnote{\url{https://github.com/xmshi-trio/MSL}} show BERT has promising performance on the medical slot-filling task, and pre-trained embedding from BERT can further improve the performance of the weak supervision method.
DialoGPT~\cite{zhang2020dialogpt}\footnote{\url{https://github.com/microsoft/DialoGPT}} is pre-trained based on GPT-2~\cite{radford2018improving} with a large in-domain dialogue dataset, and is able to generate more relevant, informative and coherent responses compared with the strong baseline based on the sequence to sequence model.
Li et al~\cite{li2020task}\footnote{\url{https://github.com/lockon-n/dapo}} proposed the dialogue-adaptive pre-training objectives (DAPO) by considering dialogue-specific features including coherence, specificity, and informativeness, which shows better performance than other language modeling objectives such as MLM and NSP.}
% Although recent studies have deployed \acsp{PLM} models in medical \acp{DS} tasks, medical \ac{DS} is still under-explored.

\textcolor{black}{
\paragraph{Summary}
We summarize all available biomedical dialogue datasets in Table~\ref{tab:DS}.
Different from using the accuracy, recall, and F1 score metrics used by previous tasks, the dialogue system task generally uses the machine translation metrics including BLEU~\cite{zhang2020dialogpt}, METEOR~\cite{banerjee2005meteor}, and NIST~\cite{doddington2002automatic}, to measure the similarity between generated responses and the ground truth based on n-gram matching.
These metrics for evaluating generated responses are limited in that they only take into account shallow lexical overlaps and do not account for paraphrasing and terminology variations. 
To address this, some automatic metrics based on pre-trained language models have been developed, such as BERTScore~\cite{zhangbertscore}, which calculates the similarity between two sentences using contextual embeddings from PLMs. 
However, these metrics have been shown to be inadequate in evaluating the faithfulness of generated responses.
While there have been efforts to develop factual consistency metrics like BARTScore~\cite{yuan2021bartscore} in the general domain, there has been less focus on developing such metrics in the biomedical domain to evaluate factual correctness.
Since the aforementioned methods utilized different datasets, it is hard to compare their performances directly.
In summary, they have demonstrated that creating more effective pre-training tasks, incorporating task-specific information, and pre-training with large in-domain dialogue datasets are effective strategies for improving the performance of series PLMs.
}
% for future research.
\begin{table} 
    \scriptsize
    \centering
    \setlength{\tabcolsep}{1pt}
    \caption{Datasets used in the biomedical dialogue system tasks.}
    \begin{tabular}{llllll}
    \toprule
    Dataset & Language & Domain & Evaluated Task & Text type & \# dialogues\\
    \midrule
    MZ~\cite{wei2018task}                   & EN & Pediatrics & \acs{DPL} & Discharge summaries &  710\\
    DX~\cite{xu2019end}                     & CN & Pediatrics & \acs{DPL} & Patient-doctor dialogues \& patient reports &  527\\
    RD~\cite{liao2020task}                  & CN & Pediatrics & \acs{DPL} & Patient-doctor dialogues \& patient reports & 1,490\\
    SD~\cite{liao2020task}                  & CN & 9 domains  & \acs{DPL} & Patient-doctor dialogues \& patient reports & 30,000\\
    CMDD~\cite{lin2019enhancing}            & CN & Pediatrics & \acs{NLU} & Patient-doctor dialogues &  2,067\\
    SAT~\cite{du2019extract}                & CN & 14 domains & \acs{NLU} & Patient-doctor dialogues & 2,950\\
    MSL~\cite{shi2020understanding}         & CN & Pediatrics & \acs{NLU} & Patient-doctor dialogues & 1,652 \\
    MIE~\cite{zhang2020mie}                 & CN & Cardiology & \acs{NLU} & Patient-doctor dialogues &   1,120\\
    CovidDialog~\cite{yang2020generation}   & CN/EN & COVID-19 & \acs{NLG} & Patient-doctor dialogues & 1,088/603 \\
    MedDG~\cite{liu2020meddg}               & CN & Gastroenterology & \acs{NLG} & Patient-doctor dialogues &  17,000\\
    MedDialog~\cite{zeng2020meddialog}      & CN/EN & 29 domains & \acs{NLG} & Patient-doctor dialogues \& patient reports &  3,407,494/257,332 \\ 
    Chunyu~\cite{lin2021graph}              & CN & - & \acs{NLG} & Patient-doctor dialogues  & 12,842 \\
    KaMed~\cite{li2021semi}                 & CN & 12 domains & \acs{NLG} & Patient-doctor dialogues & 63,754\\
    $M^2$-MedDialog-base~\cite{yan2022remedi}   & CN & 30 domains & \acs{NLU}\&\acs{DPL}\&\acs{NLG} & Patient-doctor dialogues \& patient reports & 1,557\\
    $M^2$-MedDialog-large~\cite{yan2022remedi}   & CN & 40 domains & \acs{NLG} & Patient-doctor dialogues \& patient reports & 95,408\\
    \bottomrule
    \end{tabular}
    \label{tab:DS}
\end{table} 

\subsection{Text Summarization}
Automatic text summarization aims to automatically summarize the key information of single or multiple documents with shorter and more fluent texts, which greatly decreases the time-consuming of acquiring important information. Similar to the general domain, existing methods can generally be classified into two categories: extractive summarization methods and abstractive summarization methods. 
%The former methods extract correlated sentences from given long documents and concatenate them into the final summary, while the latter methods generate new sentences based on the information of given long documents. Therefore, the extractive summarization generally is formulated into the binary classification that aims to predict  whether the sentences should be selected into the summary, while the abstractive summarization can be deemed as the conditional text generation problem.

To explore the advanced PLMs in the text summarization of the biomedical domain, the domain knowledge is incorporated by existing methods via domain fine-tuning~\cite{luo2023citationsum,xie2023factreranker}. 
For biomedical extractive summarization, Du et al~\cite{du2020biomedical} proposed a novel model BioBERTSum, which used the domain-aware pre-trained language model as the encoder and then fine-tuned it on the biomedical extractive summarization task.
\textcolor{black}{It outperforms SOTA extractive methods such as BERTSum.
Xie et al~\cite{xie2022pre}\footnote{\url{https://github.com/xashely/KeBioSum}} proposed the knowledge infusion training framework to incorporate medical knowledge to improve a series of PLMs including BERT, RoBERTa, BioBERT, and PubMedBERT.
The PubMedBERT-based method has the best performance and outperforms other strong baselines such as BERTSum and MatchSum.
Gharebagh et al~\cite{gharebagh2020attend} utilized the domain knowledge: salient medical ontological terms to help the content selection of the SciBERT-based clinical abstractive summarization model, which improves SOTA results by around 2\% in ROUGE metrics on two medical datasets MIMIC-CXR~\cite{johnson2019mimic} and OpenI~\cite{demner2016preparing}.
%Moradi et al~\cite{moradi2019clustering}\footnote{\url{https://github.com/BioTextSumm/BERT-based-Summ}} proposed the unsupervised extractive summarization with the hierarchical clustering algorithm to group the contextual embedding of sentences based on the BERT encoder and select the most informative sentences from each group to generate the final summary.
%It shows better performance than traditional unsupervised methods such as TexRank.
%Padmakumar et al~\cite{padmakumar2021unsupervised}\footnote{\url{ https: //github.com/vishakhpk/mi-unsup-summ}} proposed an unsupervised extractive summarization model, which used the GPT-2 to encode the sentences and the pointwise mutual information (PMI) to calculate the semantic similarity between sentences and documents. The proposed method has better performance than other similarity-based models on the medical journal dataset. 
Bishop et al~\cite{bishop2022gencomparesum}\footnote{\url{https://github.com/jbshp/gencomparesum}} proposed unsupervised extractive summarization method for biomedical literature with T5 and BERTScore, which achieves better performance than strong supervised methods such as BERTSum.
Xie et al~\cite{xie2022gretel}\footnote{\url{https://github.com/xashely/GRETEL_ extractive}} incorporated the neural topic model with hierarchical transformer encoder (HTE) based on PLMs, which significantly improved the performance of RoBERTa on long biomedical document summarization.
}
%Kanwal et al~\cite{kanwal2021attention} proposed to fined-tune the BERT model on the International Classification of Diseases (ICD-9) labeled MIMIC-III discharge notes for the extractive summarization of electronic health records.

For abstractive summarization, Wallace et al~\cite{wallace2020generating} utilized the Bidirectional and Auto-Regressive Transformers (BART) as the encoder for generating biomedical evidence summary of multiple clinical trials.
\textcolor{black}{They found that the summarizers can produce fluent and relevant synopses, but  the factual accuracy can't be guarantee.
Deyoug et al~\cite{deyoung2021ms2}\footnote{\url{https://github.com/allenai/ms2/}} investigated the BART model for the multi-document summarization on medical studies, which can generate coherent summaries that align with the reference summaries in evidence direction approximately 50\% of the time.
%To facilitate the development of methods for generating plain summaries toward the general public, 
Guo et al~\cite{guo2020automated}\footnote{\url{https://github.com/qiuweipku/Plain_language_summarization}} proposed a novel task of plain language summarization task on the biomedical scientific reviews, and explored pre-training BART model on general domain dataset CNN/DM and in-domain PubMed dataset. 
They found BART pre-trained using CNN/DM and PubMed abstracts demonstrate the strongest ROUGE scores, whereas the BART model pre-trained only using PubMed abstracts has the lowest level of readability.
Luo et al~\cite{luo2022readability}\footnote{\url{http://www.nactem.ac.uk/readability/}} proposed the new task of readability controllable summarization for biomedical documents, and explored the language model Longformer-Encoder-Decoder (LED) with the advanced controllable techniques including prompts and multi-head.
They demonstrate that the method can generate fluent summaries, but it lacks the capability to effectively control for readability.
Hu et al~\cite{hu2022graph} incorporated the additional knowledge with graph encoder and contrastive learning, to enhance the performance of the BioBERT.
The proposed method achieves state-of-the-art results in radiology report summarization.}
For the information acquisition of COVID-19 related scientific literature, Kieuvongngam et al~\cite{kieuvongngam2020automatic} proposed the BERT and GPT-2 based model for both extractive and abstractive summarization of COVID-19 research literature.
There are also works to build the multi-document summarization system for the information retrieval of COVID-19 research literature with the Siamese-BERT~\cite{esteva2020co}, BioBERT, and XLNet~\cite{dan2020caire}.

\textcolor{black}{Similar to the dialogue system task, the commonly used automatic metrics in the text summarization task including ROUGE~\cite{lin2004rouge}, and BERTScore~\cite{zhangbertscore}, usually evaluate the relevance and similarity between the generated summaries and the gold summaries. Moreover, the factuality metrics have attracted much attention recently to evaluate the factual correctness of generated summaries~\cite{xie2023faithful,luo2023chatgpt,xie2023factreranker}. Deyoung et al~\cite{deyoung2021ms2} introduce the $\Delta$EI metric to determine the degree of the factual accuracy of generated summaries in relation to the input medical studies.
Zhang et al~\cite{zhang2020optimizing} introduced the ChexBERT F1 score to evaluate the factual correctness between generated summaries and gold summaries of radiology reports.
In Table \ref{tab:TS}, we summarize datasets used in the biomedical text summarization task, and report performances of different methods on these datasets with the aforementioned evaluation metrics in Table \ref{tab:TSME}.
We can find that the method incorporating the domain knowledge~\cite{xie2022pre} has better performance than directly fine-tuning PLMs~\cite{du2020biomedical}, and the method~\cite{xie2022gretel} for long biomedical text summarization achieves the best performance on two biomedical literature datasets PubMed and CORD-19.
}
\begin{table}[t] 
% \begin{minipage}{.34\textwidth}
% \scriptsize
    \centering
    \caption{Datasets used in the biomedical text summarization.}
    \begin{tabular}{llll}
    \toprule
    Dataset &Text type & Type & Data Size\\
    \midrule
    COVID-19~\cite{wang2020cord} & Biomedical literature &Single& - \\
    MSˆ2~\cite{deyoung2021ms2} & Biomedical literature& Multi&470,402\\
    CDSR~\cite{guo2020automated} & Biomedical literature&Single & 7,805\\
    RCT~\cite{wallace2020generating} & Clinical trials&Multi&4,528\\
    PubMed~\cite{cohan2018discourse} &Biomedical literature &Single &119,924\\
    MIMIC-CXR~\cite{johnson2019mimic}&Radiology reports&Single&124,577\\
    OpenI~\cite{demner2016preparing}&Radiology reports&Single&3,599\\
    Readibility~\cite{luo2022readability}&Biomedical literature&Single&28,124\\
    \bottomrule
    \end{tabular}
    \label{tab:TS}
% \end{minipage}
\end{table}
\begin{table}[t] %{.65\textwidth}
% \scriptsize
    \centering
    \caption{\textcolor{black}{Performances (ROUGE-L score, counting the longest common subsequence (LCS) between the generated summary and the reference summary.) of different methods.}}
    \begin{tabular}{llllll}
    \toprule
    &PubMed &CORD-19&MSˆ2& RCT&MIMIC-CXR\\
      \midrule
    Du et al~\cite{du2019extract}&29.58&-&-&-&-\\
    Xie et al~\cite{xie2022pre}&33.28&29.10&-&-&-\\
    Bishop et al~\cite{bishop2022gencomparesum}&35.65&33.35&-&-&-\\
    Xie et al~\cite{xie2022gretel}&\textbf{38.61}&\textbf{40.01}&-&-&-\\
    Wallace et al~\cite{wallace2020generating} &-&-&-&\textbf{0.265}&-\\
    Deyoug et al~\cite{deyoung2021ms2}&-&-&20.80&0.1760&-\\
    Hu et al~\cite{hu2022graph}&-&-&-&-&46.65\\
    \bottomrule
    \end{tabular}
    \label{tab:TSME}
% \end{minipage}
\end{table} 
\subsection{Natural Language Inference}
Natural language inference (NLI, also known as text entailment) is a basic task for the natural language understanding of biomedical texts. 
It aims to infer the relation such as entailment, neutral and contradiction, between two sentences, named as the premise and hypothesis, which can further benefit biomedical downstream tasks such as commonsense comprehension, question answering and evidence inference. 
%In this task, the common neural network model is based on sentence pair modeling, which encodes the premise and hypothesis sentences with various neural networks and then classifies the relation between them with the softmax classifier layer. 

%Similar to other tasks, pre-trained language models in the biomedical domain, including BioELMo~\cite{jin2019probing} and BlueBERT~\cite{peng2019transfer}, have shown their effectiveness in the task via task-guided fine-tuning. 
To facilitate the development of methods for text inference and entailment in the medical domain, participants in the MEDIQA 2019 shared task~\cite{abacha2019overview} investigated the SciBERT, BioBERT, and ClinicalBERT in the medical NLI task.
\textcolor{black}{Among these participants, Wu et al~\cite{wu2019wtmed}\footnote{\url{https://github.com/ZhaofengWu/MEDIQA_WTMED}} achieves the best performance with 98\% accuracy in the REQ dataset~\cite{abacha2016recognizing}, which ensembled results of different base models and incorporated the syntax information.
Sharma et al~\cite{sharma2019incorporating}\footnote{\url{https://github.com/soummyaah/KGMedNLI}} incorporated the embedding of knowledge graph (UMLS) into the BioELMo to improve its performance, which shows an improvement of 0.8\% regarding the accuracy to the base BioELMo model.
Yadav et al~\cite{yadav2020medical}\footnote{\url{https://github.com/VishalPallagani/Medical-Knowledge-enriched-Textual-Entailment}} a novel framework Sem-KGN for the medical textual entailment task, which infused the medical entity information from the medical knowledge bases into the BERT model. 
They show the medical knowledge information improves the SOTA language model ClinicalBERT by 8.27\% on the REQ dataset.
He et al~\cite{he2020infusing}\footnote{\url{https://github.com/heyunh2015/diseaseBERT}} proposed to infuse the domain knowledge of disease into a series of PLMs including BERT, BioBERT, SciBERT, ClinicalBERT, BlueBERT, and ALBERT, which improves performances of these models in all cases.
Zhu et al~\cite{zhu2021discovering} utilized the neural architecture search (NAS) to automatically find a better transformer structure for language models, which improves the performance of the Chinese BERT-wwm-ext model~\cite{cui2019pre} on two Chinese NLI datasets.
We summarize all available datasets in Table \ref{tab:NLI}, and compare performances of different methods in Table \ref{tab:NLIMH}.
We can find that Wu et al~\cite{wu2019wtmed} using the ensemble method significantly outperforms other methods in RQE.
Among various PLMs including BioELMo, BERT, BioBERT, SciBERT, ClinicalBERT, BlueBERT, and ALBERT, ALBERT achieved the best performance on the MedNLI dataset.
}
\begin{table}[t] 
% \begin{minipage}{.65\textwidth}
% \scriptsize
    \centering
    \caption{Datasets used in the biomedical natural language inference.}
    \begin{tabular}{llll}
    \toprule
    Dataset & Text type & Relation Type & Data Size\\
    \midrule
   MedNLI~\cite{romanov2018lessons} & Clinical notes & Entailment, contradiction, or neutral &14,049\\
   RQE~\cite{abacha2016recognizing} & Consumer health questions & Entailment, contradiction&9,120\\
   CMFAQ~\cite{zhu2021discovering} & Consumer health questions & Entailment, contradiction&53,822\\
    \bottomrule
    \end{tabular}
    \label{tab:NLI}
\end{table}
\begin{table}
% \begin{minipage}{.34\textwidth}
% \scriptsize
    \centering
    \caption{\textcolor{black}{Performances (accuracy for MedNLI and REQ, F1 score for CMFAQ) of different methods.}}
    \begin{tabular}{llll}
    \toprule
   & MedNLI & RQE & CMFAQ\\
    \midrule
  Wu et al~\cite{wu2019wtmed}&-&\textbf{98.00}&-\\
  Sharma et al~\cite{sharma2019incorporating}&79.04&-&-\\
  Yadav et al~\cite{yadav2020medical}&-&56.17&-\\
  He et al~\cite{he2020infusing}&\textbf{79.49}&-&-\\
  Zhu et al~\cite{zhu2021discovering}&-&-&88.9\\
    \bottomrule
    \end{tabular}
    \label{tab:NLIMH}
% \end{minipage}
  \end{table} 

\subsection{Proteins/DNAs Prediction}

In this section, we only list some applications that have been well-investigated or have potential, although there are much bigger spaces in biomedical domains to make use of PLMs.
\subsubsection{Protein structure predictions}

Proteins are essential to life, and knowing their structure can facilitate  our  understanding of their function. However, the structure of only a small fraction of proteins is known \cite{jumper2021highly}. Predicting the 3D structure of a protein is based solely on its amino acid sequence, a.k.a, `protein folding problem' \cite{anfinsen1973principles}. To evaluate  protein structure predictions,  CASP  (Critical Assessment of Structure Prediction)  uses proteins with recently solved structures that have not been deposited in the PDB or publicly disclosed; it therefore, is a blind test for the participators, which is the gold-standard assessment for protein structure predictions \cite{moult1995large,kryshtafovych2019critical}. In CASP14, AlphaFold 2~\cite{jumper2021highly}, a model designed by DeepMind achieves much better performance than other participating methods (\textit{e.g.} template-based methods). The authors claim that  AlphaFold 2 could provide precise estimates and could be confidently used for protein structure predictions with high reliability. However, predictions of existing methods, including the AlphaFold 2 are more family-specific than protein-specific, and rely on the evolutionary information captured in multiple sequence alignments (MSAs). To solve these issues, cite{weissenow2021protein} proposed to use the attention head from the pre-trained protein language model ProtT5 without MSAs. Recently, Sturmfels et al~\cite{sturmfels2020profile} presented a new biologically-informed pre-training task:  predicting protein profiles derived from multiple sequence alignments, which can improve the downstream protein structure prediction task.

% \subsubsection{Protein synthesis}
\subsubsection{DNA related applications}
There are few works in DNA pre-training, among which DNABERT \cite{Ji2020} is the representative one. 
DNABERT not only achieved SOTA performance on promoter prediction, splice sites
and transcription factor binding sites, but also  identify functional genetic variants.
% DNABERT claims that `DNABERT could 1) effectively predicts proximal and core promoter regions; 2) accurately identifies transcription factor binding sites; 3)  allows visualization of important regions, contexts and sequence motifs; 4) identify functional genetic variants with DNABERT; 5)  substantially enhances performance and generalizes to other organisms.' 
Hong et al~\cite{hong2020identifying} proposed to pre-train DNA vectors to encode enhancers and promoters, and then Incorporated the attention mechanism to predict long-range enhancer–promoter interactions (EPIs). Yamada et al~\cite{yamada2021prediction} proposed a novel method based on the BERT to predict the interactions between RNA sequences and RNA-binding proteins (RBPs), in which  BERT  is pre-trained on the human reference genome. Mock et al~\cite{mock2021bertax} presented the BERTax based on BERT, for the taxonomic classification of DNA sequences.

% \qq{@benyou: 5.8.2 copy/paste a long portion of text from another article, this actions is at the edge of plagiarism. I suggest the authors to rephrase that paragraph.}

\subsection{Competitions and  Venues}
To facilitate the technological developments in biomedical text mining, many shared tasks and competitions have been organized since several years ago, focusing on various important tasks in the biomedical domain. 
\begin{itemize}

    % \item  \benyou{BLUE} BLUE benchmark consists of five different biomedicine text-mining tasks with ten corpora. Here, we rely on preexisting datasets because they have been widely used by the BioNLP community as shared tasks. These tasks cover a diverse range of text genres (biomedical literature and clinical notes), dataset sizes, and degrees of difficulty and, more importantly, highlight common biomedicine text-mining challenges.
    % \item \benyou{CBLUE}
    % \item  \benyou{BLURB leaderboard https://microsoft.github.io/BLURB/leaderboard.html}
    \item \textbf{BioNLP workshop.} The BioNLP workshop\footnote{\url{https://aclweb.org/aclwiki/SIGBIOMED}} has been organized for 20 years and continually promoted the development of the biomedical domain, in which the community proposed a series of shared tasks and benchmark datasets. In BioNLP 2019, the BioNLP Open Shared Tasks (BioNLP-OST) 2019~\cite{jin2019proceedings} and the MEDIQA 2019 Shared Task~\cite{abacha2019overview} were organized. The BioNLP-OST 2019 proposed six tasks, including the information extraction on the bacterial biotopes and phenotypes, event extraction of genetic and molecular mechanisms, pharmacological substances, compounds and proteins named entity recognition, integrated structure, semantics and coreference task, concept extraction for drug repurposing, and the information retrieval task for neuroscience. The MEDIQA 2019 aims to explore the method development on the natural language inference (NLI), recognizing question entailment (RQE), and question answering (QA) in the medical domain. In bioNLP 2021, the MEDIQA 2021~\cite{abacha2021overview} shared tasks have three tasks related to the summarization of medical documents, including the question summarization task, the multi-answer summarization task, and the radiology report summarization task. 

\item \textbf{BioNLP-OST.} The BioNLP Open Shared Tasks (BioNLP-OST)\footnote{https://2019.bionlp-ost.org/home} has been proposed since 2009 and was motivated to facilitate the development and sharing of methods on various tasks of biomedical text mining. It is organized every two years and organized at different conferences such as BioNLP and EMNLP. The latest BioNLP-OST 2019 is organized at the BioNLP 2019 as introduced aforementioned.

\item \textbf{BioASQ.} The BioASQ\footnote{http://bioasq.org} organizes workshops and challenges on biomedical semantic indexing and question answering. It has been held annually since 2013. In BioASQ 2019, the large-scale biomedical semantic indexing task, the biomedical information retrieval and question-answering task, and corresponding benchmark datasets are proposed. 

\item \textbf{BioCreAtIvE.} The Critical Assessment of Information Extraction systems in Biology (BioCreAtIvE)\footnote{https://biocreative.bioinformatics.udel.edu} organized challenge evaluations for the text mining and information extraction method on the biological domain since 2004. The latest BioCreative VII Challenge proposed five tracks, of which two tracks are related to COVID-19, including text mining and multi-label topic classification.

\item {\textbf{TREC.} The Text REtrieval Conference (TREC\footnote{\url{https://trec.nist.gov}}) organizes workshops for supporting the development of information retrieval methods based on large test collections. It was started in 1992 and held annually. It has biomedical tracks focusing on clinical decision support, precision medicine, and clinical trials et al.
}

\item \textbf{eHealth-KD.} The eHealth-KD\footnote{https://knowledge-learning.github.io/ehealthkd-2019/} organizes challenges on the structure knowledge extraction of eHealth documents in the Spanish Language. The eHealth-KD Challenge 2019 proposed the key phrases identification and classification task, and the semantic relations detection task.

\item \textbf{\#SMM4H.} The Social Media Mining for Health Applications (\#SMM4H)\footnote{https://healthlanguageprocessing.org/smm4h-2021/} held workshops and shared tasks related to natural language processing challenges in social media data for health research since 2015 annually. The shared tasks in the \#SMM4H ’21 involve the information processing methods on Twitter related to COVID-19, self-report of breast cancer, adverse effect mentions, medication regimen, and adverse pregnancy outcomes.
\end{itemize}

Moreover, there are some challenges proposed recently, such as the COVID-19 Open Research Dataset Challenge (CORD-19)\footnote{https://www.kaggle.com/allen-institute-for-ai/CORD-19-research-challenge} in response to the COVID-19 pandemic, EHR DREAM Challenge\footnote{\url{https://www.synapse.org/\#!Synapse:syn18405991/wiki/589657}} proposed in October 2019 and focusing on using electronic health record data to predict patient mortality, and ICLR 2021 workshop\footnote{https://mlpcp21.github.io/pages/challenge} devoting to propose machine learning methods for preventing and combating pandemics. Furthermore, since the continual development of pre-trained language models from 2018, in recently organized challenges, most participants proposed pre-trained language model-based methods for different tasks.

\section{Discussion}
\label{sec:discussion}

\subsection{Limitations and Concerns}
In this subsection, we will mainly discuss the limitations
of biomedical PLMs and raise some concerns about them.

\paragraph{Misinformation}
The training corpora consist of EHR, and social media may include wrong information. Thus, pre-trained language models pre-trained on them may convey some misinformation~\cite{xie2023faithful}. Furthermore, the biomedical domain itself may have misclassified disease definitions during its development process.
Misinformation has become much more serious in the biomedical domain than in the general domain since this may lead to fatal biomedical decision-making consequences. However, researchers must be aware of the complexity of routinely collected electronic health records, including ways to manage variable completeness. %This paper strongly suggests 
We believe that the predictions from pre-trained language models should be artificially calibrated by biomedical experts before it is used by end-users like patient or the public.

\paragraph{Interpretation issues}
Along with the power of neural networks, there is a growing concern about the interpretability of deep neural networks (DNNs). While in the biomedical domain, the consequence of bad decisions/predictions may be deadly; thus, a well-interpreted model is more crucial. The interpretation in the biomedical domain may come from two aspects: (1) biomedical models should be easily understood, and the predictions could be simulated from the raw input, (2) a (textual) reason should be provided for each prediction. The basic example of the former (a.k.a, transparency \cite{lipton2018mythos}) is decision trees that could clearly illustrate the decision path.  However, such a transparency goal is hardly achieved in modern natural language processing, especially with pre-trained language models. More efforts could be made for the latter; one has to find some textual explanation for each prediction/decision, based on what doctors and patients could make their own decisions.

\paragraph{Identifying causalities from correlations} 
Similar to interpretability, causality may provide the underlying explanation of the model decisions. Causality is crucial in many tasks of biomedical knowledge, \textit{e.g.}, diagnosis, pathology, or systems biology. Causal associations between biological entities, events, and processes are central to most claims of interest; see an early review from \cite{kleinberg2011review}. With automatic causality recognition, it could suggest possible causal connections that may be beneficial for biomedical decisions, which hence greatly reduces the human workload \cite{mihuailua2013biocause}.
% \paragraph{Recall vs. precision} not only about recall, precision matters.

\paragraph{Trade-off between coverage or quality?} There are no large-scaled and high-quality training corpora in the biomedical domain. This means one has to sacrifice its coverage to obtain a high-quality vertical application, or train a general model with large-scaled yet low-quality corpora. Pre-trained language models typically consist of many transformer layers that have many parameters, which usually require a massive amount of plain text. This may lead to a general model with great coverage, but a smaller proportion of high-quality expert knowledge.  

\paragraph{Heterogeneous training data}
For biomedical understandings, there is heterogeneous information, including tables, figures, graphs (fMRI), etc. For example, tables and numbers are crucial in scientific literature. But most PLMs are unable to interpret tables and numbers well. To deeply capture the information in these heterogeneous data, both in-depth data prepossessing and model adaption may be needed. Especially, multi-domain pre-trained language models in biomedical should be paid much more attention. 

\paragraph{Ethics and bias} 
With the rapid development of AI systems and applications in industrial products, it should be aware that they should not introduce any bias for special groups or populations \cite{mehrabi2021survey}, and some of the efforts were taken in the NLP field \cite{zhao2017men,sun2019mitigating,blodgett2020language,garrido2021survey}. This becomes more crucial in these
 sensitive environments in the biomedical domain that involves life-changing decisions, like surgery \cite{rudzicz2020ethics}.
It should ensure that the decisions cannot reflect discriminatory or biased behavior toward specific groups or certain populations. 
A few works have quantified the ethics and bias issues in the domain of pre-trained language models..
\cite{zhang2020hurtful} quantifies biases in clinical contextual word embeddings.  
The reason behind this may arise due to the training itself is biased with respect to various attributes like gender, language, race, age, ethnicity, and marital status. For example, in the MIMIC-III dataset \cite{johnson2016mimic}, one can find: 1) gender bias: males have more heart disease than females, and 2) ethnicity bias: black patients have fewer clinical studies than other groups \cite{kalyan2021ammu}.
Considering the complexity of directly reducing biases in training corpora, existing works explore identifying bias by adversarial training \cite{zhang2020hurtful}, or data augmentation \cite{minot2021interpretable}.
% It is necessary to identify and reduce any form of bias that allows the model to take fair decisions without favoring any group. There are few works that identified and addressed bias in transformer-based biomedical language models. 

% \cite{zhang2020hurtful} further showed that adversarial pretraining debiasing has little impact in reducing bias. Minot et al. [177] proposed an approach based on data augmentation to identify and reduce gender bias in patient notes. This is an area that needs to be explored further to improve reduce bias and improve the fairness in model decisions ] \benyou{rewriting}
\paragraph{Privacy} 
Although most corpora used in biomedical pre-training like scientific publications and social medical are open-access. Some EHRs are private since some organizations do not want to expose their data. For example, clinical records may contain patient visits and  medical history; these sensitive information may bring some physical and mental harm to patients  if exposed \cite{nakamura2020kart}. Note that de-identification of
these sensitive information in EHR records (like  MIMIC III) is not always safe; recent works showed that there is data leakage from pre-trained models in the general domain, \textit{i.e.}  recovering Personal Health Information (PHI) from  pre-trained models trained from  is possible ~\cite{lehman2021does}. Therefore, we warn the public release of  pre-trained models, if PHI is risky to be exposed. Recently, Nakamura et al \cite{nakamura2020kart} proposed a  framework called `KART' to assess the  sensitive information from pre-trained biomedical language models using various attacks. Also, the federated learning \cite{yang2019federated,li2020federated} framework may help when different organizations and end-users could collaboratively learn a shared prediction model while keeping all the training data on a private side.
% \paragraph{ xxxxx }
% \benyou{anyone who can elaborate `The whole story about the proposed algorithms, not just cherry-pick...'}
% Inspired by ExplainaBoard \url{http://explainaboard.nlpedia.ai/}. 
% Recent papers on systematic evaluation from Google, ``The Benchmark Lottery'', 
%  https://www.researchgate.net/profile/Arlene-Gallagher-2/publication/277894779_Data_resource_profile_Clinical_Practice_Research_Datalink_CPRD/links/55925de908ae15962d8e5e4f/Data-resource-profile-Clinical-Practice-Research-Datalink-CPRD.pdf
% action="https://www.softconf.com/emnlp2021/papers/login/scmd.cgi?scmd=login"
\subsection{Future trends}
\label{sec:trend}

We further suggest some future trends in this subsection.
\paragraph{Standardized benchmark}
In general NLP fields,  evaluation criteria and standard benchmarks are a driving force for the NLP community.  For example, BERT \cite{devlin2018bert} was widely accepted in benchmarks \cite{wang2018glue,rajpurkar2016squad} makes it spread to various tasks in NLP.  On the other hand, lacking an effective evaluation criterion is one of the  bottlenecks of text generation \cite{celikyilmaz2020evaluation}. In the biomedical domain, various pre-trained models and their fine-tuning applications have been proposed (as introduced in Sec. \ref{sec:pertraining} and Sec. \ref{sec:finetuning}). However, they are generally not well-compared. 
Although a few efforts have been made to standardize  benchmarks for biomedical pre-trained models, which include but are not limited to \cite{zhang2021cblue,gu2020domain}.
This becomes much more difficult in the cross-discipline domain like the biomedical domain since papers are usually from different communities like informatics, medicine, and computer science. An open standardized and well-categorized  benchmark (like in \cite{lewis2020pretrained,li2023huatuo}) should be proposed to make use of the advantages of each work and collaboratively push the development of biomedical NLP. This survey is the first step to introducing the biomedical pre-trained language models and their applications in downstream tasks. More efforts are expected to be made to design fine-grained taxonomy and define each SOTA approach in various applications, based on what incremental work could be better evaluated.

\paragraph{Open culture}
In general NLP fields, a lot of effort is made to make better-available resources, including open-source resources (released training data and models), and fairly implemented approaches. In addition, open culture  makes  researchers could easily contribute to the community.
For example, the NLP community has been largely developed thanks to the model collections \cite{wolf2020huggingfaces,fan2017matchzoo}.  In addition, most accepted papers in top conferences tend to release codes, models, and data.
Biomedical NLP fields also benefit a lot from such open culture and standard systematic evaluations. For instance, pre-trained models in Huggingface \footnote{\url{https://huggingface.co/}} largely fascinated their applications in the biomedical domain.

\paragraph{Efficiency on pre-trained language models} Compared to previous SOTA methods training from scratch based on neural networks such as LSTM or CNN, before Transformer, pre-trained language models are much bigger in terms of model scale and much slower due to the increasing number of parameters. This is more expensive for deployment that requires more computing resources. One may have to refer to \cite{tay2020efficient} for efficient transformers. For example, current work explores quantization ~\cite{zhang2020ternarybert,bai2020binarybert}, weights pruning~\cite{hou2020dynabert}, and knowledge distillation~\cite{jiao2020tinybert,sanh2020distilbert} for BERT. Therefore, in the biomedical domain, pre-training language models with lower computation complexity are a direction needed to pay more attention.

\paragraph{Generation-based PLMs are under-investigated}
% underestimated
Most works focused on encoder-based models, and a few works involve decoder or encoder-decoder architectures that enable generations. This may be due to the fact that classification tasks may be widely used in downstream biomedical tasks. 
Very recently, \cite{kraljevic2021medgpt} proposes GPT models using temporal electronic health records and  \cite{phan2021scifive} trained a T5-based biomedical pre-trained model.
We believe that generation-based PLMs (e.g. GPT, T5, and BART) have great potential in the biomedical domain, but it is currently under-investigated. {\color{black} Very recently, we have witnessed some work that uses large generation-based PLMs in the biomedical domain, see especially BIOGPT~\cite{luo2022biogpt},  PubMedGPT~\footnote{https://crfm.stanford.edu/2022/12/15/pubmedgpt.html}, and Flan-PaLM \cite{singhal2022large}.}

\paragraph{Few-shot learning}
\cite{Perez2021} evaluates the few-shot ability of LMs when held-out examples are unavailable for choosing hyperparameters or prompts and finds that LMs do not perform well compared to random selection and under-perform selection based on held-out examples. 
In other words, previous methods overestimate the few-shot capability of LMs based on more realistic settings. 
This might be even worse for biomedical LMs. 
% \todo[inline]{\benyou{check papers of \cite{kalyan2020secnlp}, \cite{khattak2019survey} and  other papers and summarize the related Corpora here}}

\paragraph{In non-English or low-resource language} Most works in biomedical  pre-trained language models are with English corpora, and a few about Chinese \cite{zhang2020conceptualized}, German \cite{10.1093/bioinformatics/btaa668}, Japanese \cite{kawazoe2020clinical,wada2020pre}, Spanish \cite{akhtyamova2020named,miranda2020named,akhtyamova2020testing,lopez2021pre}, Korean \cite{kim2020korean}, Russian \cite{tutubalina2021russian}, Italian \cite{catelli2020crosslingual}, Arabic \cite{antoun2020arabert,boudjellal2021abioner}, French \cite{copara2020contextualized}, Portuguese \cite{schneider-etal-2020-biobertpt,schneider2021gpt} etc. For non-English biomedical tasks, there are two mainstream solutions: a single non-English language paradigm and a multi-linguistic paradigm. The former uses a single language, while the latter uses multiple languages. The multi-linguistic paradigm could be more beneficial for low-resource since biomedical knowledge itself is language-independent, and information in a second language could be complementary. 

\paragraph{Multi-modal pre-training}
Multi-modal pre-training \cite{radford2021learning,ramesh2021zeroshot} has attracted much attention in image classification and generation tasks, because it only needs cheap but large-scale publicly available online resources. This shows great potential in machine learning since less human annotation is needed. It is expected that various modalities could provide complementary information. For example, making use of biomedical codes,  medical images, waveforms, and genomics in pre-training  models would be beneficial but challenging due to its multi-modal nature. 
% [PDF] DLI-IT: A deep learning approach to drug label identification through image and text embedding% MUFASA: Multimodal Fusion Architecture Search for Electronic Health Records

\paragraph{Injecting biomedical knowledge in pre-trained language model}
Before the pre-training age, some works \cite{petroni2019language} have explored injecting medical knowledge into embeddings that provide potentially better ML features. Recently, existing work claims that pre-trained language models could be a soft knowledge base that captures knowledge. Despite this, \cite{colon2021combining,xu2021k,zhang2023injecting} also tried to inject knowledge into pre-trained language models explicitly. In the biomedical domain, which is knowledge-intensive; knowledge-injected models could have great potential in the future. For example, \cite{michalopoulos2020umlsbert} integrates domain knowledge (\textit{i.e.}, Unified Medical Language System (UMLS) Metathesaurus) in pre-training via a knowledge augmentation strategy.
% Learning a Health Knowledge Graph from Electronic Medical Record% Scientific Language Models for Biomedical Knowledge Base Completion: An Empirical Study
% https://arXiv.org/abs/2106.09700

\paragraph{Interpretability in biomedical PLMs}
Neural networks were criticized for having limited interpretability. Pre-trained language models are typically huge neural network models, which is more challenging in terms of interpretability. One may expect to understand the working mechanism related to the medical characteristics in pre-trained language models. For example, probing pre-trained language models have been widely used to understand pre-trained language models, see \cite{vulic2020probing,koto2021discourse,wu2020perturbed,lin2020birds}. For biomedical pre-trained language models, \cite{alghanmi2021probing} aims to evaluate pre-trained language models about the disease knowledge. 
\cite{vig2021bertology} exhaustively analyzing attention in protein Transformer models, providing many interesting findings to understand the working mechanisms better. 
\cite{jin2019probing} conducts some probing experiments to determine what additional information is carried intrinsically by BioELMo and BioBERT.
{
Another direction of interpretability in  the biomedical field is to mine the  causality (rather than correlation) due to its crucial relevance in establishing clinical interventions and public health policies. Correlation merely indicates a statistical relationship between two variables, which is valuable in generating hypotheses, but provides limited insights into the underlying mechanisms. Conversely, causality moves beyond associative relationships to delineate direct cause-effect relationships. This deeper understanding is pivotal in biomedical research, as it provides the foundation for intervention studies and enables the development of effective treatments. Identifying a causal relationship, for instance, between a specific genetic mutation and a disease, allows for targeted therapies and personalized medicine. Thus, while correlation provides a starting point for scientific exploration, it is the discernment of causality that truly advances biomedical knowledge and contributes to the development of life-saving interventions.}

% We believe that more efforts are expected for interpretability in biomedical PLMs.

%Corrected
%same
%1/2
%Saved
%674 words
{\color{black}

\paragraph{Dialogue-based medical consultation}

Transitional medical consultation is to obtain medical suggestions and treatment from clinicians. Recently, AI communities have tried to solve medical consultations through online ways using artificial intelligence tools, especially for pre-consultation and psychological treatment. Meanwhile, online medical consultation is another natural playground for current state-of-the-art AI algorithms under the `AI for Science' Trend.  Some existing work formulate medical consultation as a question-answering task in the sense that it could leverage many existing question-answering pairs. However,  medical consultation is complicated in the sense that static and single-turn question-answering pairs could not solve individually-dependent consultation; especially, medical consultations are more likely to be dependent on individual backgrounds, like historical diseases and treatment, genes, and dietary habits. We believe dialogue-based consultation systems could better fit medical scenarios than single-question-answering systems.
Existing medical dialogue systems have shown some potential but also perform much worse than the expectation. Very recently, motivated by the great success of Open AI ChatGPT which uses giant language models to meet human consultation needs, we believe using giant medical language models could largely improve the quality of medical consultation. More optimistically,  we believe this might, at least to some extent, revolutionize the current medical industry, see \cite{kung2022performance,https://doi.org/10.48550/arxiv.2301.10035} as some preliminary work.

\paragraph{Scale law in medical PLMs}

Not only in dialogue systems, large-scale PLMs are as popular as it in the general domain. The reasons are twofold. First, the adaption of SOTA PLMs to the medical domain takes time, and it is usually more than half a year late after a general PLM is released. Secondly, non-generative language models are insensitive to huger scales,  and their performance becomes saturated when they are beyond 24 layers (the scale of BERT-large). Meanwhile, most works use non-generative language models (e.g., BERT, RoBERTa, and Electra) in the biomedical domain while very few generative language models are used. With huger language models, we might see some emergent abilities in medical applications. Fortunately, we have witnessed a preliminary sign that we started to investigate large language models in medical/clinical tasks~\cite{luo2022biogpt,singhal2022large,lievin2022can}, especially BIOGPT~\cite{luo2022biogpt},  PubMedGPT~\footnote{https://crfm.stanford.edu/2022/12/15/pubmedgpt.html}, and Flan-PaLM \cite{singhal2022large}.

% \zhihong{Large Language Models Encode Clinical Knowledge: https://arxiv.org/abs/2212.13138}

% \zhihong{Can large language models reason about medical questions?: https://arxiv.org/abs/2207.08143}

% PubMedGPT: https://www.mosaicml.com/blog/introducing-pubmed-gpt

% https://academic.oup.com/bib/advance-article-abstract/doi/10.1093/bib/bbac409/6713511 BIOGPT  

% \benyou{reasoning needed}

% https://arxiv.org/abs/2207.08143

% \benyou{modular}

% \benyou{text normalization}

% \benyou{privacy preserving and data sharing?}

% \benyou{long text: https://arxiv.org/abs/2201.11838  Clinical-Longformer and Clinical-BigBird: Transformers for long clinical sequences    \cite{li2022clinical}

% % https://arxiv.org/pdf/2210.10341.pdf

% https://academic.oup.com/jamia/advance-article/doi/10.1093/jamia/ocac225/6855145  long text  \cite{10.1093/jamia/ocac225}  A comparative study of pre-trained language models for long clinical text}

\paragraph{Data collection and sharing protocol} 

The need for data in biomedicine is tremendous since data is the fuel for learning.
The reasons that hinder medical data collection and sharing are manyfold. Firstly, it has a legal risk regarding privacy issues, especially because this also involves cross-border or cross-organization data transfer. Secondly, an individual hospital might adopt different standards in terminology, this issue becomes more severe in developing counties than in developed counties. The merge between two data sources will be difficult due to the inconsistency of terminology. Therefore, it requires a well-defined protocol to deal with this, including solving terminology inconsistency and data privacy. From an NLP perspective, we need to normalize word terminology and data desensitization. For other perspectives, this needs some high-level data-sharing protocol, e.g., federated learning \cite{10.1145/3298981}.

\paragraph{Dealing with long sequences}

The computation of self-attentions in Transformers is quadratic to the length of sequences. This means the longer sequences would necessarily make transformer-based PLMs much more time-consuming. 
Sequences in biomedicine are usually  long; it varies from DNA/protein sequences to texts. First,  DNA/protein sequences are long especially for big protein sequence  which has lengths that are longer than 4096, i.e., the typical maximum sequence length in language models.  
% it needs to design specific PLMs to deal with long sequences. 
Biomedical texts,  including EHRs, biomedical encyclopedias, and biomedical literature, are usually longer than the general domain (e.g., the maximum sentence length used in GLUE is usually 128); { for instance, there is usually  text redundancy in  clinical notes}. Therefore, we need to design more efficient and effective models tailored to long sequences, see some existing recent works \cite{li2023comparative,li2022clinical}.

% 脱敏

% \benyou{RAG}

% \benyou{causality!}

% \paragraph{More investigation from biomedical sequences}

% Roshan Rao, Joshua Meier, Tom Sercu, Sergey Ovchinnikov, and Alexander Rives. Transformer
% protein language models are unsupervised structure learners. In 9th International Conference on
% Learning Representations, ICLR 2021, Virtual Event, Austria, May 3-7, 2021. OpenReview.net,
% 2021b. URL https://openreview.net/forum?id=fylclEqgvgd.   \cite{rao2020transformer}

% \zhihong{ to be added}

}
% https://arxiv.org/abs/2204.03905 BioBART \cite{yuan2022biobart}

% \benyou{more citations in this section}

% explore the border, what PLMs can do and what they cannot

% \paragraph{knowledge transfer}
% Transfer Learning from Medical Literature for Section Prediction in Electronic Health Records

% More complicated reasoning
% \paragraph{explore what PLMs do and what it cannot}

% \paragraph{engineering for robust medical nlp } Evaluation and accurate diagnoses of pediatric diseases using artificial intelligence

% \paragragh{interpretability}

% \paragraph{It lacks large-scale and high-quality Corpora}
% The biomedical corpora have many issues, first,

% PLMs as memorization engines,  

\section{Conclusion}
\label{sec:conclusion}
This paper systematically summarizes recent advances of pre-trained language models in the biomedical domain, including background, why and how pre-trained language models are used in the biomedical domain, existing biomedical pre-trained language models, data sources in the biomedical domain, application of pre-trained language models in various biomedical downstream tasks. Furthermore, we also discuss some limitations and future trends. Finally, we expect that the pre-trained language model in the general NLP domain could also help the specific biomedical domain.

\section*{Acknowledgment}

This work is supported by Chinese Key-Area Research and Development Program of Guangdong Province (2020B0101350001), the Shenzhen Science and Technology Program (JCYJ20220818103001002), the Guangdong Provincial Key Laboratory of Big Data Computing, The Chinese University of Hong Kong, Shenzhen, Shenzhen Key Research Project (C10120230151) and Shenzhen Doctoral Startup Funding (RCBS20221008093330065).

%
% ---- Bibliography ----
%
% BibTeX users should specify bibliography style 'splncs04'.
% References will then be sorted and formatted in the correct style.
%

{
% \fontsize{15pt}{18pt}\selectfont
% \scriptsize
% \fontsize{15pt}{18pt}
% \fontsize{10pt}{12pt}\selectfont
% \bibliographystyle{ACM-Reference-Format}
\bibliographystyle{abbrv}
\bibliography{bplm}
}

\end{document}